%% file: main.tex
\documentclass[10pt,twocolumn,letterpaper]{article}

\usepackage{cvpr}             

\usepackage{graphicx}
\usepackage{amsmath}
\usepackage{amssymb}
\usepackage{booktabs}
\usepackage[accsupp]{axessibility}
\usepackage{fmtcount}

\input{packages}

\input{macros}

\usepackage[pagebackref,breaklinks,colorlinks]{hyperref}

\usepackage[capitalize]{cleveref}
\crefname{section}{Sec.}{Secs.}
\Crefname{section}{Section}{Sections}
\Crefname{table}{Table}{Tables}
\crefname{table}{Tab.}{Tabs.}

\newcommand{\projectname}[0]{Infinigen}
\newcommand{\parbf}[1]{\par \noindent \textbf{#1}.}

\def\cvprPaperID{1975}
\def\confName{CVPR}
\def\confYear{2023}

\begin{document}

\title{Infinite Photorealistic Worlds using Procedural Generation }

\author{
Alexander Raistrick$^*$,\, Lahav Lipson$^*$,\, Zeyu Ma$^*$, {\small($^*$equal contribution; alphabetical order)} \\
Lingjie Mei,\, 
Mingzhe Wang,\,
Yiming Zuo,\, 
Karhan Kayan,\,
Hongyu Wen,\,
Beining Han,\\
Yihan Wang,\,
Alejandro Newell\footnotemark[2],\,
Hei Law\footnotemark[2],\,
Ankit Goyal\footnotemark[2],\,
Kaiyu Yang\footnotemark[2],\,
Jia Deng \\
Department of Computer Science, Princeton University\\
}

\maketitle
\input{texts/0-abstract}
\footnotetext[2]{work done while a student at Princeton University}
\input{texts/1-introduction}

\input{texts/2-related_work}
\input{texts/3-method}

\input{texts/4-experiments}
\input{texts/6-contributions}

{\small
\bibliographystyle{ieee_fullname}
\bibliography{egbib}
}

\clearpage

\appendix

\input{supp.tex}

\end{document}

%% file: packages.tex
\usepackage{adjustbox}
\usepackage{array}
\usepackage{booktabs}
\usepackage{colortbl}
\usepackage{wrapfig}
\usepackage{hhline}
\usepackage{multirow}
\usepackage{subcaption} 

\usepackage{amsmath,amsfonts,amssymb}
\usepackage{bm}
\usepackage{nicefrac}
\usepackage{microtype}

\usepackage{changepage}
\usepackage{extramarks}
\usepackage{fancyhdr}
\usepackage{lastpage}
\usepackage{setspace}
\usepackage{soul}
\usepackage{xspace}
\usepackage{afterpage}

\usepackage{url}

\usepackage{enumerate}
\usepackage{enumitem}  

\usepackage{makecell}

\usepackage{pifont} 

\usepackage{float}
\usepackage{graphicx}
\usepackage{multirow}
 
\usepackage{listings}

\definecolor{codegreen}{rgb}{0,0.6,0}
\definecolor{codegray}{rgb}{0.5,0.5,0.5}
\definecolor{codepurple}{rgb}{0.58,0,0.82}
\definecolor{backcolour}{rgb}{0.95,0.95,0.92}
\definecolor{red}{rgb}{0.9,0,0}

\lstdefinestyle{mystyle}{
    backgroundcolor=\color{backcolour},   
    commentstyle=\color{codegreen},
    keywordstyle=\color{magenta},
    numberstyle=\tiny\color{codegray},
    stringstyle=\color{codepurple},
    basicstyle=\ttfamily\footnotesize,
    breakatwhitespace=false,         
    breaklines=true,                 
    captionpos=b,                    
    keepspaces=true,                 
    numbers=left,                    
    numbersep=5pt,                  
    showspaces=false,                
    showstringspaces=false,
    showtabs=false,                  
    tabsize=2
}

%% file: macros.tex
\newcolumntype{L}[1]{>{\raggedright\let\newline\\\arraybackslash\hspace{0pt}}m{#1}}
\newcolumntype{C}[1]{>{\centering\let\newline\\\arraybackslash\hspace{0pt}}m{#1}}
\newcolumntype{R}[1]{>{\raggedleft\let\newline\\\arraybackslash\hspace{0pt}}m{#1}}

\newcommand{\sectapp}[1]{Appendix~\ref{#1}}

\newcommand{\fig}[1]{Fig.~\ref{#1}}

\newcommand{\ignore}[1]{}

\DeclareMathAlphabet{\mathbfit}{OML}{cmm}{b}{it}

\makeatletter
\DeclareRobustCommand\onedot{\futurelet\@let@token\@onedot}
\def\@onedot{\ifx\@let@token.\else.\null\fi\xspace}

\makeatother

\definecolor{MyDarkBlue}{rgb}{0,0.08,1}
\definecolor{MyAqua}{rgb}{0,0.7,0.7}
\definecolor{MyDarkGreen}{rgb}{0.02,0.6,0.02}
\definecolor{MyDarkRed}{rgb}{0.8,0.02,0.02}
\definecolor{MyDarkOrange}{rgb}{0.40,0.2,0.02}
\definecolor{MyPurple}{RGB}{111,0,255}
\definecolor{MyRed}{rgb}{1.0,0.0,0.0}
\definecolor{MyGold}{rgb}{0.75,0.6,0.12}
\definecolor{MyDarkgray}{rgb}{0.66, 0.66, 0.66}

\usepackage{caption}
\DeclareCaptionLabelFormat{cont}{#1~#2\alph{ContinuedFloat}}
\usepackage{sidecap}
\usepackage{pifont}

 \newcommand{\galleryRowCompare}[7]{
   \vspace*{-0.075cm}
   \includegraphics[width=0.2\textwidth]{#1/#2} &
   \includegraphics[width=0.2\textwidth]{#1/#3} &
   \includegraphics[width=0.2\textwidth]{#1/#4} &
   \includegraphics[width=0.2\textwidth]{#1/#5} &
   \includegraphics[width=0.2\textwidth]{#1/#6} &
   \includegraphics[width=0.2\textwidth]{#1/#7}
   \tabularnewline 
 }

%% file: texts/0-abstract.tex
\begin{abstract}
We introduce \projectname{}, a procedural generator of photorealistic 3D scenes of the natural world. \projectname{} is entirely procedural: every asset, from shape to texture, is generated from scratch via randomized mathematical rules, using no external source and allowing infinite variation and composition.  \projectname{} offers broad coverage of objects and scenes in the natural world including plants, animals, terrains, and natural phenomena such as fire, cloud, rain, and snow. \projectname{} can be used to generate unlimited, diverse training data for a wide range of computer vision tasks including object detection, semantic segmentation, optical flow, and 3D reconstruction. We expect \projectname{} to be a useful resource for computer vision research and beyond. Please visit \href{https://infinigen.org}{infinigen.org} for videos, code and pre-generated data. 
\end{abstract}

%% file: texts/1-introduction.tex
\vspace{-7mm}
\section{Introduction}
\label{sec:intro}

Data, especially large-scale labeled data~\cite{deng2009imagenet,lin2014microsoft}, has been a critical driver of progress in computer vision. At the same time, data has also been a major challenge, as many important vision tasks remain starved of high-quality data. This is especially true for 3D vision, where accurate 3D ground truth is difficult to acquire for real images.

Synthetic data from computer graphics is a promising solution to this data challenge. Synthetic data can be generated in unlimited quantity with high-quality labels. Synthetic data has been used in a wide range of tasks~\cite{gta5koltun2016playing, crestereo, law2022label, bai2022deep, lipson2022coupled, ma2022multiview, deitke2022procthor}, with notable successes in 3D vision, where models trained on synthetic data can perform well on real images zero-shot~\cite{teed2021droid, teed2022deep, lipson2021raft, teed2020raft, wang2021tartanvo, haugaard2022surfemb, teed2021raft}. 

Despite its great promise, the use of synthetic data in computer vision remains much less common than real images. We hypothesize that a key reason is the limited diversity of 3D assets: for synthetic data to be maximally useful, it needs to capture the diversity and complexity of the real world, but existing freely available synthetic datasets are mostly restricted to a fairly narrow set of objects and shapes, often driving scenes (e.g. \cite{gta5koltun2016playing, huang2018deepmvs}) or human-made objects in indoor environments (e.g. \cite{deepfurniture, robotrix}). \looseness=-1

In this work, we seek to substantially expand the coverage of synthetic data, particularly objects and scenes from the natural world. We introduce \projectname{}, a procedural generator of photorealistic 3D scenes of the natural world. Compared to existing sources of synthetic data, \projectname{} is unique due to the combination of the following properties: 
\begin{itemize}
    \item \emph{Procedural:} \projectname{} is not a finite collection of 3D assets or synthetic images; instead, it is a \emph{generator} that can create infinitely many distinct shapes, textures, materials, and scene compositions. Every asset, from shape to texture, is entirely procedural, generated from scratch via randomized mathematical rules that allow infinite variation and composition. This sets it apart from datasets or dataset generators that rely on external assets. \looseness=-1
    
    \item \emph{Diverse:} \projectname{} offers a broad coverage of objects and scenes in the natural world, including plants, animals, terrains, and natural phenomena such as fire, cloud, rain, and snow. 
    
    \item \emph{Photorealistic:} \projectname{} creates highly photorealistic 3D scenes. It achieves high photorealism by procedurally generating not only coarse structures but also fine details in geometry and texture. 
    
    \item \emph{Real geometry}: unlike in video game assets, which often use texture maps to fake geometrical details (e.g.\@ a surface appears rugged but is in fact flat), all geometric details in \projectname{} are real. This ensures accurate geometric ground truth for 3D reconstruction tasks. 
    
    \item \emph{Free and open-source:} \projectname{} builds on top of Blender~\cite{blender}, a free and open-source graphics tool. \projectname{}'s code is released for free under the BSD\footnote[3]{See \url{https://opensource.org/license/bsd-3-clause/}} license. Anyone can freely use \projectname{} to obtain unlimited assets and renders. 
\end{itemize}

\vspace{-2mm}
\begin{figure*}
\includegraphics[width=\linewidth]{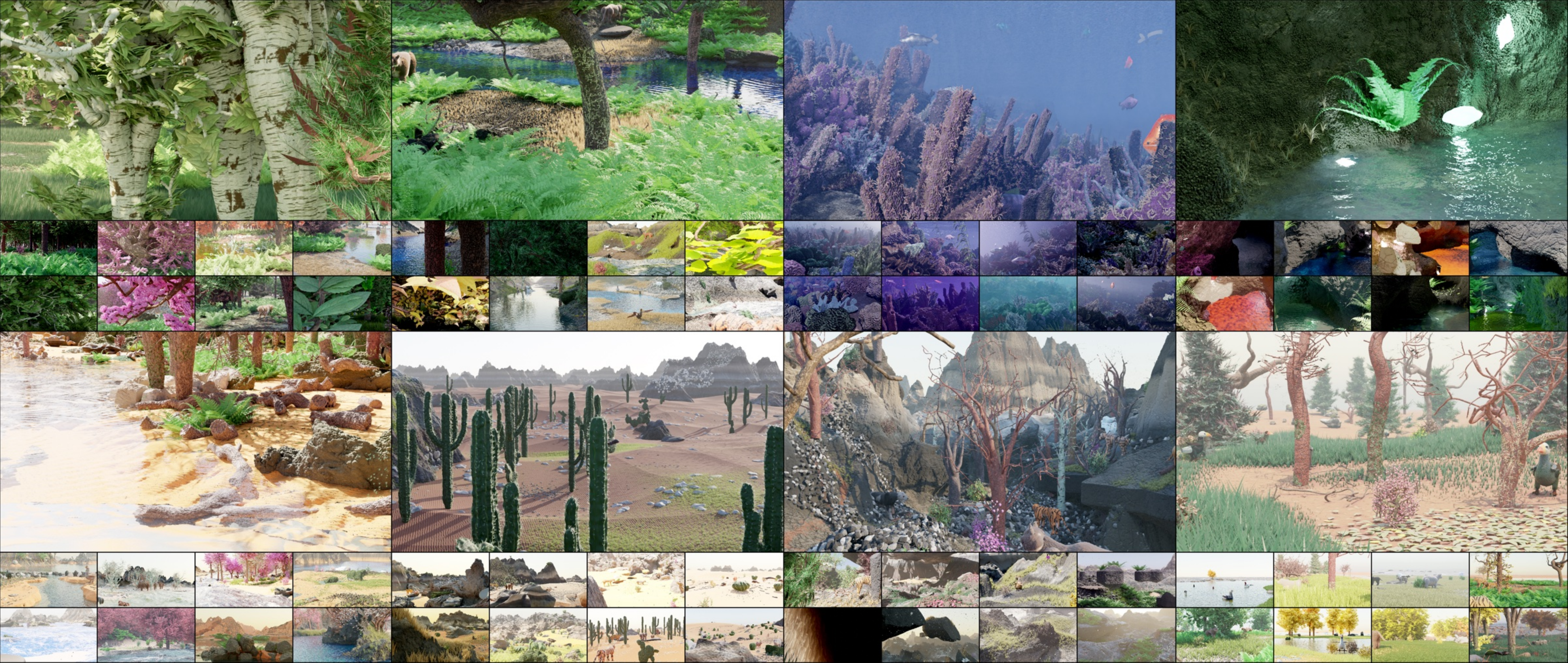}
 \caption{Randomly generated, \textit{non cherry-picked} images produced by our system. From top left to bottom right: Forest, River, Underwater, Caves, Coast, Desert, Mountain and Plains. See Appendix for more samples.}
\label{fig:teaser}
\end{figure*}

\projectname{} focuses on the natural world for two reasons. First, accurate perception of natural objects is demanded by many applications, including geological survey, drone navigation, ecological monitoring,  rescue robots, agriculture automation, but existing synthetic datasets have limited coverage of the natural world. Second, we hypothesize that the natural world alone can be sufficient for pretraining powerful ``foundation models''---the human visual system was evolved entirely in the natural world; exposure to human-made objects was likely unnecessary. 

\projectname{} is useful in many ways. It can serve as a generator of unlimited training data for a wide range of computer vision tasks, including object detection, semantic segmentation, pose estimation, 3D reconstruction, view synthesis, and video generation. Because users have access to all the procedural rules and parameters underlying each 3D scene, \projectname{} can be easily customized to generate a large variety of task-specific ground truth. 
\projectname{} can also serve as a generator of 3D assets, which can be used to build simulated environments for training physical robots as well as virtual embodied agents. The same 3D assets are also useful for 3D printing, game development, virtual reality, film production, and content creation in general. 

We construct \projectname{} on top of Blender~\cite{blender}, a graphics system that provides many useful primitives for procedural generation. Utilizing these primitives we design and implement a library of procedural rules to cover a wide range of natural objects and scenes. In addition, we develop utilities that facilitate creation of procedural rules and enable all Blender users including non-programmers to contribute; the utilities include a transpiler that automatically converts Blender node graphs (intuitive visual representation of procedural rules often used by Blender artists) to Python code. We also develop utilities to render synthetic images and extract common ground truth labels including depth, occlusion boundaries, surface normals, optical flow, object category, bounding boxes, and instance segmentation. Constructing \projectname{} involves substantial software engineering: the latest main branch of the \projectname{} codebase consists of 40,485 lines of code.  

In this paper, we provide a detailed description of our procedural system. We also perform experiments to validate the quality of the generated synthetic data; our experiments suggest that data from \projectname{} is indeed useful, especially for bridging gaps in the coverage of natural objects. Finally, we provide an analysis on computational costs including a detailed profiling of the generation pipeline.  

We expect \projectname{} to be a useful resource for computer vision research and beyond.  In future work, we intend to make \projectname{} a living project that, through open-source collaboration with the whole community, will expand to cover virtually everything in the visual world. \looseness=-1

%% file: texts/2-related_work.tex
\begin{figure*}[th!]
\includegraphics[width=\textwidth]{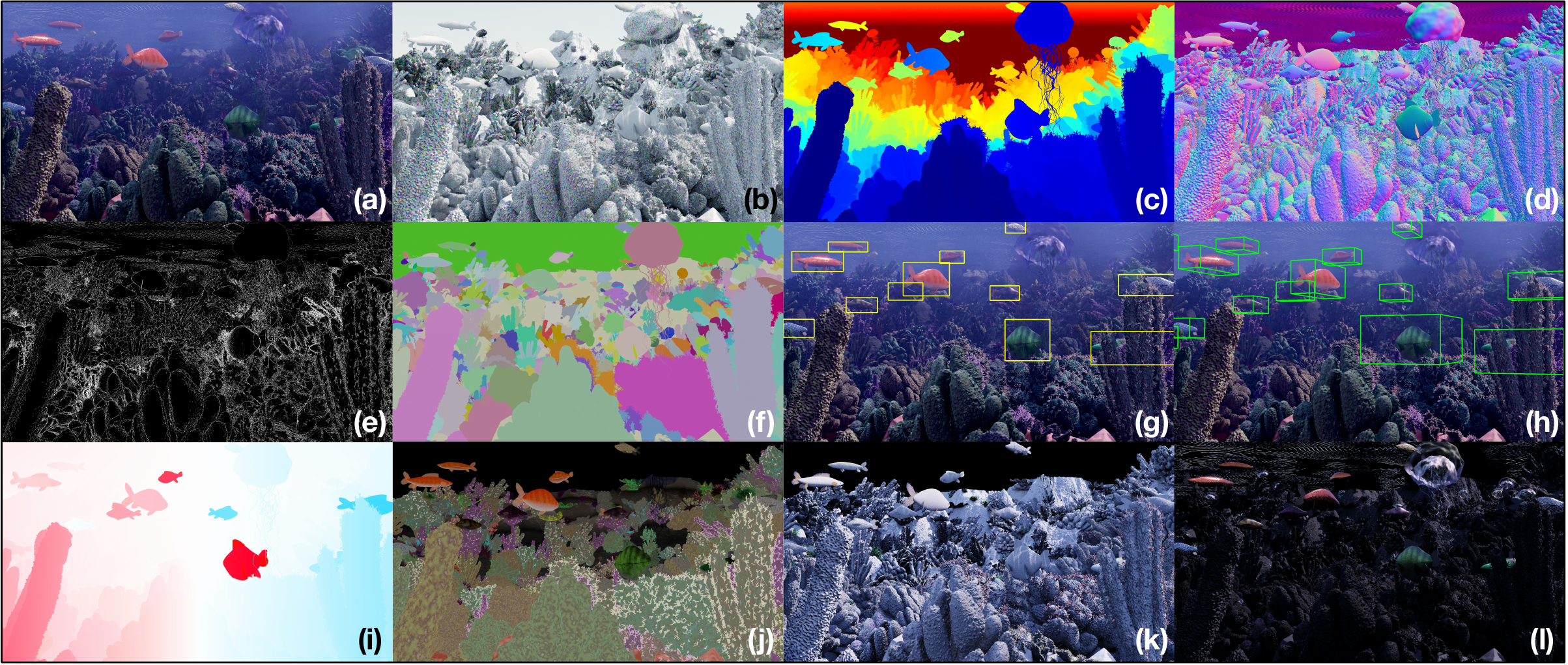}
\caption{For each image (a), we have a high-res mesh (b), which readily yields Depth (c), Surface Normals (d), Occlusion Boundaries (e), Instance Segmentation masks (f), and 2D / 3D bounding boxes (g/h). From rendering metadata, we obtain Optical Flow (i), and material parameters such as Albedo (j), Lighting Intensity (k) and Specular Reflection (l).}
\label{fig:ground_truth}
\end{figure*}

\section{Related Work}
\label{sec:relatedwork}
\newcommand{\crossmark}{\ding{53}}
\newcommand{\voxel}{\ding{114}}
\begin{table*}
\label{tab:relateddatasets}
\centering
\resizebox{\linewidth}{!}{
\begin{tabular}{l|c|c|c|c|c|c|c|c|l} \toprule
\multirow{2}{*}{Synthetic Dataset} & \multirow{2}{*}{Domain} & \# Triangles & \# Scenes & \# Assets & Free & Procedural & Procedural & Provides & \multirow{2}{*}{External Asset Source}  \\ 
 & & Per-Scene & in Total & in Total & Assets & Arrangement & Assets & Procedural Code &\\
 \midrule
GTA-V~\cite{gta5koltun2016playing} & Driving, Urban & - & - & - & No & No & No & N/A & Grand Theft Auto \\
MOTSynth~\cite{fabbri2021motsynth} & Urban & - & - & - & No & No & No & N/A & Grand Theft Auto \\
MVS-Synth~\cite{huang2018deepmvs} & Driving, Urban & - & - & - & No & No & No & N/A & Grand Theft Auto \\
DeformingThings4D~\cite{li20214dcomplete} & Animals/Humanoids & - & 2K & 2K & Yes & No & No & N/A & Adobe Mixamo~\cite{adobemixamo}\\
DeepFurniture~\cite{deepfurniture} & Indoor & - & 20K & - & No & No & No & N/A & Professional Designers \\
Robotrix~\cite{robotrix} & Indoor & - & 16 & - & No & No & No & N/A & UE4Arch, UnrealEngine Marketplace~\cite{unrealenginemarket} \\ 
SUNCG~\cite{suncg} (+~\cite{cgpbr,zhang2017physically,li2018cgintrinsics}) & Indoor & - & 46K & 2.6K & No & No & No & N/A & Planner5D\cite{planner5d} \\
TartanAir~\cite{wang2020tartanair} & In/Outdoor, Natural/Urban & - & 30 & - & No & No & No & N/A & UnrealEngine Marketplace~\cite{unrealenginemarket} \\ 
 Hypersim~\cite{hypersim} & Indoor & 100K-11M  & $461$ & 59K & No (\$6000) & No & No & N/A & Evermotion Architectures~\cite{evermotionarch} \\
 OpenRooms~\cite{li2021openrooms} & Indoor & 1M  & 1.3K & 3K & No (\$500) & No & No & N/A & Scan2CAD~\cite{avetisyan2019scan2cad}, ShapeNet~\cite{shapenet}, Adobe Stock~\cite{adobestock} \\
 Sintel~\cite{sintel} & Medieval, Natural & 300K  & 27 & - & No (\$12) & No & No & N/A & Blender Foundation~\cite{blender} \\
 Spring~\cite{mehl2023spring} & Natural & - & 47 & - & No (\$12) & No & No & N/A & Blender Foundation~\cite{blender} \\
  Structured3D~\cite{structured3d} & Indoor & - & 22K & 472K & No & No & No & N/A & Professional Designers \\
  SceneNet-RGBD~\cite{scenenet_rgbd} & Indoor & 420K & 57 & 5.1K & Yes & No & No & N/A & ShapeNet~\cite{shapenet}, SceneNet~\cite{scenenet} \\
 3D-Front~\cite{3dfront} & Indoor & 60K & 19K & 13K & Yes & No & No & N/A & 3D-FUTURE~\cite{3dfuture} \\ 
Jiang et al.~\cite{jiang2018configurable} & Indoor & - & $\infty$ & 54K & No & Yes & No & No & ShapeNet~\cite{shapenet}, Planner 5D~\cite{planner5d}\\
 InteriorNet~\cite{interiornet} & Indoor & - & 22M & 1M & No & Yes & No & No & Manufacturers / Kujiale\cite{kujiale} \\
  FaceSynthetics~\cite{wood2021fake} & Faces & 7.4K & - & $\infty$ & No & N/A & Partial & No & Artist-Created Faces (textures, hair, clothing)\\
 Meta-Sim2~\cite{metasim2} & Driving, Urban & - & $\infty$  & $\infty$ & No & Yes & Partial & No & -\\
 Synscapes~\cite{wrenninge2018synscapes, tsirikoglou2017procedural} & Driving, Urban & - & 25K & - & No & Yes & Partial & No & 7D-Labs~\cite{7dlabs} \\
 ProcSy~\cite{procsy} & Driving, Urban & - & $\infty$ & $\infty$ & Yes & Yes & Partial & No & CityEngine, OpenStreetMap~\cite{haklay2008openstreetmap}, Manual Annotation\\ 
 ProcTHOR~\cite{deitke2022procthor} & Indoor & - & $\infty$ & 1.6K & Yes & Yes & No & Yes & AI2-THOR~\cite{ai2thor}, Professional Designers \\ 
 Kubric~\cite{kubric} & Scattered Objects & 161K  & $\infty$ & 52K & Yes & Yes & No & Yes & ShapeNet~\cite{shapenet}, Google Scanned Objects~\cite{googlescannedobjects} \\ 
 \midrule
\multirow{2}{*}{\projectname{} (Ours)} & \multirow{2}{*}{Natural} & Dynamic & \multirow{2}{*}{$\infty$} & \multirow{2}{*}{$\infty$} & \multirow{2}{*}{Yes} & \multirow{2}{*}{Yes} & \multirow{2}{*}{Yes} & \multirow{2}{*}{Yes} & \multirow{2}{*}{None} \\
& & (16M @ 1080p) & & & & & & &\\
 \bottomrule
\end{tabular}
}

\caption{Comparison to existing synthetic datasets or generators. Ours is entirely procedural, relying on no external assets, and can produce infinite original assets and scenes. Many existing datasets use external, static asset libraries. Procedural generation is often limited to object placement or a subset of objects. The vast majority of datasets are also restricted to the built environment, especially indoor scenes. In terms of accessibility, many do not provide free assets or make code available. Many works do not report average triangles per scene; where possible, we calculate this using generous assumptions from the numbers they do report. Dashes represent numbers we were not able to obtain or estimate. In counting the number of assets, we exclude trivial modifications like re-lighting and re-scaling.} \vspace{-3mm}

\label{tab:compare_related_work}
\end{table*}

\begin{figure}
\includegraphics[width=\linewidth]{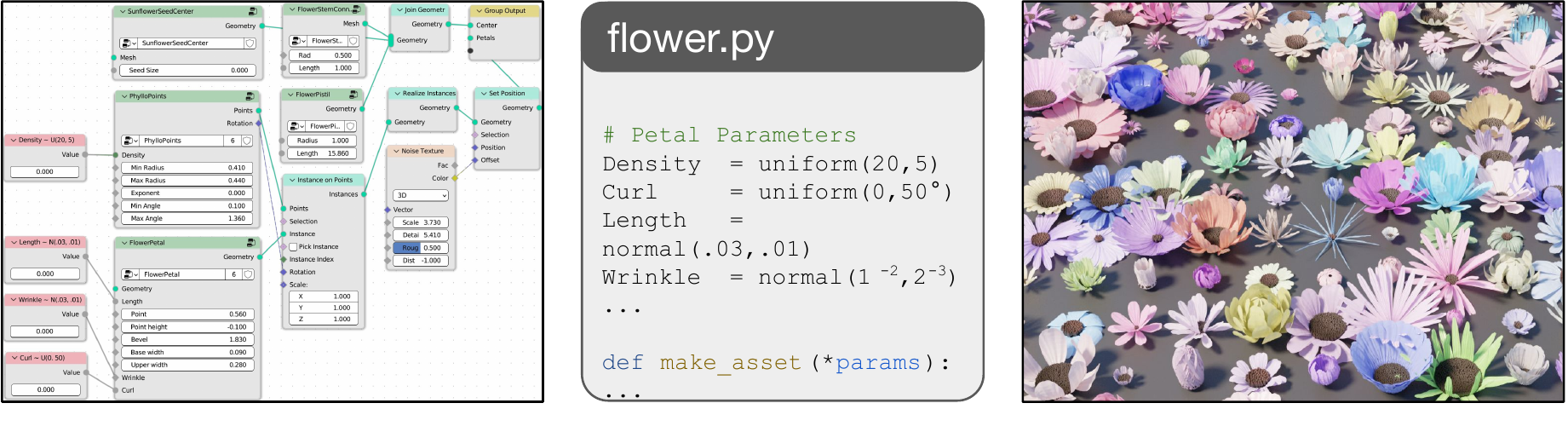}
\caption{Our \emph{Node Transpiler} converts artist-friendly Node-Graphs (left) to procedural code (middle) which produces assets (right).} 
\label{fig:transpiler}
\end{figure}

\begin{figure}
\includegraphics[width=\linewidth]{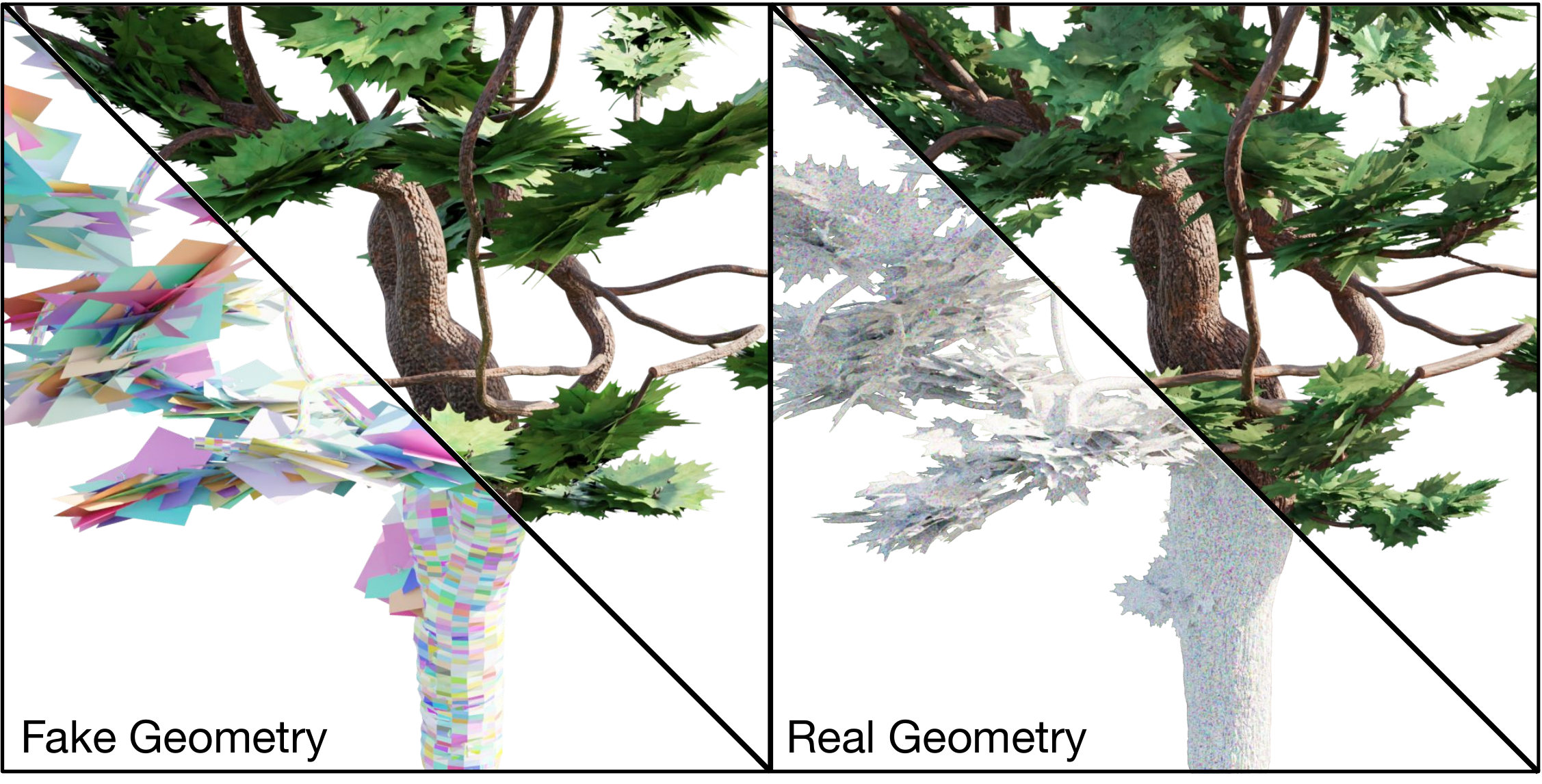}
\caption{Real-time optimized assets (left) often use low res. geometry in conjunction with shading tricks and alpha-masked image textures to give the illusion of geometric detail. \projectname{} assets (right) instead model objects in full geometric detail. Bottom left triangles show a random color per mesh face.\vspace{-0.1in}} 
\label{fig:fakerealgeo}
\end{figure}

\begin{figure}
\includegraphics[width=\linewidth]{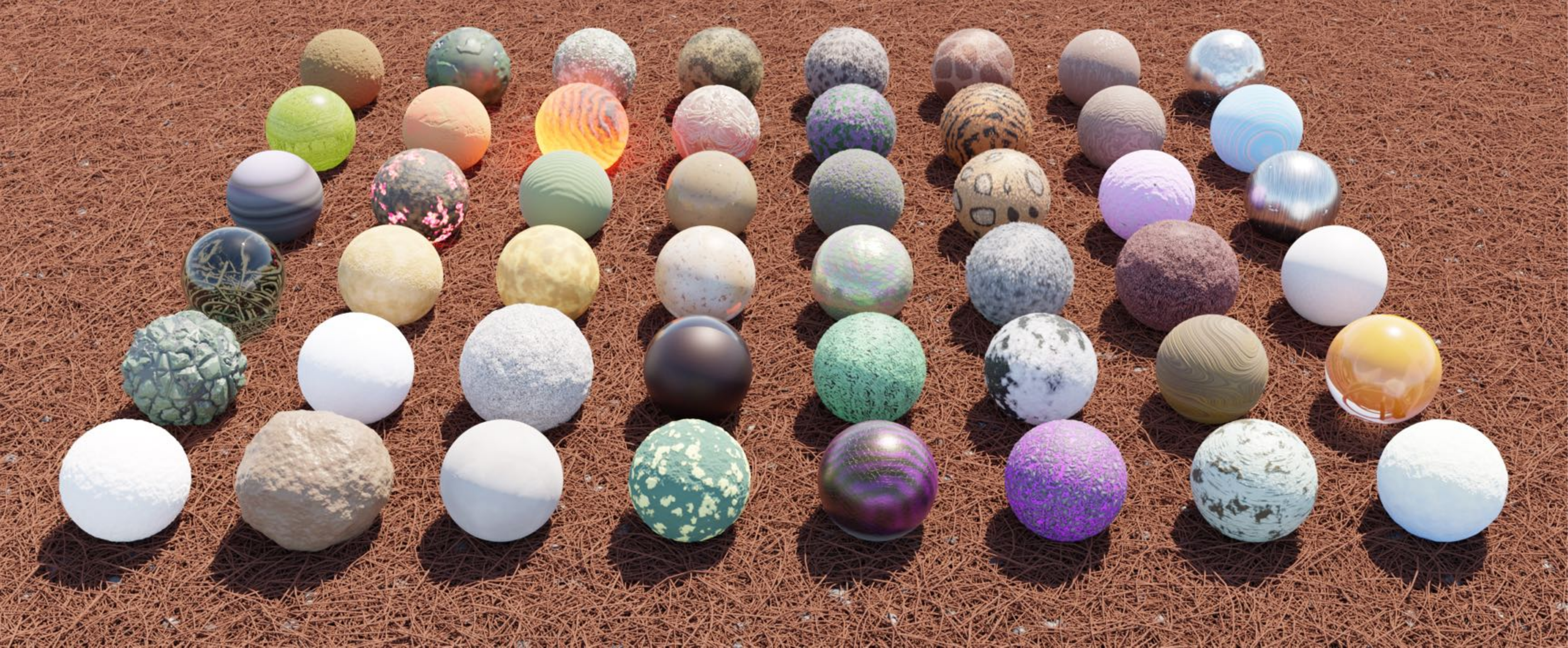}
\caption{Examples of a subset of our material generators. Columns 1-4 are for terrain, 5-7 are for creatures, and 8 is miscellaneous.\vspace{-0.2in}}
\label{fig:materials}
\end{figure}

\begin{figure*}
\includegraphics[width=\linewidth]{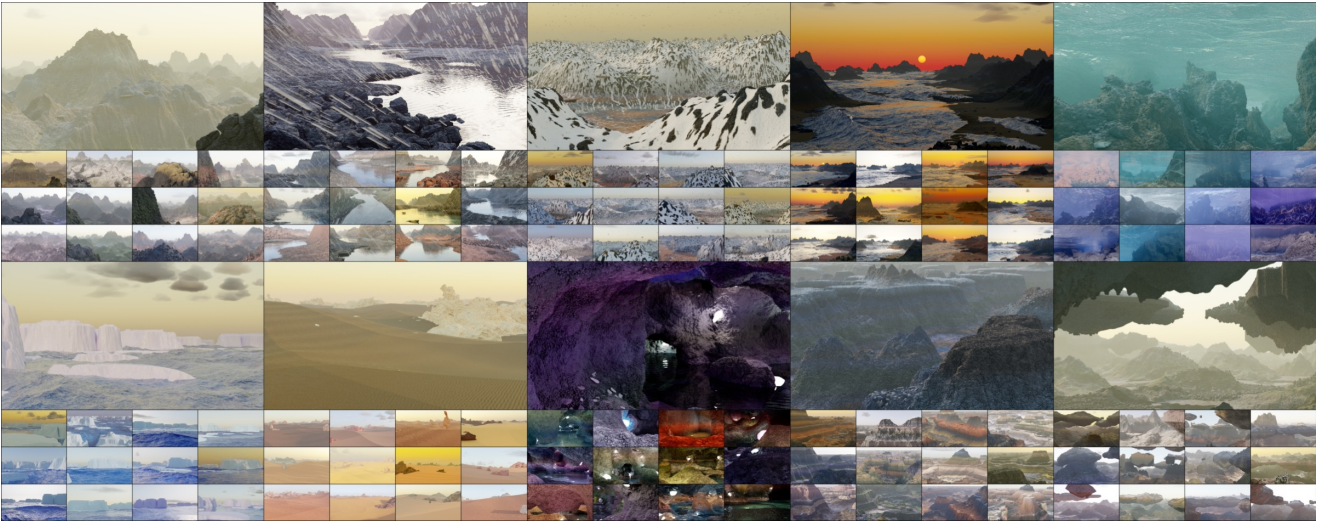}
\caption{Random, \textit{non cherry-picked} terrain-only scenes. We sample 13 images for various natural scene types. From top left to bottom right; Mountains, Rainy river, Snowy mountains, Coastal sunrise, Underwater, Arctic icebergs, Desert, Caves, Canyons and Floating islands. See Appendix for more samples.\vspace{-2mm}
}
\label{fig:terrain}
\end{figure*}

\begin{figure}
\includegraphics[width=\linewidth]{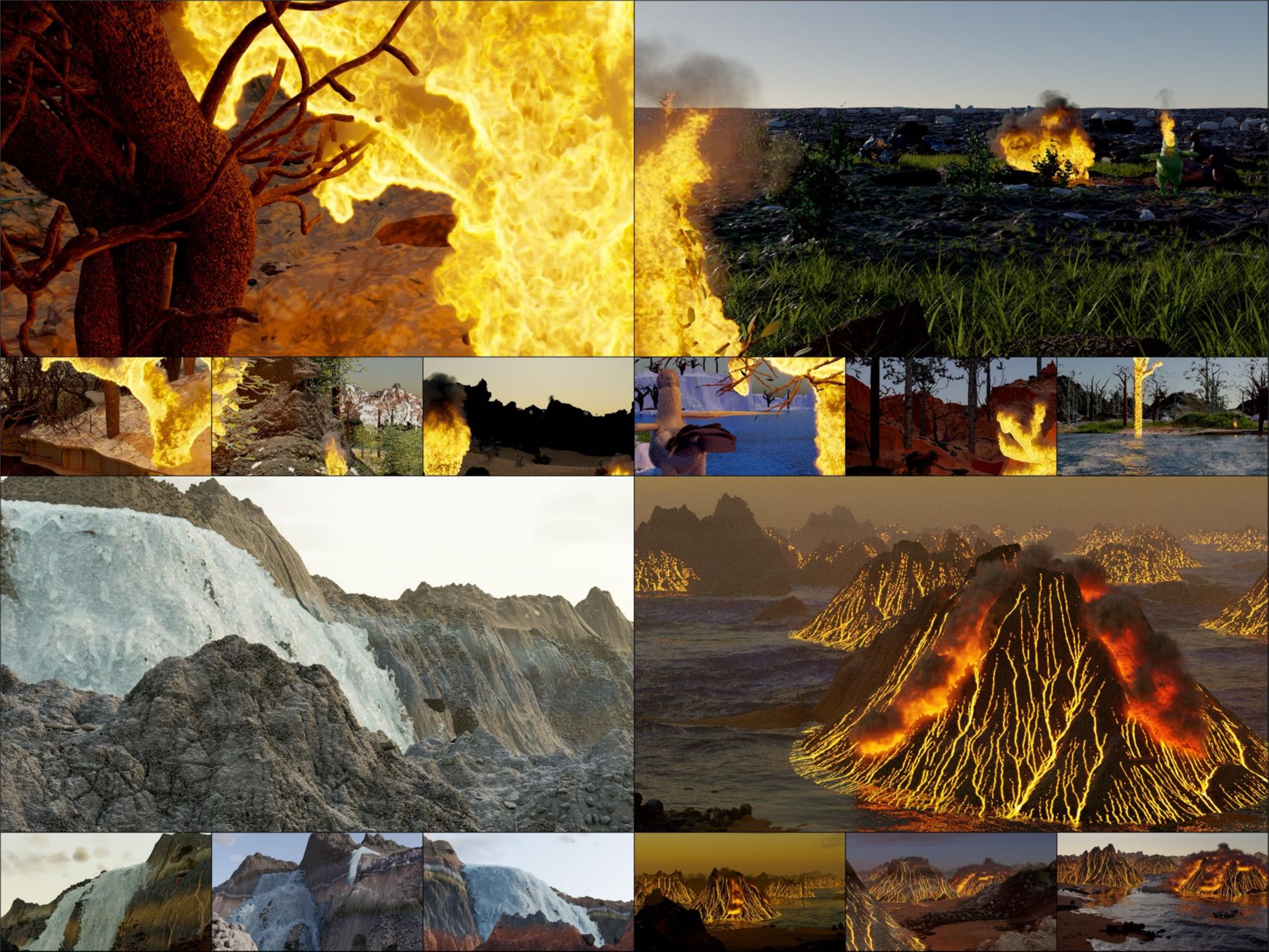}
\caption{Random, \emph{non cherry-picked} images of simulated fire, smoke, waterfalls, and volcano eruptions.\vspace{-4mm}} 
\label{fig:fluidsim}
\end{figure}

\begin{figure}
\includegraphics[width=\linewidth]{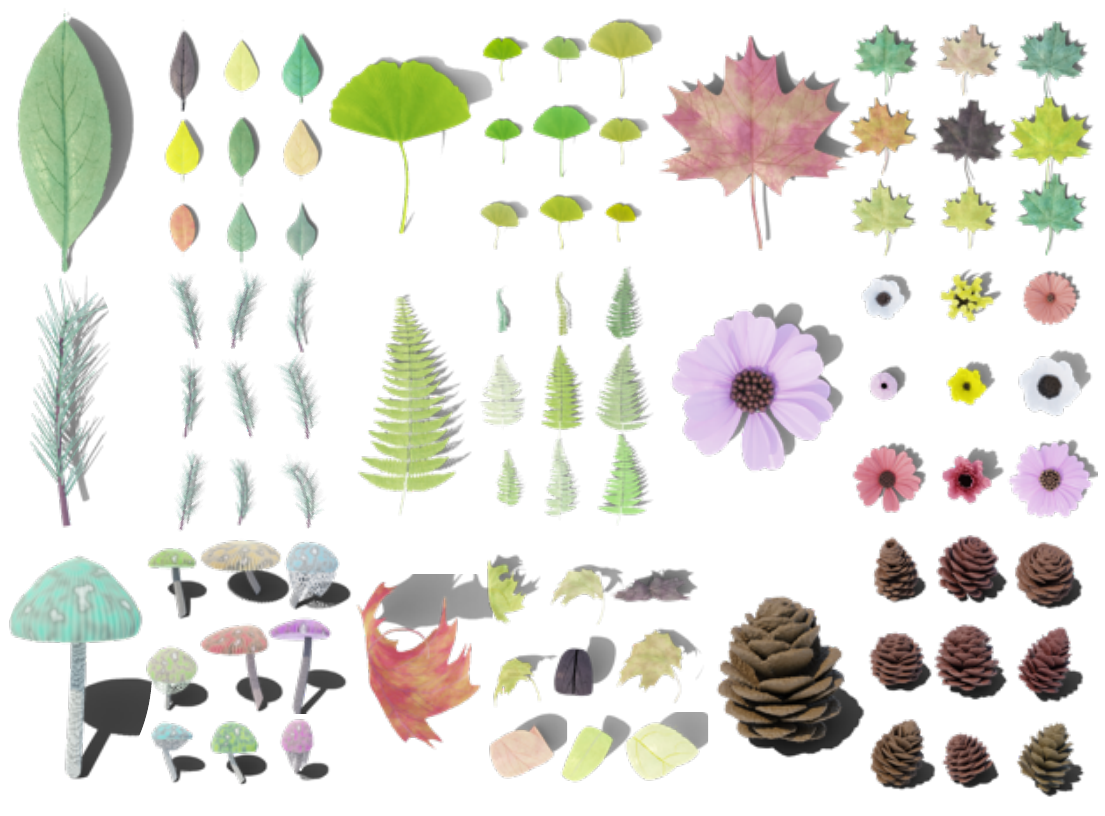}
\caption{Random, \emph{non cherry-picked}  leaves, flowers, mushrooms and pinecones.\vspace{-0.2in}} 
\label{fig:leaves}   
\end{figure}

\begin{figure*}
\includegraphics[width=0.76\linewidth]{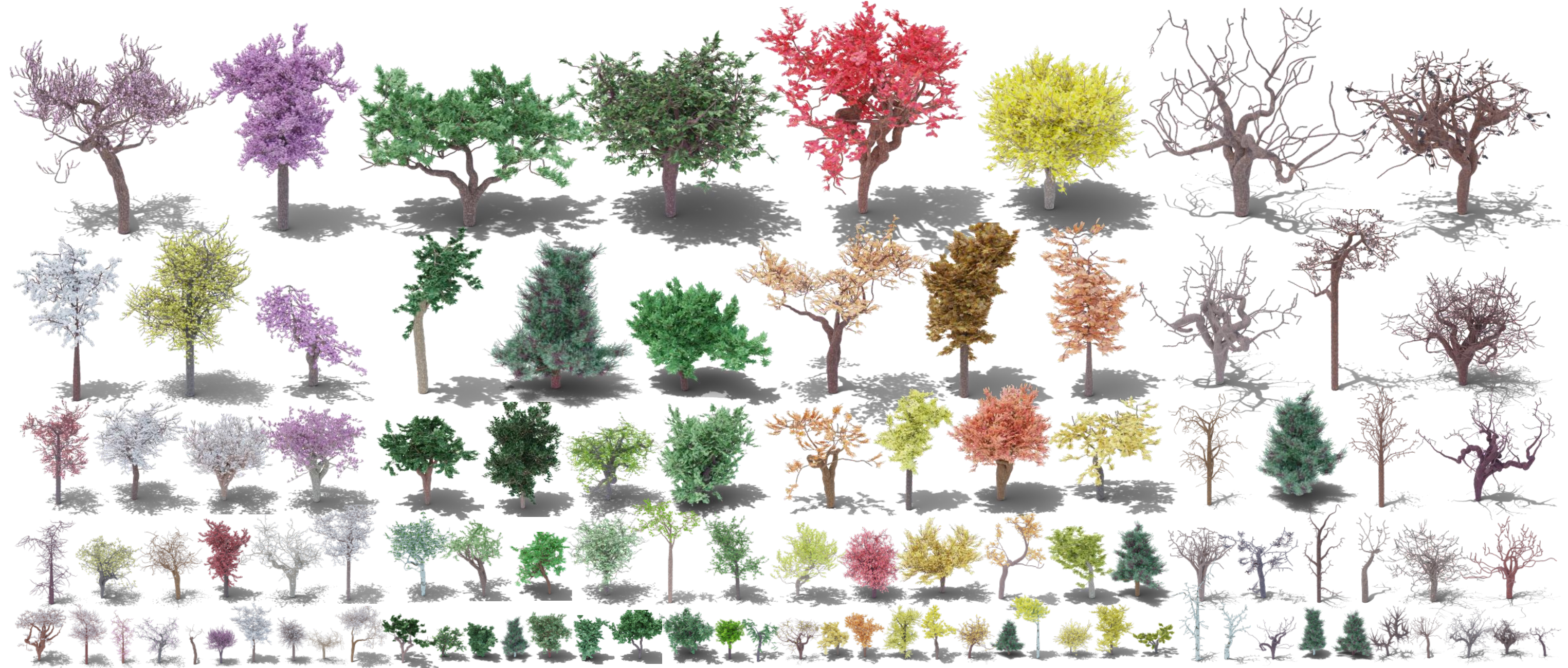}
\hspace{-1mm}
\includegraphics[width=0.237\linewidth]{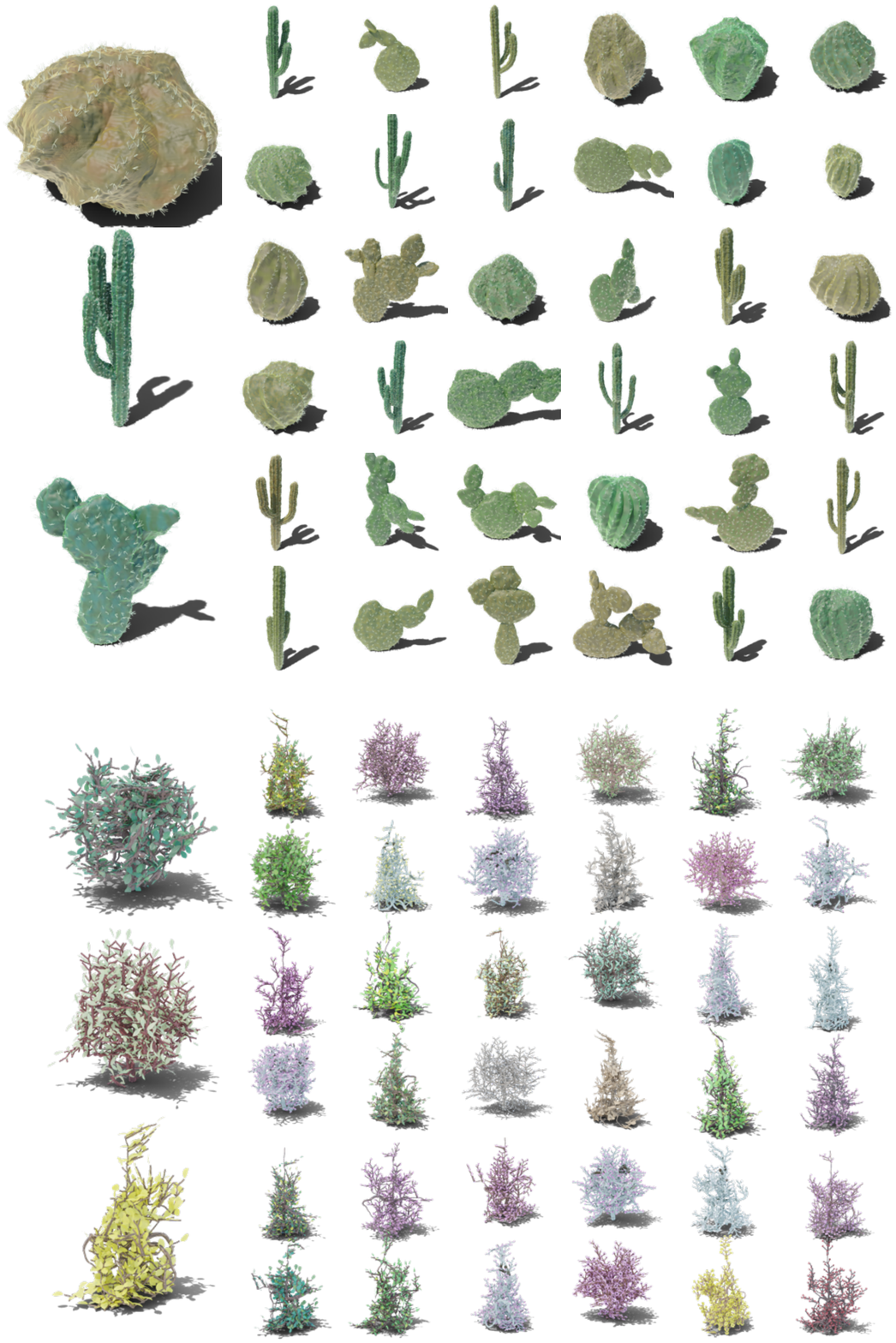}
\caption{Random, \emph{non cherry-picked} procedural trees (left), cacti (top right) and bushes (bottom right).\vspace{-2mm}} 
\label{fig:trees}   
\end{figure*}

\begin{figure}
\includegraphics[width=\linewidth]{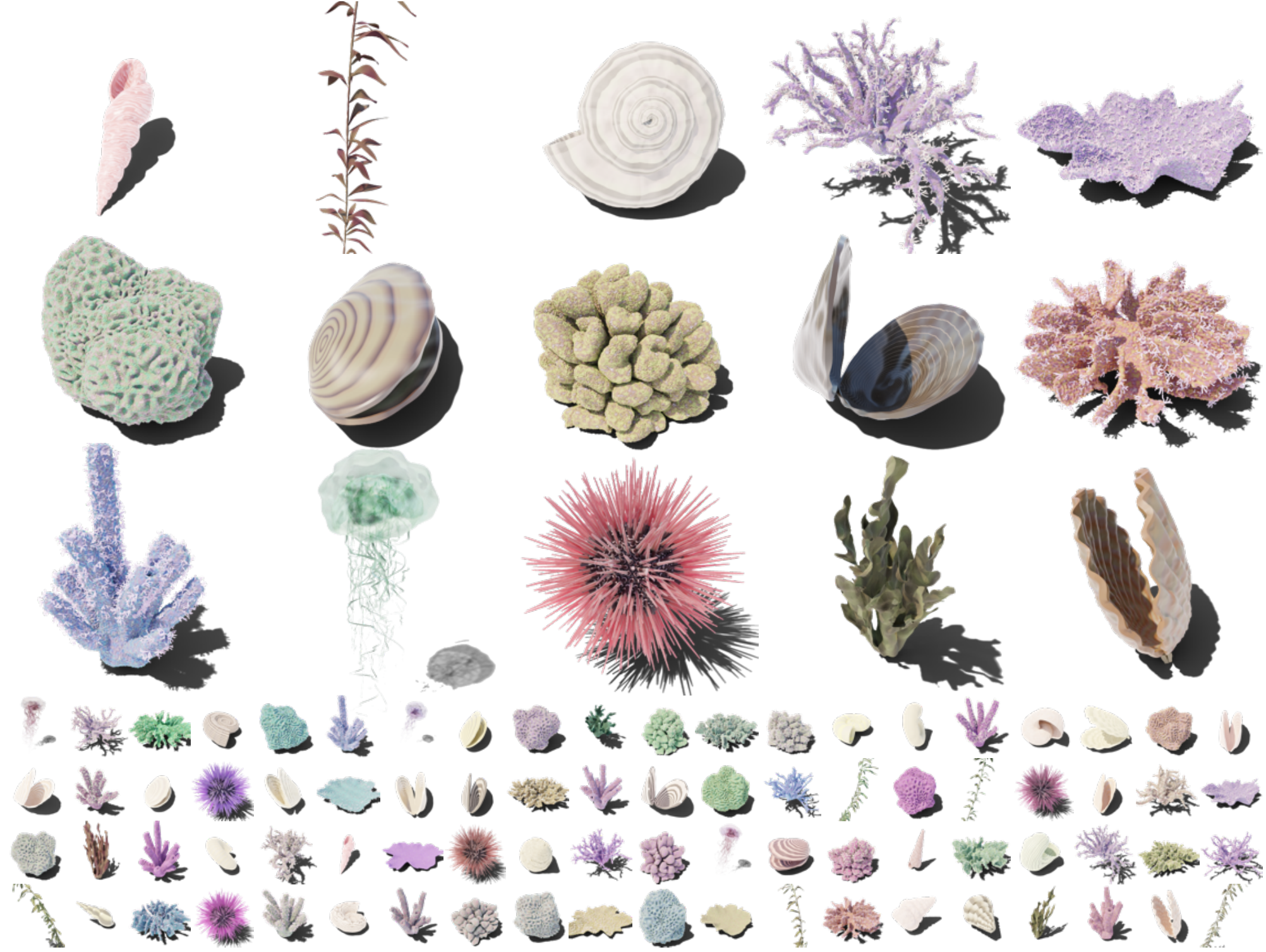}
\vspace{-3mm}
\caption{Random, \emph{non cherry-picked}  underwater objects.\vspace{-2mm}} 
\label{fig:corals_main}
\end{figure}

\begin{figure}
\includegraphics[width=\linewidth]{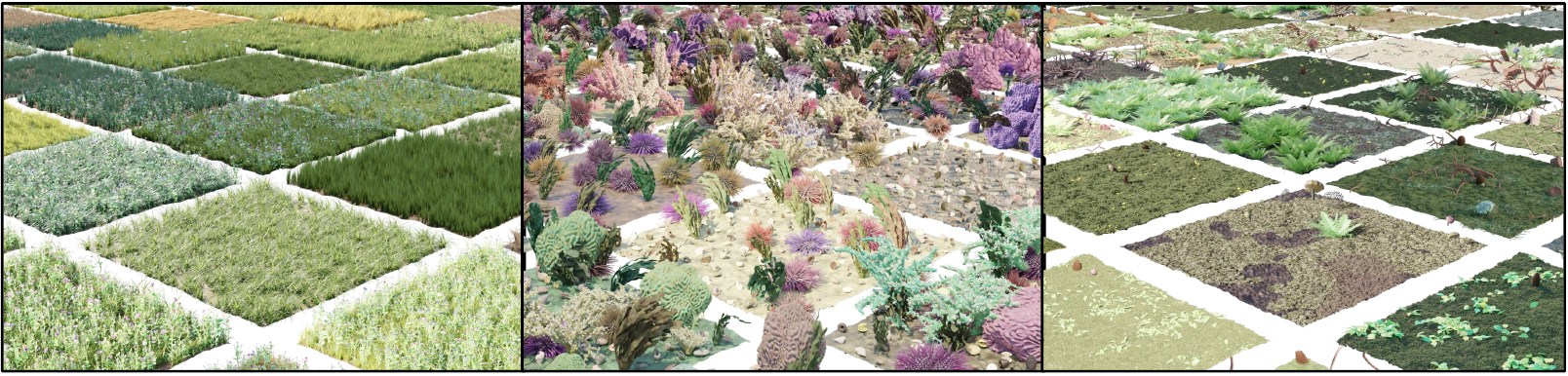}
\caption{Random, \emph{non cherry-picked} surface scatters. Dense coverage with procedural assets turns any surface into a convincing {\it grassland}, {\it sea floor} or {\it forest floor} environment.\vspace{-2mm}} 
\label{fig:scatters}   
\end{figure}

\begin{figure*}
\includegraphics[width=0.28\linewidth]{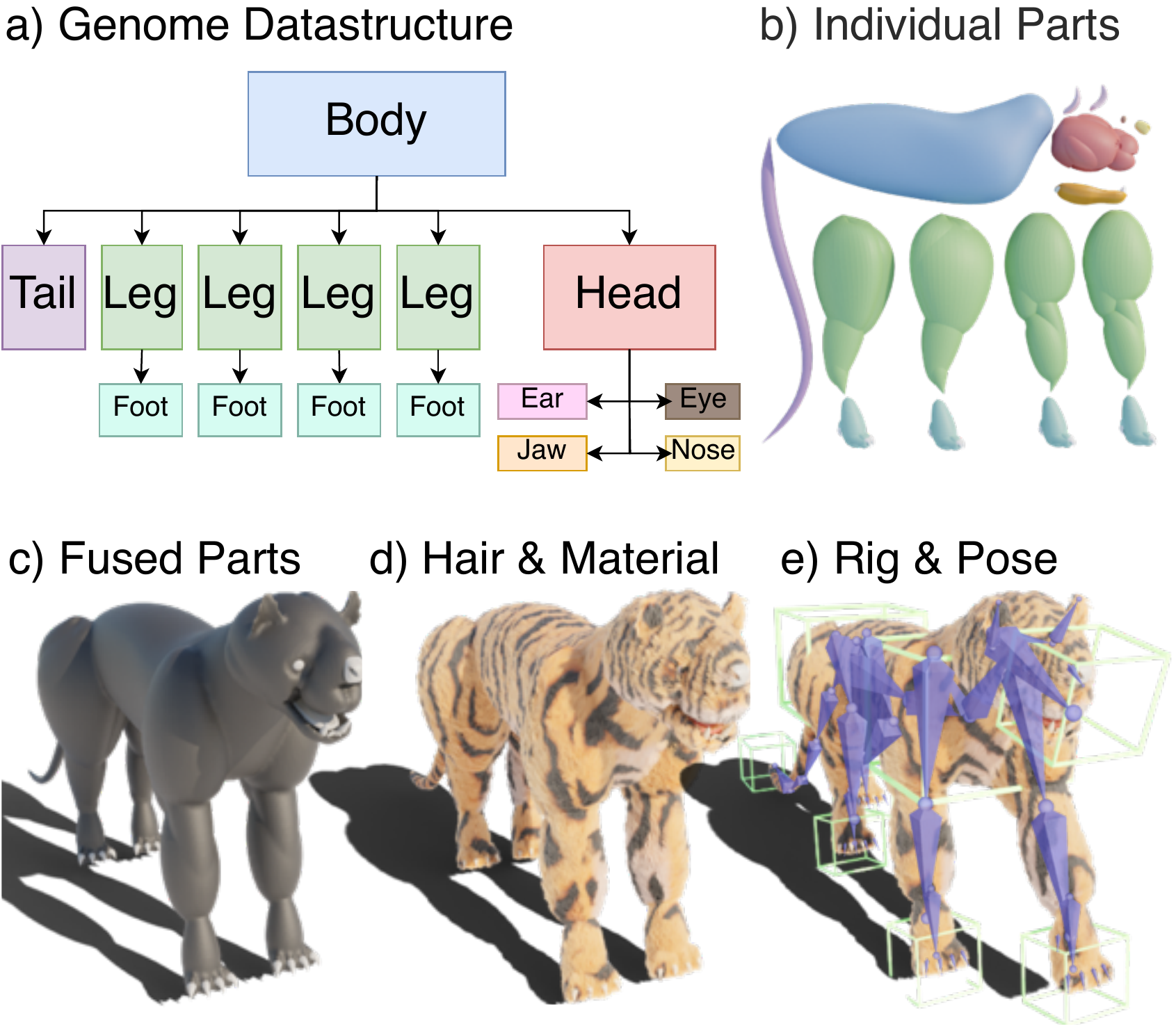}
\includegraphics[width=0.72\linewidth]{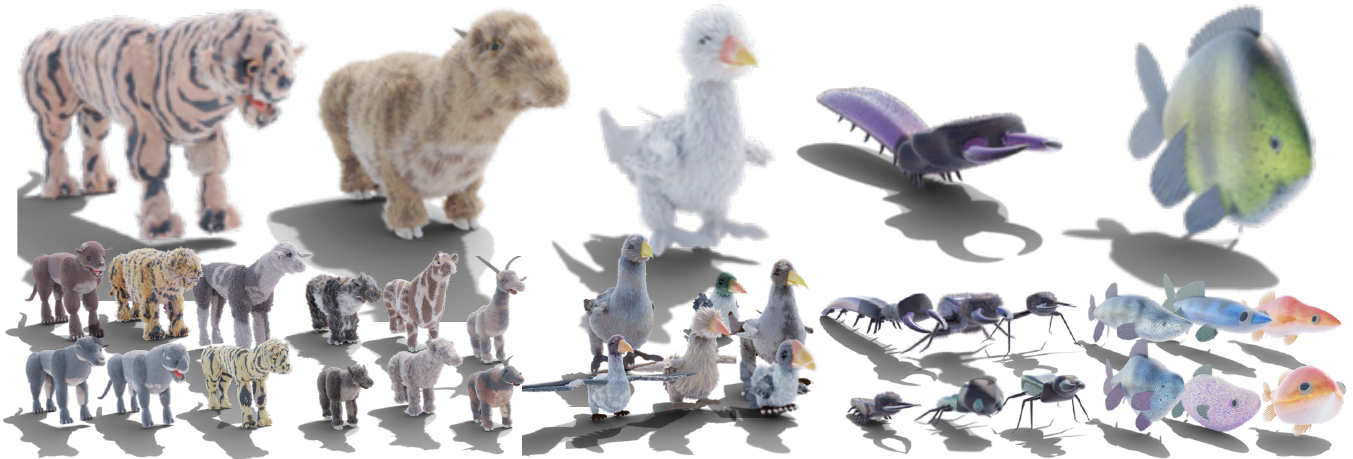}
\caption{Creature Generation. Our system automatically generates genomes (a), parts (b), assembly (c), materials (d) and animation rigs (e). On the right, we show random, \emph{non cherry-picked} samples from our \textit{Carnivore}, \textit{Herbivore}, \textit{Bird}, \textit{Beetle}, and \textit{Fish} generators.\vspace{-2mm}} 
\label{fig:creatures_sample}
\end{figure*}

\begin{figure*}
\includegraphics[width=\linewidth]{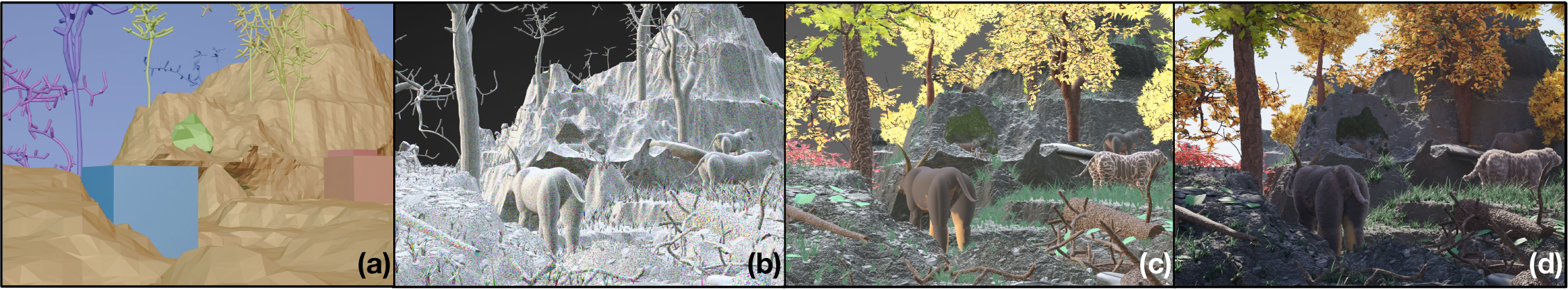}
\caption{Data Generation Pipeline. We procedurally compose a scene layout (a) with random camera poses. We generate all necessary assets (b, showing a color per mesh face), and apply procedural materials and displacement (c). Finally, we render a photo-real image (d).\vspace{-2mm}} 
\label{fig:pipeline}
\end{figure*}

\begin{figure}
\includegraphics[width=\linewidth]{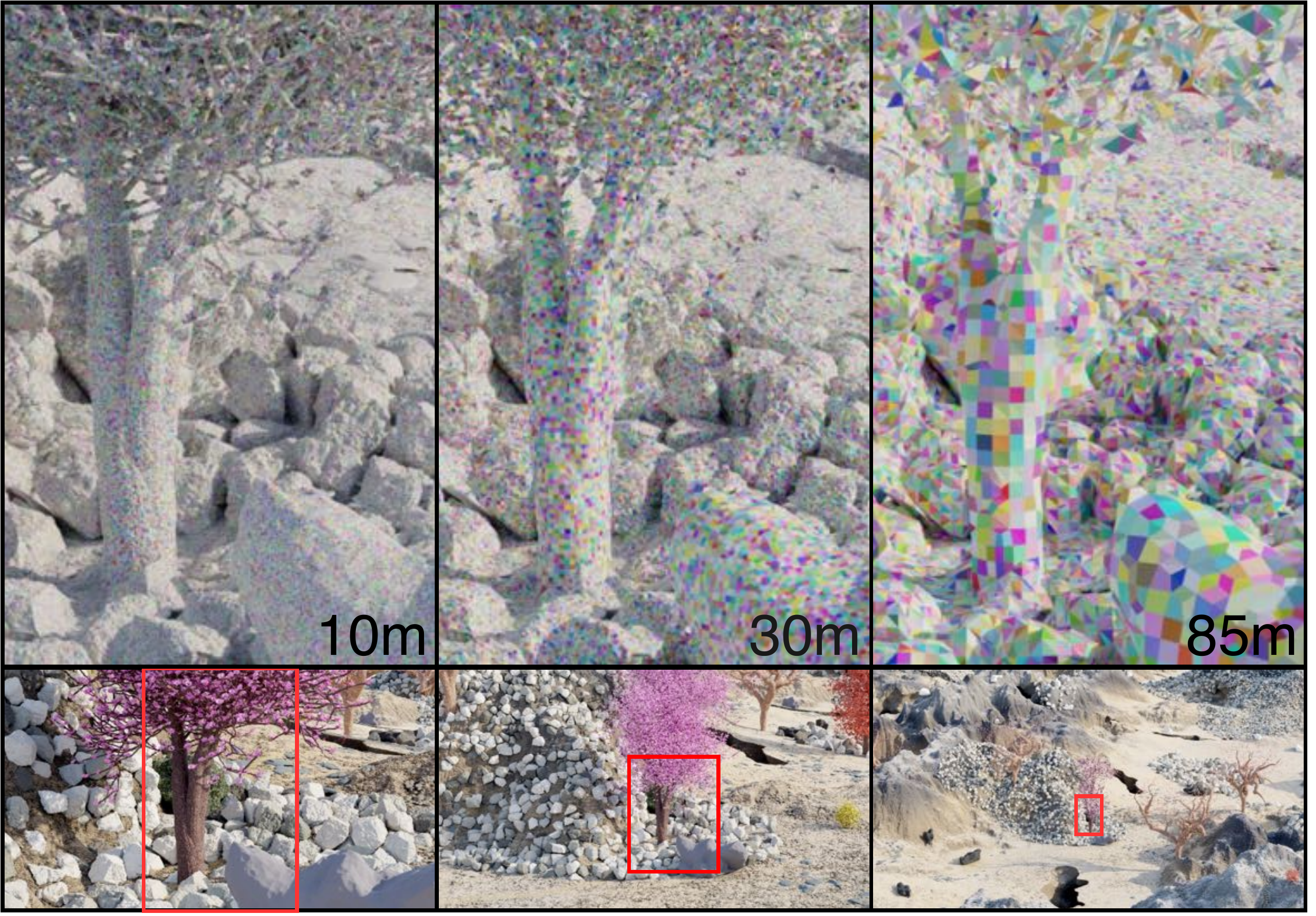}
\caption{Dynamic Resolution Scaling. We show close-up mesh visualizations (top) of the same content for three different camera distances. Despite differing mesh resolution, no changes are visible in the final images (bottom).\vspace{-2mm}} 
\label{fig:dynamic_scaling}
\end{figure}

Synthetic data from computer graphics have been used in computer vision for a wide range of tasks\cite{gta5koltun2016playing, deepfurniture}. We refer the reader to ~\cite{nikolenko2021synthetic} for a comprehensive survey. Below we categorize existing work in terms of application domain, generation method, and accessibility. Tab.~\ref{tab:compare_related_work} provides detailed comparisons. 

\parbf{Application Domain} Synthetic datasets or dataset generators have been developed to cover a variety of domains. The built environment has been covered by the largest amount of existing work~\cite{sintel, wang2020tartanair, li20214dcomplete, crestereo, sceneflow2016, kubric, fallingthings, fabbri2021motsynth} especially indoor scenes~\cite{deepfurniture, robotrix, suncg, cgpbr,zhang2017physically,li2018cgintrinsics,hypersim,li2021openrooms,structured3d,straub2019replica,jiang2018configurable,deitke2022procthor,li2021igibson} and urban scenes~\cite{huang2018deepmvs, gta5koltun2016playing, procsy, metasim2, wrenninge2018synscapes, tsirikoglou2017procedural, fabbri2021motsynth}. A significant source of synthetic data for the built environment comes from simulated platforms for embodied AI, such as AI2-THOR~\cite{ai2thor}, Habitat~\cite{szot2021habitat}, BEHAVIOR~\cite{srivastava2022behavior}, SAPIEN~\cite{xiang2020sapien}, RLBench~\cite{james2020rlbench}, CARLA~\cite{dosovitskiy2017carla}. Some datasets, such as TartanAir~\cite{wang2020tartanair} and Sintel~\cite{sintel}, include a mix of built and natural environments. There also exist datasets such as FlyingThings~\cite{sceneflow2016}, FallingThings~\cite{fallingthings} and Kubric~\cite{kubric} that do not render realistic scenes and instead scatter (mostly artificial) objects against simple backgrounds. Synthetic humans are another important application domain, where high-quality synthetic data have been generated for understanding faces~\cite{wood2021fake}, pose~\cite{andriluka20142d,von2018recovering}, and activity~\cite{soomro2012ucf101, kay2017kinetics}. Some datasets focus on objects, not full scenes, to serve object-centric tasks such as non-rigid reconstruction~\cite{li20214dcomplete}, view synthesis~\cite{mildenhall2021nerf}, and 6D pose~\cite{hodan2020bop}.\looseness=-1

We focus on natural objects and natural scenes, which have had limited coverage in existing work. Even though natural objects do occur in many existing datasets such as urban driving, they are mostly on the periphery and have limited diversity.  

\parbf{Generation Method} Most synthetic datasets are constructed by using a \emph{static} library of 3D assets, either externally sourced or made in house. The downside of a static library is that the synthetic data would be easier to overfit. Procedural generation has been involved in some existing datasets or generators~\cite{deitke2022procthor, kubric, jiang2018configurable, interiornet,he2021semi}, but is limited in scope. Procedural generation is only applied to either object arrangement or a subset of objects, e.g.\@ only buildings and roads but not cars~\cite{procsy,wrenninge2018synscapes}. In contrast, \projectname{} is entirely procedural, from shape to texture, from macro structures to micro details, without relying on any external asset. 

\parbf{Accessibility} A synthetic dataset or generator is most useful if it is maximally accessible, i.e.\@ it provides free access to assets and code with minimum use restrictions. However, few existing works are maximally accessible. Often the rendered images are provided, but underlying 3D assets are unavailable, not free, or have significant use restrictions. Moreover, the code for procedural generation, if any, is often unavailable.\looseness=-1 

\projectname{} is maximally accessible. Its code is available under the BSD license. Anyone can freely use \projectname{} to generate unlimited assets.

%% file: texts/3-method.tex
\section{Method\vspace{-0.05in}}
\label{sec:method}

\begin{table}
\centering
\footnotesize
\begin{tabular}{l|rr} \toprule
 Asset Type & \ Num. Generators & \ Interpretable DOF \\
 \midrule
 Terrain & 26 & 17  \\
 Materials & 50 & 271 \\
 Weather, Fluid & 19 & 61 \\
 Rocks & 4 & 12 \\
 Small Plants & 30 & 258 \\ 
 Trees & 3 & 26 \\
 Creatures & 39 & 315 \\
 Scattering & 11 & 110 \\
 \midrule
 Total & 182 & 1070 \\ 
 \bottomrule
\end{tabular}
\caption{Approximate degrees of freedom, as a proxy of overall diversity. We count only distinct human-understandable parameters with useful ranges of interpolation, with the caveat that this could be an overestimate as not all parameters are fully independent. Some asset classes (e.g terrain) are based on physics simulation and have many more \emph{internal} degrees of freedom not counted here. See Appendix~\ref{thesec:dof} for a full list of named parameters.\vspace{-0.2in}} 
\label{tab:asset_dimensions}
\end{table}

\parbf{Procedural Generation}
Procedural generation refers to the creation of data through generalized rules and simulators. Where an artist might manually create the structure of a single tree by eye, a procedural system creates infinite trees by coding their structure and growth in generality. Developing procedural rules is a form of world modeling using compact mathematical language.

\parbf{Blender Preliminaries} We develop procedural rules primarily using Blender, an open-source 3D modelling software that provides various primitives and utilities. Blender represents scenes as a hierarchy of posed objects. Users modify this representation by transforming objects, adding primitives, and editing meshes. Blender provides import/export for most common 3D file-formats. Finally, all operations in Blender can be automated using its Python API, or by inspecting its open-source code. 

For more complex operations, Blender provides an intuitive node-graph interface. Rather than directly edit shader code to define materials, artists edit \textit{Shader Nodes} to compose primitives into a photo-realistic material. Similarly, \textit{Geometry Nodes} define a mesh using nodes representing operators such as Poisson disk sampling, mesh boolean, extrusion etc. A finalized Geometry Node Tree is a generalized parametric CAD model, which produces a unique 3D object for each combination of its input parameters. These tools are intuitive and widely adopted by 3D artists.

Although we use Blender heavily, not all of our procedural modeling is done using node-graphs; a significant portion of our procedural generation is done outside Blender and only loosely interacts with Blender. 

\parbf{Node Transpiler} As part of \projectname{}, we develop a suite of new tools to speed up our procedural modeling. A notable example is our \textit{Node Transpiler}, which automates the process of converting node-graphs to Python code, as shown in Fig. \ref{fig:transpiler}. The resulting code is more general, and allows us to randomize graph \textit{structure} not just input parameters. This tool makes node-graphs more expressive and allows easy integration with other procedural rules developed directly in Python or C++.
It also allows non-programmers to contribute Python code to \projectname{} by making node-graphs. See Appendix~\ref{thesec:transpiler} for more details.

\input{texts/3.1-asset_generation}

\input{texts/3.2-dataset_generation}

%% file: texts/3.1-asset_generation.tex
\parbf{\textit{Generator} Subsystems}
Infinigen is organized into \textit{generators}, which are probabilistic programs each specialized to produce one subclass of assets (e.g. mountains or fish). Each has a set of high-level parameters (e.g. the overall height of a mountain), which reflect the external degrees of freedom controllable by the user. By default, we randomly sample these parameters according to distributions tuned to mirror the natural world, with no input from the user. However, users can also override any parameter using our Python API to achieve fine grained control of data generation. 

Each  probabilistic program involves many additional internal, low-level degrees of freedom (e.g. the heights of every point on a mountain). Randomizing over both the internal and external degrees of freedom leads to a distribution of assets which we sample from for unlimited generation. Tab.~\ref{tab:asset_dimensions} summarizes the number of human-interpretable degrees of freedom in \projectname{}, with the caveat that the numbers could be an over-estimation because not all parameters are fully independent. Note that it is hard to quantify the internal degrees of freedom, so the external degrees of freedom serve as a lower bound of the total degrees of freedom for our system. \looseness=-1

\parbf{Material Generators} We provide 50 procedural material generators (\fig{fig:materials}). Each is composed of a randomized shader, specifying color and reflectance, and a local geometry generator, which generates corresponding fine geometric details. \looseness=-1

The ability to produce accurate ground-truth geometry is a key feature of our system. This precludes the use of many common graphics techniques such as Bump Mapping and Phong Interpolation \cite{bump_mapping, phong}. Both manipulate face normals to give the illusion of detailed geometric textures, but do so in a way that cannot be represented as a mesh. Similarly, artists often rely on image textures or alpha channel masking to give the illusion of high res. meshes where none exist. All such shortcuts are excluded from our system. See Fig.~\ref{fig:fakerealgeo} for an illustrative example of this distinction. 

\parbf{Terrain Generators} We generate terrain (Fig.~\ref{fig:terrain}) using SDF elements derived from fractal noise~\cite{FastNoiseLite} and simulators ~\cite{FastNoiseLite, Hutton_landlab_2020, barnhart2020landlab, hobley2017creative, SoilMachine}. We evaluate these to a mesh using marching cubes \cite{marchingcubes}. We generate boulders via repeated extrusion, and small stones using Blender's built-in addon. We simulate dynamic fluids (Fig.~\ref{fig:fluidsim}) using FLIP \cite{brackbill_flip_1988}, sun/sky light using the Nishita sky model \cite{nishitalighting}, and weather with Blender's particle system.

\parbf{Plants \& Underwater Object Generators} We model tree growth with random walks and space colonization \cite{runions2007modeling}, resulting in a system with diverse coverage of various trees, bushes and even some cacti (Fig. \ref{fig:trees}). We provide generators for a variety of corals (\fig{fig:corals_main}) using Differential Growth \cite{diff_growth}, Laplacian Growth \cite{Kobayashi1993}, and Reaction-Diffusion \cite{gray1994chemical}. We produce Leaves (Fig. \ref{fig:leaves}), Flowers \cite{jean1983introductory}, Seaweed, Kelp, Mollusks and Jellyfish using geometry node-graphs. 

\parbf{Surface Scatter Generators} Some natural environments are characterized by a dense coverage of smaller objects. To this end, we provide several scatter generators, which combine one or more existing assets in a dense layer (Fig. \ref{fig:scatters}). In the forest floor example, we generate fallen tree logs by procedurally fracturing entire trees from our tree system. 

Due to space constraints, all specific implementation details of the above are available in Appendix~\ref{thesec:imp}. 

\parbf{Creature Generators}
 The genome of each creature is represented as a tree data-structure (Fig. \ref{fig:creatures_sample} a). This reflects the topology of real creatures, whose limbs don't form closed loops. Nodes contain part parameters, and edges specify part attachment. We provide generators for 5 classes of realistic creature genomes, shown in \fig{fig:creatures_sample}. We can also combine creature parts at random, or interpolate similar genomes. See Appendix~\ref{thesec:creature} for details. 

 Each part generator is either a transpiled node-graph, or a non-uniform rational basis spline (NURBS). NURBS parameter-space is high-dimensional, so we randomize NURBS parameters under a factorization inspired by lofting, composed of deviations from a center curve. To tune the random distribution, we modelled 30 example heads and bodies, and ensured that our distribution supports them. 

Our system produces high-quality animation rigs, and optionally simulates realistic surface folding, sagging and motion of creature skin using cloth simulation. For hair, we use the transpiler to automate the process of grooming hairs, as usually performed by human character artists.

%% file: texts/3.2-dataset_generation.tex
\begin{figure*}
\centering
\begin{subfigure}[b]{.195\linewidth}
\includegraphics[width=\linewidth]{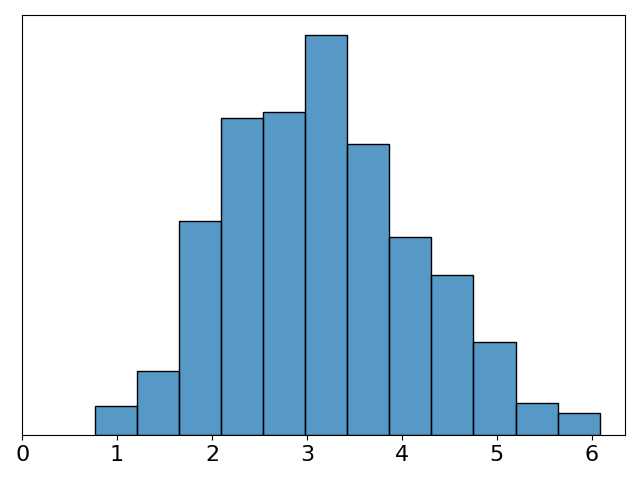}
\caption{Wall Time (Hours)}\label{fig:elapsedhours}
\end{subfigure}
\begin{subfigure}[b]{.195\linewidth}
\includegraphics[width=\linewidth]{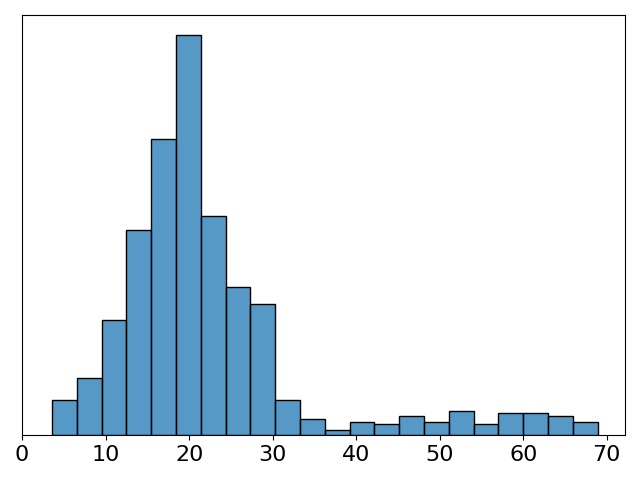}
\caption{Memory (GB)}\label{fig:maxmemory}
\end{subfigure}
\begin{subfigure}[b]{.195\linewidth}
\includegraphics[width=\linewidth]{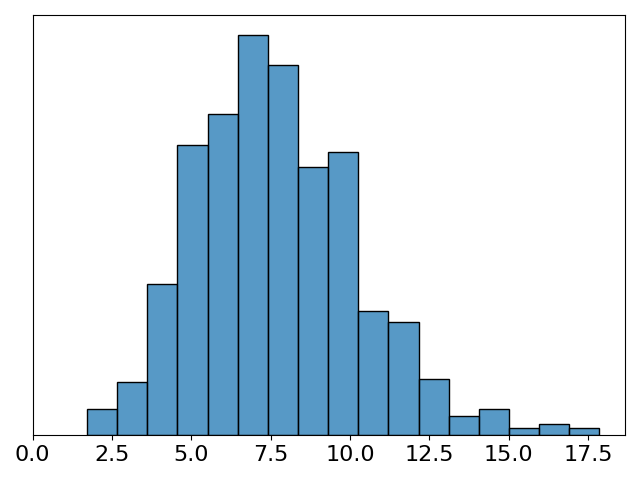}
\caption{CPU Hours}\label{fig:cpuhours}
\end{subfigure}
\begin{subfigure}[b]{.195\linewidth}
\includegraphics[width=\linewidth]{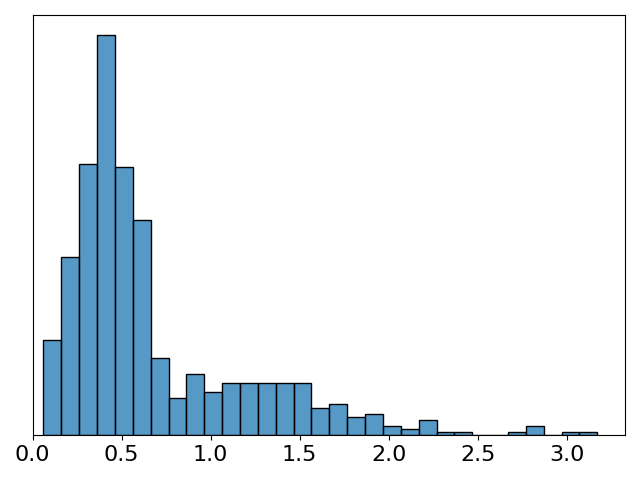}
\caption{GPU Hours}\label{fig:gpuhours}
\end{subfigure}
\begin{subfigure}[b]{.195\linewidth}
\includegraphics[width=\linewidth]{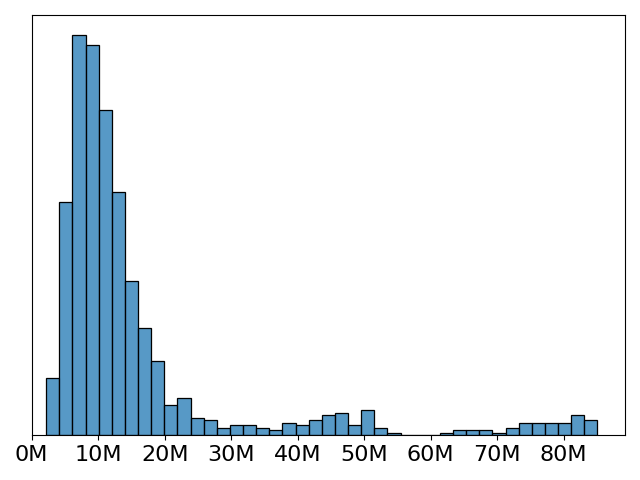}
\caption{\# Triangles per scene}\label{fig:tricounts}
\end{subfigure}
\caption{Resource requirements to create a pair of stereo 1080p images using \projectname{}.}
\label{fig:resourcerequirements}
\end{figure*}

\parbf{Dynamic Resolution Scaling}
With the camera location fixed, we evaluate our procedural assets at precisely the level of detail such that each face is $<1$px in size when rendered. This process is visualized in Fig. \ref{fig:dynamic_scaling}. For most assets, this entails evaluating a parametric curve at the given pixel size, or using Blender's built-in subdivision or re-meshing. For terrain, we perform Marching Cubes on SDF points in \emph{spherical coordinates}. For densely scattered assets (incl. all assets in Fig. \ref{fig:scatters}) we use \textit{instancing} - that is, we generate a fixed number of assets of each type, and reuse them with random transforms within a scene. Even with this effort in optimization, the average complete scene has 16M polygons.

\parbf{Image Rendering \& Ground Truth Extraction}
We render images using Cycles, Blender's physically-based path tracing renderer. We provide code to extract ground truth for common tasks, visualized in Fig. \ref{fig:ground_truth}. 

Cycles individually traces photons of light to accurately simulate diffuse and specular reflection, transparent refraction and volumetric effects. We render at $1920\times 1080$ resolution using $10,000$ random samples per-pixel, which is standard for blender artists and ensures almost no sampling noise in the final image. 

Prior datasets \cite{kubric, hasson19_obman, crestereo, sintel, he2021semi} rely on blender's built-in render-passes to obtain dense ground truth. However, these rendering passes are a byproduct of the rendering pipeline and not intended for training neural networks. Specifically, they are often incorrect due to translucent surfaces, volumetric effects, motion blur, focus blur or sampling noise. See Appendix~\ref{thesec:groundtruthexplanation} for examples of these issues. 

Instead, we provide OpenGL code to extract surface normals, depth and occlusion boundaries from the mesh directly without relying on blender. This solution has many benefits in addition to its accuracy. Users can exclude objects not relevant to their task (e.g. water, clouds, or any other object) independently of whether they are rendered. Many annotations like occlusion boundaries are also plainly not supported by Blender. Finally, our implementation is modular, and we anticipate that users will generate task-specific ground truth not covered above via simple extensions to our codebase. 

\parbf{Runtime}
We benchmark our \projectname{} on 2 \textit{Intel(R) Xeon(R) Silver 4114 @ 2.20GHz} CPUs and 1 NVidia-GPU across 1000 independent trials. The wall time to produce a pair of 1080p images is 3.5 hours. Statistics are shown in Fig. \ref{fig:resourcerequirements}.

%% file: texts/4-experiments.tex
\begin{table}[t]
\centering
\resizebox{\linewidth}{!}{
\tabcolsep=0.08cm
\begin{tabular}{l|cccccccccc|c} \toprule
 Training Dataset & Adirondack & \textbf{Jadeplant} & Motorcycle & Piano & Pipes & Playroom & Playtable & Recycle & Shelves & Vintage & Avg\\
 \midrule
FallingThings~\cite{fallingthings} & 8.3 & 43.3 & \textbf{12.3} & 18.2 & 25.3 & 29.7 & 50.0 & 10.4 & 43.3 & 45.6 & 28.6\\ 
 Sintel-Stereo~\cite{sintel} & 35.7 & 62.9 & 31.1 & 24.1 & 31.9 & 41.7 & 60.1 & 30.8 & 55.8 & 76.1 & 45.0\\ 
 HR-VS~\cite{hrvs} & 43.5 & 43.2 & 17.0 & 29.6 & 32.1 & 34.6 & 68.4 & 24.7 & 57.4 & 34.9 & 38.5 \\ 
 Li et al.~\cite{crestereo} & 23.9 & 80.2 & 40.7 & 32.0 & 40.3 & 49.1 & 67.5 & 36.6 & 51.7 & 42.3 & 46.4\\ 
 SceneFlow~\cite{sceneflow2016} & \textbf{7.4} & 41.3 & 14.9 & 16.2 & 33.3 & \textbf{18.8} & \textbf{38.6} & \textbf{10.2} & \textbf{39.1} & 29.9 & \textbf{25.0}\\ 
 TartanAir~\cite{wang2020tartanair} & 15.5 & 45.1 & 18.1 & \textbf{12.9} & 28.4 & 25.6 & 51.0 & 20.9 & 49.1 & \textbf{28.2} & 29.5\\ 
 InStereo2K~\cite{bao2020instereo2k} & 17.1 & 59.7 & 21.3 & 23.8 & 35.8 & 33.9 & \textbf{36.4} & 20.0 & \textbf{33.4} & 44.1 & 32.5\\ 
 Ours (\projectname{} 30K) & \textbf{7.4} & \textbf{35.2} & 15.2 & 20.7 & \textbf{24.7} & 29.3 & 50.0 & 12.6 & 55.1 & 46.9 & 29.7\\ 
 \bottomrule
\end{tabular}
}
\caption{Bad 3.0 (\%) $\downarrow$ error on the Middlebury~\cite{middlebury} validation set. \projectname{} generalizes well to natural objects (e.g. Jadeplant). However, natural objects contain very few planar or textureless surfaces; models trained exclusively on natural objects generalize less well on Middlebury's indoor scenes.\vspace{-2mm}}
\label{tab:middlebury_train}
\end{table}

\begin{figure}[t]
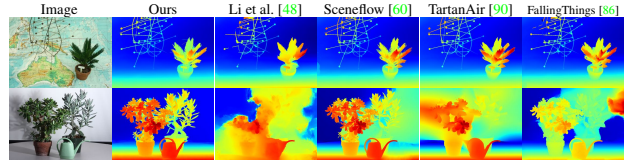

  \begin{center}
  \resizebox{\linewidth}{!}{
    \setlength{\tabcolsep}{0.7pt}
    \begin{tabular}{ccccccc}
      {\Large Image} &
      {\Large Ours} &
      {\Large Li et al.~\cite{crestereo}} &
      {\Large Sceneflow~\cite{sceneflow2016}} & 
      {\Large TartanAir~\cite{wang2020tartanair}} &
      {\large FallingThings~\cite{fallingthings}} & \\
    \galleryRowCompare{figures/stereo_qual_results/row12}{input.pdf}{pg_v4_large.pdf}{cre_stereo_large.pdf}{sceneflow_v4_large.pdf}{tartan_air_large.pdf}{falling_things_large.pdf}
    \galleryRowCompare{figures/stereo_qual_results/row13}{input.pdf}{pg_v4_large.pdf}{cre_stereo_large.pdf}{sceneflow_v4_large.pdf}{tartan_air_large.pdf}{falling_things_large.pdf}
    \end{tabular}
  }
  \end{center}
  \vspace{-1.5em}
  \caption{Qualitative results on the \emph{Plant} and \emph{Australia}  Middlebury~\cite{middlebury} test images. RAFT-Stereo trained using \projectname{} generalizes well to images with natural objects.\vspace{-2mm}}
  \label{fig:middlebury_plants}
\end{figure}

\section{Experiments}
\label{sec:experiments_sec}
To evaluate \projectname{}, we produced 30K image pairs with ground truth for rectified stereo matching. We train RAFT-Stereo~\cite{lipson2021raft} on these images from scratch and compare results on the Middlebury validation (Tab. \ref{tab:middlebury_train}) and test sets (Fig. \ref{fig:middlebury_plants}). See Appendix for qualitative results on in-the-wild natural photographs.

%% file: texts/6-contributions.tex
\vspace{-1mm}
\section{Contributions \& Acknowledgements}\vspace{0mm}
Alexander Raistrick, Lahav Lipson and Zeyu Ma contributed equally, and are ordered alphabetically by first name; each has the right to list their name first in their CV. Alexander Raistrick performed team coordination, and developed the creature system, transpiler and scene composition. Lahav Lipson trained models and implemented dense annotations and rendering. Zeyu Ma developed the terrain system and camera selection. Lingjie Mei created coral, sea invertebrates, small plants, boulders \& moss. Mingzhe Wang developed materials for creatures, plants and terrain. Yiming Zuo developed trees and leaves. Karhan Kayan created liquids and fire. Hongyu Wen \& Yihan Wang developed creature parts and materials. Beining Han created ferns and small plants. Alejandro Newell designed the tree \& bush system. Hei Law created weather and clouds. Ankit Goyal developed terrain materials. Kaiyu Yang developed an initial prototype with randomized shapes. Jia Deng conceptualized the project, led the team, and set the directions.

We thank Zachary Teed for helping with the early prototypes.
This work was partially supported by the Office of Naval Research under Grant N00014-20-1-2634 and the National Science Foundation under Award IIS-1942981.

%% file: supp.tex
\renewcommand{\thefigure}{\Alph{figure}}
\setcounter{figure}{0}

\renewcommand{\thetable}{\Alph{table}}
\setcounter{table}{0}
\setcounter{tocdepth}{3}
\section*{Appendix}


\input{texts/13-extended-figures.tex}
\input{texts/11-experiments.tex}
\input{texts/16-Dataset-Generation}
\input{texts/12-degrees-of-freedom.tex}

\input{texts/17-transpiler-details.tex}
\input{texts/15-composition-in-detail.tex}

\input{texts/14-assets-in-detail.tex}

\clearpage
\input{texts/18-dof-tables.tex}
\clearpage



%% file: texts/13-extended-figures.tex
\section {Figures Extended}
\label{thesec:figuresextended}
\begin{figure*}
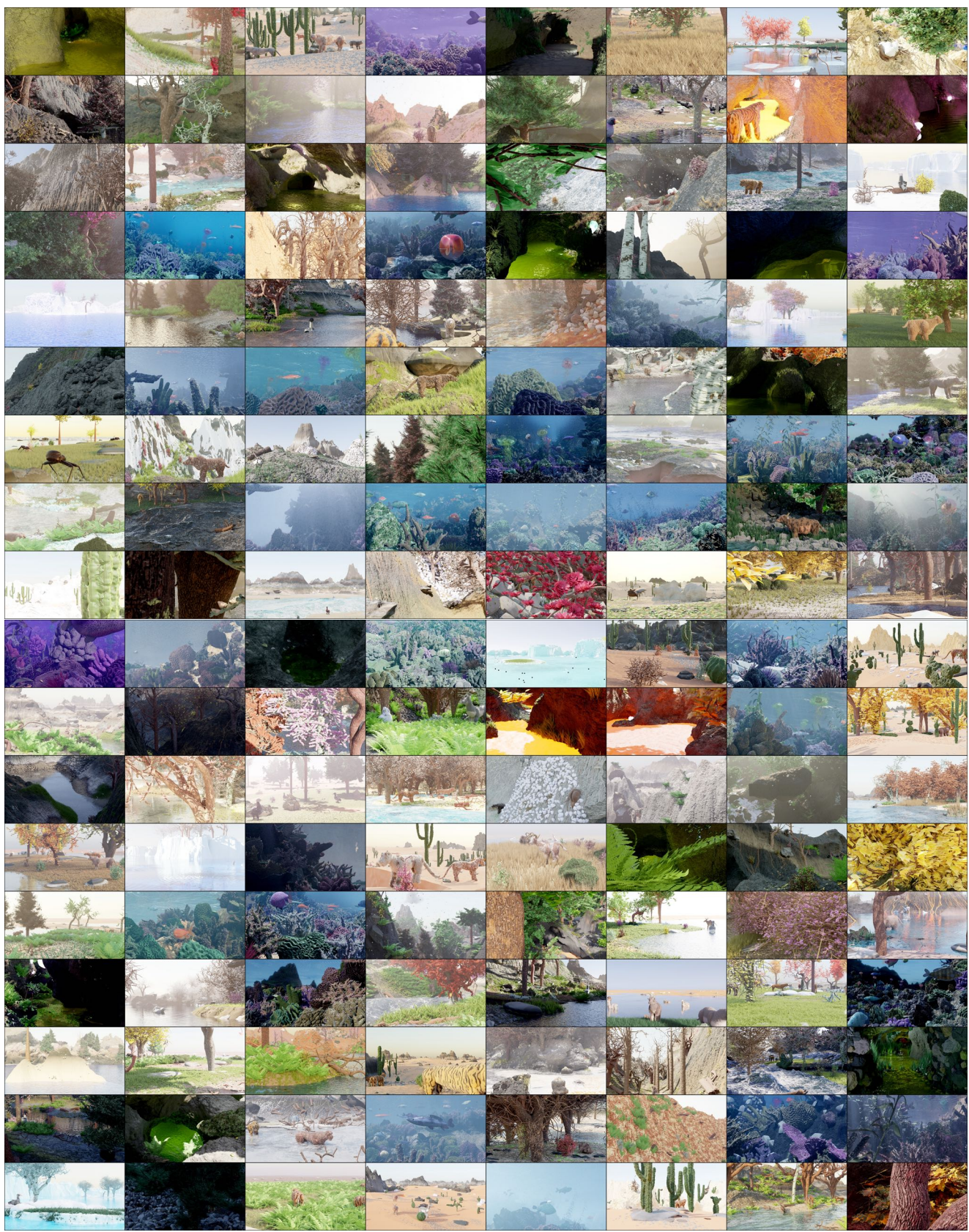

\includegraphics[width=\linewidth]{figures/extended_fullrandom_sample/1_1.pdf}
\includegraphics[width=\linewidth]{figures/extended_fullrandom_sample/1_2.pdf}
\caption{576 randomly generated, non-cherry-picked images produced by our system (Part 1 of 4). Images are compressed due to space constraints - please see \href{https://infinigen.org}{infinigen.org}}
\label{thefig:rand_sample_1}
\end{figure*}

First, we provide a large sample of random, non-cherry-picked RGB images from our dataset generator ( Figs. \ref{thefig:rand_sample_1}, \ref{thefig:rand_sample_2}, \ref{thefig:rand_sample_3}, \ref{thefig:rand_sample_4}).  Both this sample and Fig. 1 in the main paper are randomly selected, however unlike in Fig. 1, here we do not group the images by scene type. We limit the sample to 576 JPEG images of resolution 960x540 only due to space constraints - in practice we can generate infinitely many such images as 1080P PNGs, with a full suite of accompanying ground truth data.

Second, we show an extended random sample of our terrain system (Fig. \ref{thefig:terrain_new0} \ref{thefig:terrain_new1}), grouped by scene type as in Fig. 6 of the main paper.

Please visit \href{https://infinigen.org}{infinigen.org} for videos, code, and extended hi-res. random samples.

\begin{figure*}
\includegraphics[width=\linewidth]{figures/extended_fullrandom_sample/2_1.pdf}
\includegraphics[width=\linewidth]{figures/extended_fullrandom_sample/2_2.pdf}
\caption{576 randomly generated, non-cherry-picked images produced by our system (Part 2 of 4).   Images are compressed due to space constraints - please see \href{https://infinigen.org}{infinigen.org}}
\label{thefig:rand_sample_2}
\end{figure*}

\begin{figure*}
\includegraphics[width=\linewidth]{figures/extended_fullrandom_sample/3_1.pdf}
\includegraphics[width=\linewidth]{figures/extended_fullrandom_sample/3_2.pdf}
\caption{576 randomly generated, non-cherry-picked images produced by our system (Part 3 of 4). Images are compressed due to space constraints - please see \href{https://infinigen.org}{infinigen.org}}
\label{thefig:rand_sample_3}
\end{figure*}

\begin{figure*}
\includegraphics[width=\linewidth]{figures/extended_fullrandom_sample/4_1.pdf}
\includegraphics[width=\linewidth]{figures/extended_fullrandom_sample/4_2.pdf}
\caption{576 randomly generated, non-cherry-picked images produced by our system (Part 4 of 4). Images are compressed due to space constraints - please see \href{https://infinigen.org}{infinigen.org}}
\label{thefig:rand_sample_4}
\end{figure*}

  \begin{figure*}[t]
    \centering
    \includegraphics[width=0.93\linewidth]{figures/supp_terrain_new0.pdf}
    
  \caption{144 randomly generated, non-cherry-picked images of terrain produced by our system (Part 1 of 2). Images are compressed due to space constraints - please see \href{https://infinigen.org}{infinigen.org}}

  \label{thefig:terrain_new0}
  
\end{figure*}

  \begin{figure*}[t]
    \centering
    \includegraphics[width=0.93\linewidth]{figures/supp_terrain_new1.pdf}
    
  \caption{144 randomly generated, non-cherry-picked images of terrain produced by our system (Part 2 of 2). Images are compressed due to space constraints - please see \href{https://infinigen.org}{infinigen.org}}

  \label{thefig:terrain_new1}
  
\end{figure*}

%% file: texts/11-experiments.tex
\section {Experiments}
\label{thesec:supp_experiments}

\begin{figure*}[t]
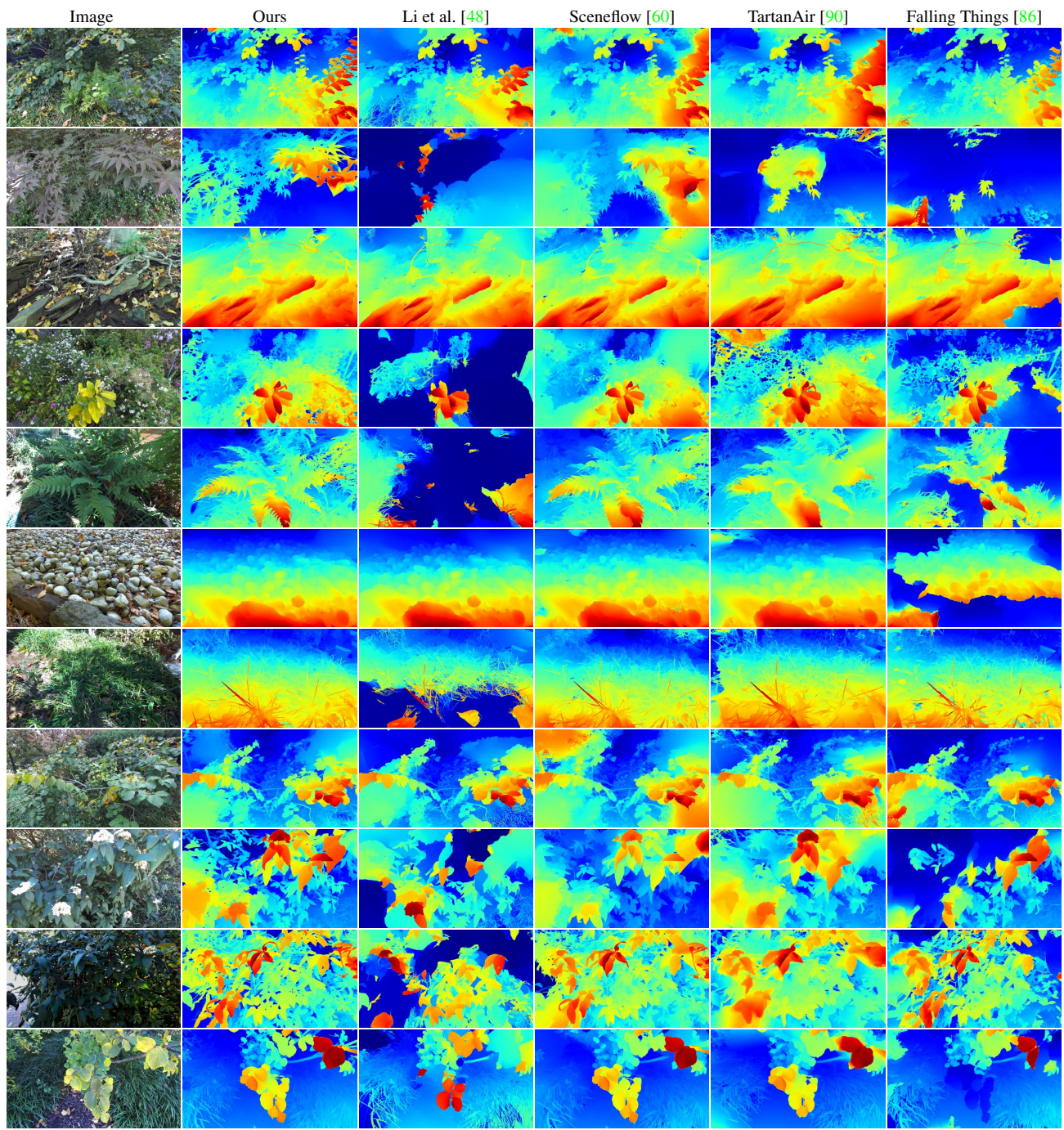
 
  \begin{center}
  \resizebox{\linewidth}{!}{
    \setlength{\tabcolsep}{0.7pt}
    \begin{tabular}{ccccccc}
      Image &
      Ours &
      Li et al.~\cite{crestereo} &
      Sceneflow~\cite{sceneflow2016} &
      TartanAir~\cite{wang2020tartanair} & 
      Falling Things~\cite{fallingthings} \\
    \galleryRowCompare{figures/stereo_qual_results/row1}{input.pdf}{pg_v4_large.pdf}{cre_stereo_large.pdf}{sceneflow_v4_large.pdf}{tartan_air_large.pdf}{falling_things_large.pdf}
    \galleryRowCompare{figures/stereo_qual_results/row2}{input.pdf}{pg_v4_large.pdf}{cre_stereo_large.pdf}{sceneflow_v4_large.pdf}{tartan_air_large.pdf}{falling_things_large.pdf}
    \galleryRowCompare{figures/stereo_qual_results/row8}{input.pdf}{pg_v4_large.pdf}{cre_stereo_large.pdf}{sceneflow_v4_large.pdf}{tartan_air_large.pdf}{falling_things_large.pdf}
    \galleryRowCompare{figures/stereo_qual_results/row4}{input.pdf}{pg_v4_large.pdf}{cre_stereo_large.pdf}{sceneflow_v4_large.pdf}{tartan_air_large.pdf}{falling_things_large.pdf}
    \galleryRowCompare{figures/stereo_qual_results/row5}{input.pdf}{pg_v4_large.pdf}{cre_stereo_large.pdf}{sceneflow_v4_large.pdf}{tartan_air_large.pdf}{falling_things_large.pdf}
    \galleryRowCompare{figures/stereo_qual_results/row6}{input.pdf}{pg_v4_large.pdf}{cre_stereo_large.pdf}{sceneflow_v4_large.pdf}{tartan_air_large.pdf}{falling_things_large.pdf}
    \galleryRowCompare{figures/stereo_qual_results/row7}{input.pdf}{pg_v4_large.pdf}{cre_stereo_large.pdf}{sceneflow_v4_large.pdf}{tartan_air_large.pdf}{falling_things_large.pdf}
    \galleryRowCompare{figures/stereo_qual_results/row3}{input.pdf}{pg_v4_large.pdf}{cre_stereo_large.pdf}{sceneflow_v4_large.pdf}{tartan_air_large.pdf}{falling_things_large.pdf}
    \galleryRowCompare{figures/stereo_qual_results/row9}{input.pdf}{pg_v4_large.pdf}{cre_stereo_large.pdf}{sceneflow_v4_large.pdf}{tartan_air_large.pdf}{falling_things_large.pdf}
    \galleryRowCompare{figures/stereo_qual_results/row10}{input.pdf}{pg_v4_large.pdf}{cre_stereo_large.pdf}{sceneflow_v4_large.pdf}{tartan_air_large.pdf}{falling_things_large.pdf}
    \galleryRowCompare{figures/stereo_qual_results/row11}{input.pdf}{pg_v4_large.pdf}{cre_stereo_large.pdf}{sceneflow_v4_large.pdf}{tartan_air_large.pdf}{falling_things_large.pdf}
    \end{tabular}
  }
  \end{center}
  \vspace{-1.5em}
  \caption{Qualitative results on natural stereo photographs. Rectified input images are captured at $2208\times 2484$ resolution using a calibrated ZED 2 stereo camera~\cite{zed2}. Our data generator helps RAFT-Stereo generalize well to real images of natural objects.
  }
  \label{thefig:wildresults}
\end{figure*}

\begin{figure*}[t]
  \begin{center}
  \resizebox{\linewidth}{!}{
    \setlength{\tabcolsep}{0.7pt}
    \begin{tabular}{ccccccc}
      Image &
      Ours &
      Li et al.~\cite{crestereo} &
      Sceneflow~\cite{sceneflow2016} & 
      TartanAir~\cite{wang2020tartanair} &
      FallingThings~\cite{fallingthings} & \\
    \galleryRowCompare{figures/stereo_qual_results/row12}{input.pdf}{pg_v4_large.pdf}{cre_stereo_large.pdf}{sceneflow_v4_large.pdf}{tartan_air_large.pdf}{falling_things_large.pdf}
    \galleryRowCompare{figures/stereo_qual_results/row13}{input.pdf}{pg_v4_large.pdf}{cre_stereo_large.pdf}{sceneflow_v4_large.pdf}{tartan_air_large.pdf}{falling_things_large.pdf}
    \end{tabular}
  }
  \end{center}
  \vspace{-1.5em}
  \caption{Qualitative results on the \emph{Plant} and \emph{Australia}  Middlebury~\cite{middlebury} test images. RAFT-Stereo trained using \projectname{} generalizes well to images with natural objects.}
  \label{thefig:middlebury_plants}
\end{figure*}

To validate the usefulness of the generated data, we produce 30K image pairs for training rectified stereo matching. We train RAFT-Stereo~\cite{lipson2021raft} on these images from scratch and compare against the same architecture trained on other synthetic datasets. Models are trained for 200k steps using the same hyper-parameters from~\cite{lipson2021raft}.

\parbf{Real Images of Natural Scenes} Because images from \projectname{} consist of entirely of natural scenes and are devoid of any human-made objects, we expect models trained on \projectname{} data to perform better on images with natural environments and worse on images without (e.g. indoor scenes). However, quantitative evaluation on real-world natural scenes is currently infeasible because  
there does not exist a real-world benchmark that evaluates depth estimation for natural scenes. Existing real-world benchmarks consist almost entirely of images of indoor environments dominated by artificial objects. In addition, it is challenging to obtain 3D ground truth for real-world natural scenes, because real-world natural scenes are often highly complex and non-static (e.g.\@ moving tree leaves and animals), making high-resolution laser-based 3D scanning impractical.  

 Due to the difficulty of obtaining 3D ground truth for real images of natural scenes, we perform qualitative evaluation instead. We collected high-resolution rectified stereo images of real-world natural scenes using the ZED 2 Stereo Camera~\cite{zed2} and visualize the predicted disparity maps from RAFT-Stereo~\cite{lipson2021raft} in Fig.~\ref{thefig:wildresults}. Our results show that a model trained entirely on synthetic scenes from \projectname{} can perform well on real images of natural scenes zero-shot. The model trained on \projectname{} data produces noticeably better results than models trained on existing datasets, suggesting that \projectname{} is useful in that it provides training data for a domain that is poorly covered by existing datasets.

\parbf{Middlebury Dataset} We evaluate our trained model on the Middlebury Dataset~\cite{middlebury}, which is a standard evaluation benchmark for stereo matching. It consists of megapixel image-pairs of cluttered indoor spaces, 10 with public ground-truth and 10 without. This benchmark is challenging due to its abundance of objects, textureless surfaces, and thin structures. In Tab.~\ref{thetab:middlebury_train}, we see that our \projectname{}-trained model struggles on images with exclusively artificial objects but performs well on  the only image with natural objects (Jadeplant). In Fig~\ref{thefig:middlebury_plants}, we qualitatively evaluate our model on Middlebury images without public ground-truth and observe that our model generalizes well to the natural scenes.

\begin{table}[h]
\centering
\resizebox{0.8\linewidth}{!}{
\begin{tabular}{l|c|l} \toprule
 Training Dataset & Bad 3.0 (\%) $\downarrow$ & \# Image Pairs \\
 \midrule
 InStereo2K~\cite{bao2020instereo2k} & 25.282 & 2K \\ 
 FallingThings~\cite{fallingthings} & 12.199 & 62K \\ 
 Sintel-Stereo~\cite{sintel} & 10.253 & 2K\\ 
 HR-VS~\cite{hrvs} & 9.296 & 780 \\ 
 Li et al.~\cite{crestereo} & 9.271 & 177K \\ 
 SceneFlow~\cite{sceneflow2016} & 7.837 & 35K \\ 
 TartanAir~\cite{wang2020tartanair} & 6.504 & 296K\\
 Ours (\projectname{} 30K) & \textbf{5.527} & 31K\\ 
 \bottomrule
\end{tabular}
}
\caption{Performance on 400 independent \projectname{} evaluation  scenes. No assets are shared between our \projectname{} training and evaluation scenes. }
\label{thetab:stereoquant}
\end{table}

\parbf{Synthetic Images of Natural Scenes} Although quantitative evaluation is not currently feasible on real-world natural scenes, it can be done using synthetic natural scenes from \projectname{}, 
with the caveat that we rely on the assumption that performance on \projectname{} images is a good proxy to real-world performance, as suggested by the qualitative results in Fig.~\ref{thefig:wildresults}. In Tab.~\ref{thetab:stereoquant}, we evaluate our \projectname{}-trained model on an independent set of 400 image pairs from \projectname{}. No assets are shared between our training and evaluation sets. Tab.~\ref{thetab:stereoquant} also compares the model trained on \projectname{} to models trained on other datasets. We see that the model trained on \projectname{} data has a significant lower error than those trained on other datasets. These quantitative results suggest that the distribution of \projectname{} images is significantly different from existing datasets and that \projectname{}  can serve as a useful supplement to existing datasets.

\begin{table*}[t]
\centering
\resizebox{0.8\linewidth}{!}{
\begin{tabular}{l|cccccccccc|c} \toprule
 Training Dataset & Adirondack & \textbf{Jadeplant} & Motorcycle & Piano & Pipes & Playroom & Playtable & Recycle & Shelves & Vintage & Avg\\
 \midrule
FallingThings~\cite{fallingthings} & 8.3 & 43.3 & \textbf{12.3} & 18.2 & 25.3 & 29.7 & 50.0 & 10.4 & 43.3 & 45.6 & 28.6\\ 
 Sintel-Stereo~\cite{sintel} & 35.7 & 62.9 & 31.1 & 24.1 & 31.9 & 41.7 & 60.1 & 30.8 & 55.8 & 76.1 & 45.0\\ 
 HR-VS~\cite{hrvs} & 43.5 & 43.2 & 17.0 & 29.6 & 32.1 & 34.6 & 68.4 & 24.7 & 57.4 & 34.9 & 38.5 \\
 Li et al.~\cite{crestereo} & 23.9 & 80.2 & 40.7 & 32.0 & 40.3 & 49.1 & 67.5 & 36.6 & 51.7 & 42.3 & 46.4\\ 
 SceneFlow~\cite{sceneflow2016} & \textbf{7.4} & 41.3 & 14.9 & 16.2 & 33.3 & \textbf{18.8} & \textbf{38.6} & \textbf{10.2} & \textbf{39.1} & 29.9 & \textbf{25.0}\\ 
 TartanAir~\cite{wang2020tartanair} & 15.5 & 45.1 & 18.1 & \textbf{12.9} & 28.4 & 25.6 & 51.0 & 20.9 & 49.1 & \textbf{28.2} & 29.5\\ 
 InStereo2K~\cite{bao2020instereo2k} & 17.1 & 59.7 & 21.3 & 23.8 & 35.8 & 33.9 & \textbf{36.4} & 20.0 & \textbf{33.4} & 44.1 & 32.5\\ 
 Ours (\projectname{} 30K) & \textbf{7.4} & \textbf{35.2} & 15.2 & 20.7 & \textbf{24.7} & 29.3 & 50.0 & 12.6 & 55.1 & 46.9 & 29.7\\ 
 \bottomrule
\end{tabular}
}
\caption{Bad 3.0 (\%) $\downarrow$ error on the Middlebury~\cite{middlebury} validation set. \projectname{} helps models generalize to images with natural objects (e.g. Jadeplant). On the other hand,  natural objects contain very few planar or texture-less surfaces; models trained exclusively on natural objects can generalize less well on indoor datasets like Middlebury.}
\label{thetab:middlebury_train}
\end{table*}

%% file: texts/16-Dataset-Generation.tex
\section{Dataset Generation}

 \newcommand{\galleryRowGT}[5]{
   \includegraphics[width=0.25\linewidth]{#1/#2} &
   \includegraphics[width=0.25\linewidth]{#1/#3} &
   \includegraphics[width=0.25\linewidth]{#1/#4} &
   \includegraphics[width=0.25\linewidth]{#1/#5}
  \tabularnewline 
 }

\begin{figure*}[t]
  \begin{center}
  \resizebox{\linewidth}{!}{
    \setlength{\tabcolsep}{0.2pt}
    \renewcommand{\arraystretch}{0.2}
    \begin{tabular}{cccc}
      \multirow{2}{*}{RGB} &
      \multirow{2}{*}{Depth} &
      Surface Normals + &
      Instance\vspace{0.1cm} \\
      & & Occlusion Boundaries & Segmentation \\
    \galleryRowGT{figures/new_ground_truth/row1}{Noisy_Image0100_00_00.pdf}{depth.pdf}{normals.pdf}{segmentation_better.pdf}
    \galleryRowGT{figures/new_ground_truth/row2}{Noisy_Image0100_00_00.pdf}{depth.pdf}{normals.pdf}{segmentation_better.pdf}
    \galleryRowGT{figures/new_ground_truth/row5}{Noisy_Image0100_00_00.pdf}{depth.pdf}{normals.pdf}{segmentation_better.pdf}
    \galleryRowGT{figures/new_ground_truth/row6}{Noisy_Image0100_00_00.pdf}{depth.pdf}{normals.pdf}{segmentation_better.pdf}
    \galleryRowGT{figures/new_ground_truth/row7}{Noisy_Image0100_00_00.pdf}{depth.pdf}{normals.pdf}{segmentation_better.pdf}
    \galleryRowGT{figures/new_ground_truth/row9}{Noisy_Image0100_00_00.pdf}{depth.pdf}{normals.pdf}{segmentation_better.pdf}
    \galleryRowGT{figures/new_ground_truth/row10}{Noisy_Image0100_00_00.pdf}{depth.pdf}{normals.pdf}{segmentation_better.pdf}
    \galleryRowGT{figures/new_ground_truth/row11}{Noisy_Image0100_00_00.pdf}{depth.pdf}{normals.pdf}{segmentation_better.pdf}
    \end{tabular}
  }
  \end{center}
  \caption{High-Resolution Ground Truth Samples. We show select ground truth maps for 8 example \projectname{} images. For space reasons, we show only Depth, Surface Normals / Occlusion and Instance Segmentation. Our instance segmentation is highly granular, but classes can be grouped arbitrarily using object metadata. See Sec.\ref{thesec:groundtruthexplanation} for a full explanation.}
  \label{thefig:groundtruthviz}
\end{figure*}

\begin{figure}[ht!]
\centering
\begin{subfigure}{.99\linewidth}
\includegraphics[width=\linewidth]{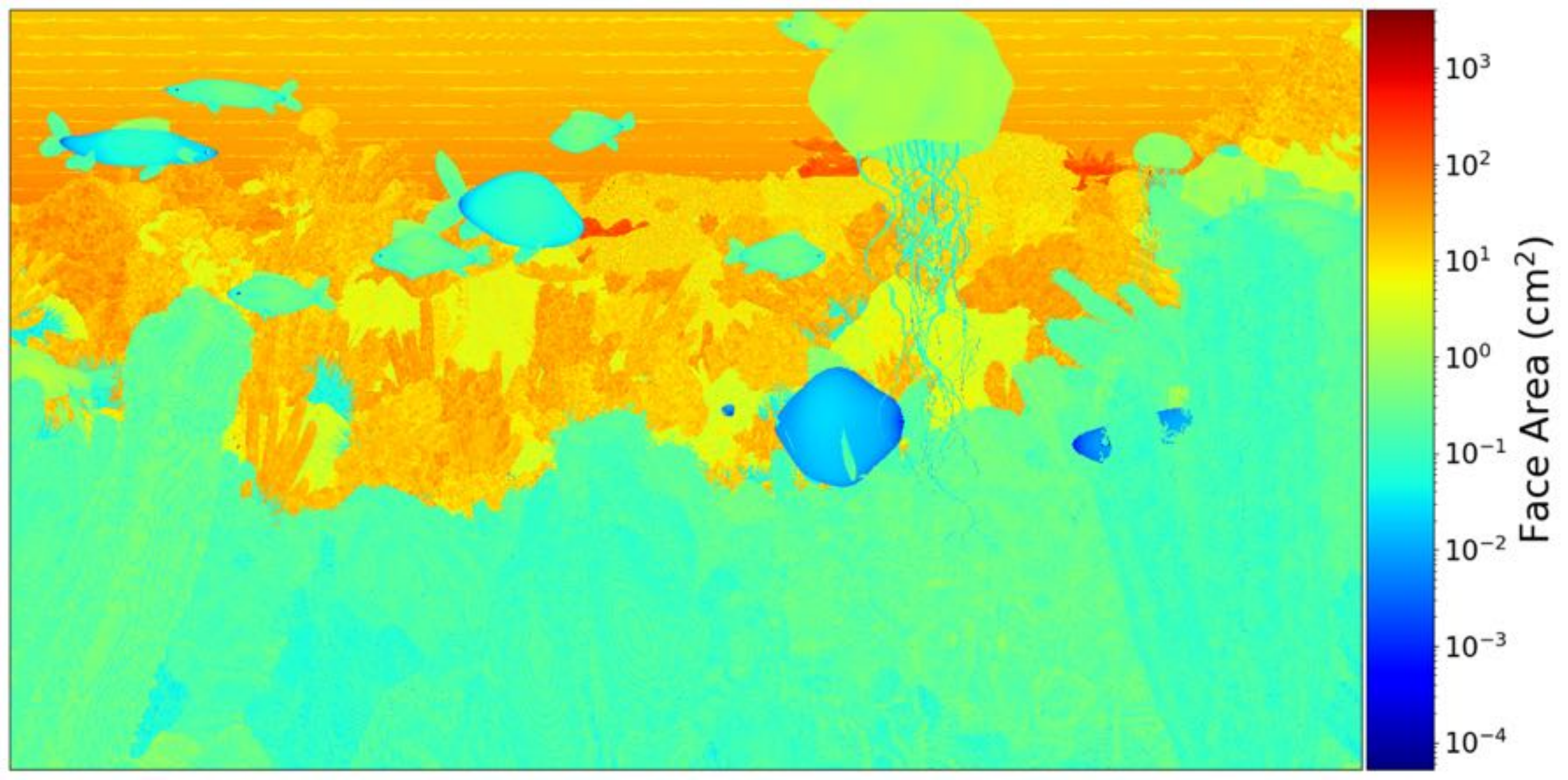}
\caption{The area of mesh faces in cm$^2$. Our dynamic-resolution scaling causes faces closer to the camera to be smaller.}\label{thefig:facesize}
\end{subfigure}
\begin{subfigure}{.94\linewidth}
\includegraphics[width=\linewidth]{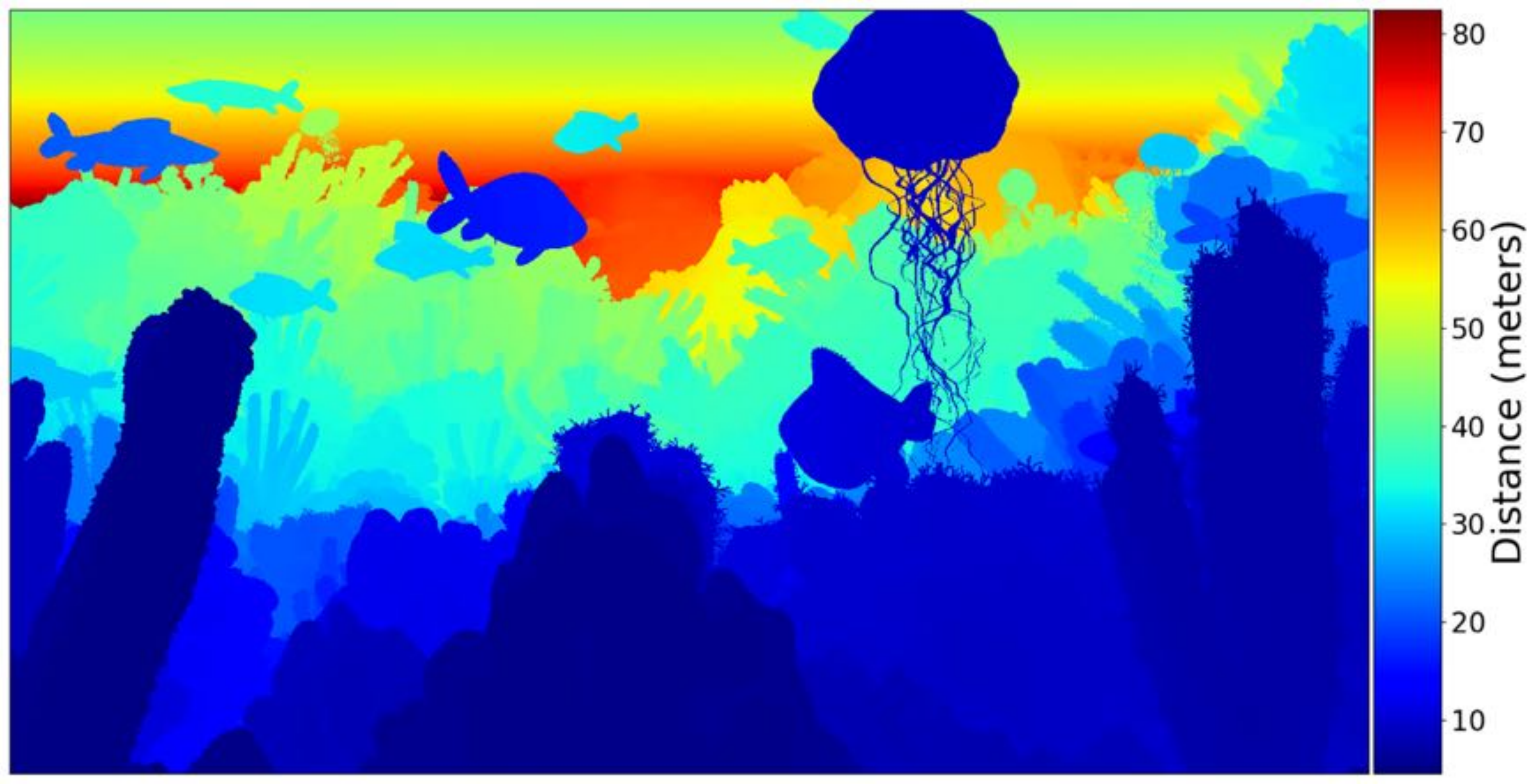}
\caption{Distance of faces from the camera (i.e. depth). Distance is proportional to the area of faces.}\label{thefig:distance}
\end{subfigure}
\begin{subfigure}{.99\linewidth}
\includegraphics[width=\linewidth]{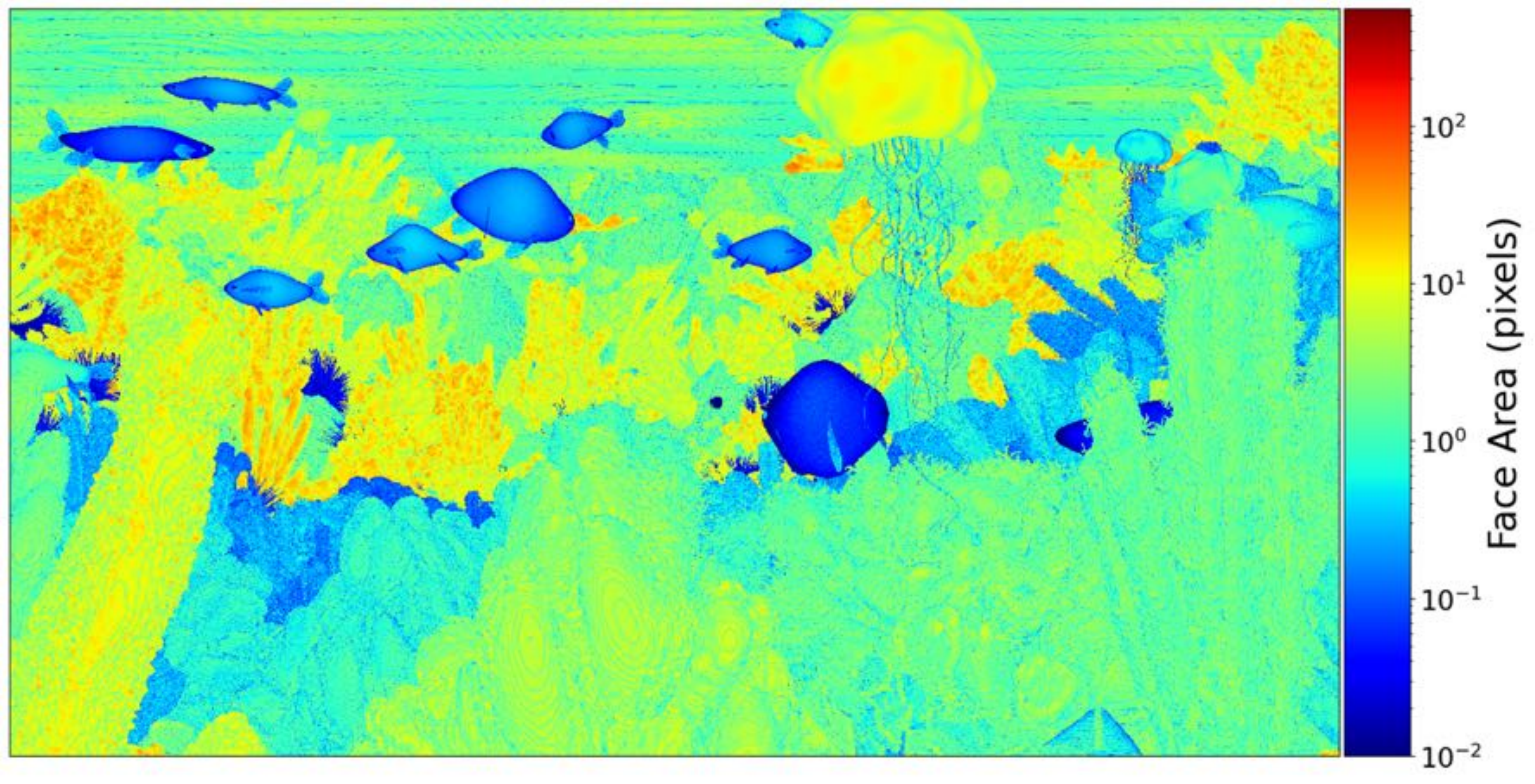}
\caption{Face area measured in pixels. Our dynamic resolution scaling causes individual mesh faces to appear approximately one pixel across. }\label{thefig:areapixel}
\end{subfigure}
\caption{Dynamic Resolution Scaling. Faces further from the camera are made smaller (a) such that they appear to be the same size from the camera's perspective (c). We show the depth map for reference (b).}
\label{thefig:face_area_heatmaps}
\end{figure}

\begin{figure}[ht!]
\centering
\begin{subfigure}{.87\linewidth}
\includegraphics[width=\linewidth]{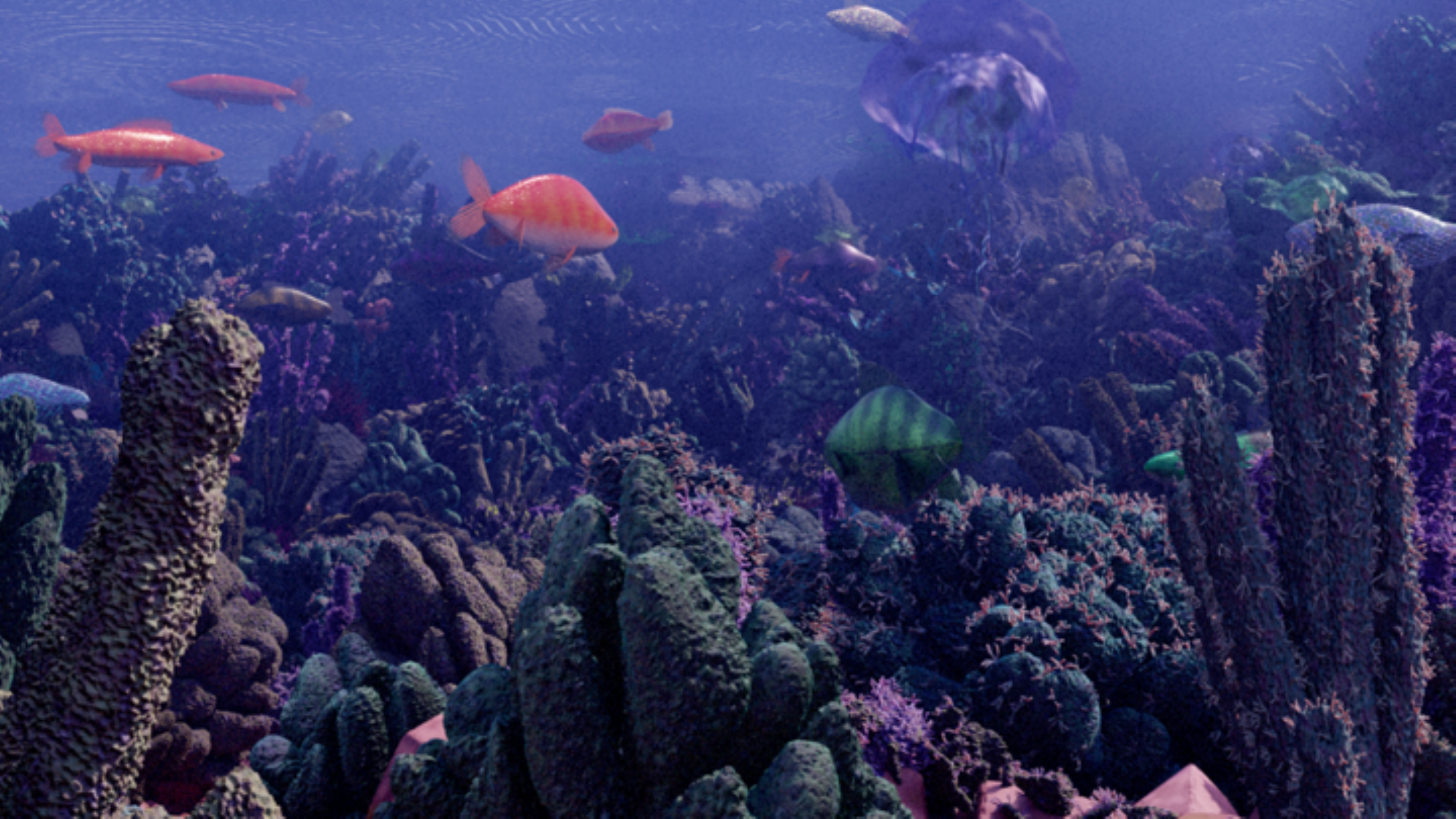}
\caption{Input Image for reference.\vspace{.4cm}}\label{thefig:badgt_input}
\end{subfigure}
\begin{subfigure}{.87\linewidth}
\includegraphics[width=\linewidth]{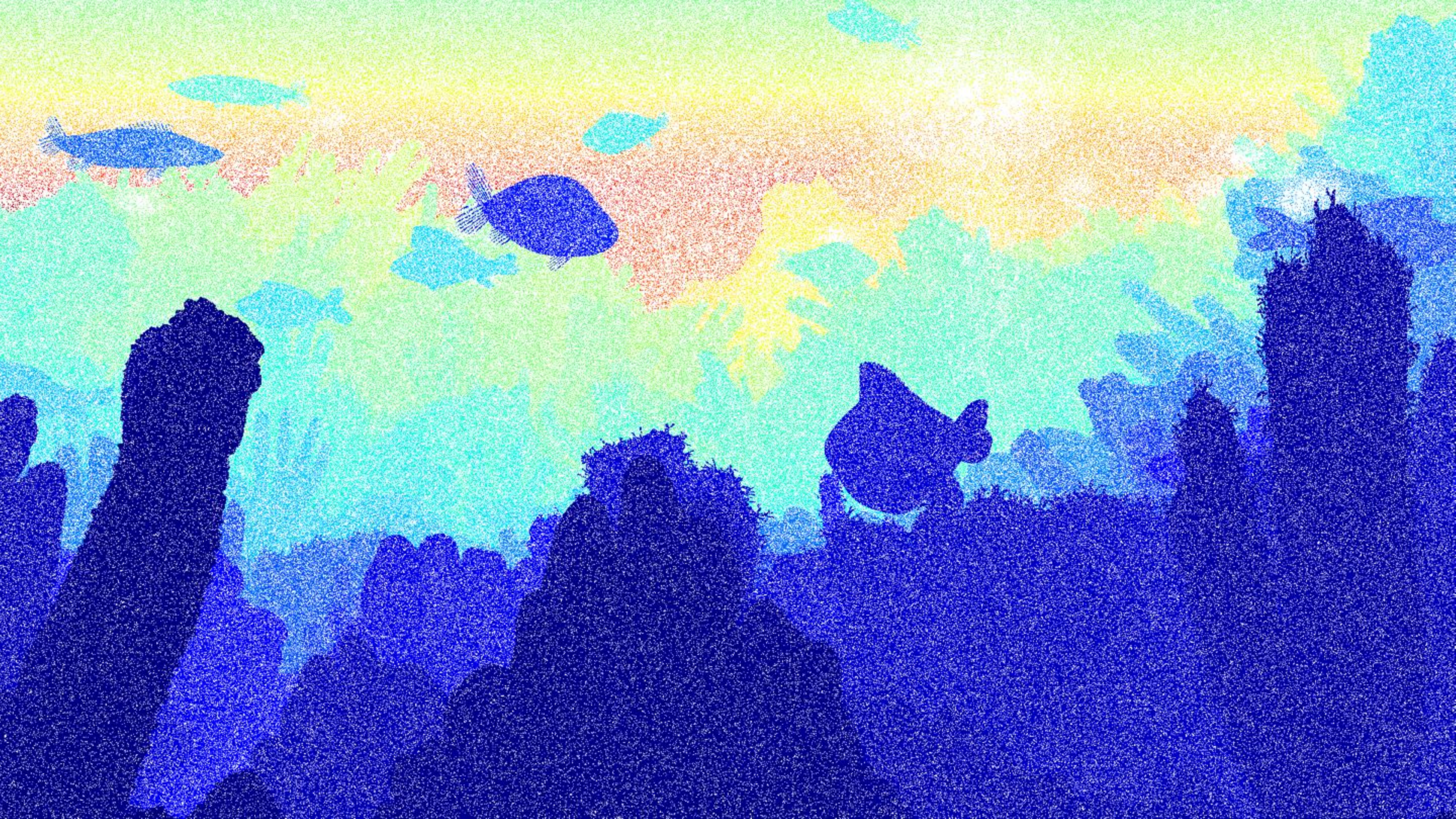}
\caption{Depth from Blender's built-in render passes.}\label{thefig:badgt_depth}
\end{subfigure}
\caption{Our ground truth is computed directly from the underlying geometry and is always exact. Prior methods \cite{kubric, hasson19_obman, crestereo, sintel, he2021semi} generate ground-truth from Blender's render-passes, which leads to noisy depth for volumetric effects such as water, fog, smoke, and semi-transparent objects.}
\label{thefig:builtingroundtruth}
\end{figure}

\subsection{Image Rendering}
We render images using Cycles, Blender's physically-based path tracing renderer. Cycles individually traces photons of light to accurately simulate diffuse and specular reflection, transparent refraction and volumetric effects. We render at $1920\times 1080$ resolution using $10,000$ random samples per-pixel. 

\subsection{Ground Truth Extraction}
\label{thesec:groundtruthexplanation}
We develop custom code for extracting ground-truth directly from the geometry. Prior datasets \cite{kubric, hasson19_obman, crestereo, sintel, he2021semi} rely on blender's built-in render-passes to obtain dense ground truth. However, these rendering passes are a byproduct of the rendering pipeline and not intended for training ML models. Specifically, they are incorrect for translucent surfaces, volumetric effects, or when motion blur, focus blur or sampling noise are present. 

We contribute OpenGL code to extract surface normals, depth, segmentation masks, and occlusion boundaries from the mesh directly without relying on blender. 

\parbf{Depth}
We show several examples of our depth maps in Fig.~\ref{thefig:groundtruthviz}. In Fig.~\ref{thefig:builtingroundtruth}, we visualize the alternative approach of naively producing depth using blender's built-in render passes. 

\parbf{Occlusion Boundaries}
We compute occlusion boundaries using the mesh geometry. Blender does not natively produce occlusion boundaries, and we are not aware of any other synthetic dataset or generator which provides exact occlusion boundaries.

\parbf{Surface Normals}
We compute surface normals by fitting a plane to the local depth map around each pixel. Sampling the geometry directly instead can lead to aliasing on high-frequency surfaces (e.g. grass).

The size of the plane used to fit the local depth map is configurable, effectively changing the resolution of the surface normals. We can also configure our sampling operation to exclude values which cannot be reached from the center of each plane without crossing an occlusion boundary; planes with fewer than 3 samples are marked as invalid. We show these occlusion-augmented surface normals in Fig.~\ref{thefig:groundtruthviz}. These surface normals appear only surfaces with sparse occlusion boundaries, and exclude surfaces like grass, moss, lichen, etc.\looseness=-1

\parbf{Segmentation Masks}
We compute instance segmentation masks for all objects in the scene, shown in Fig.~\ref{thefig:groundtruthviz}. Object meta-data can be used to group certain objects together arbitrarily (e.g. all grass gets the same label, a single tree gets one label, etc).

\parbf{Customizable}
Since our system is controllable and fully open-source, we anticipate that users will generate countless task-specific ground truth not covered above via simple extensions to our codebase.  

\subsection{Runtime}
\begin{figure*}
\centering
\begin{subfigure}[b]{.19\linewidth}
\includegraphics[width=\linewidth]{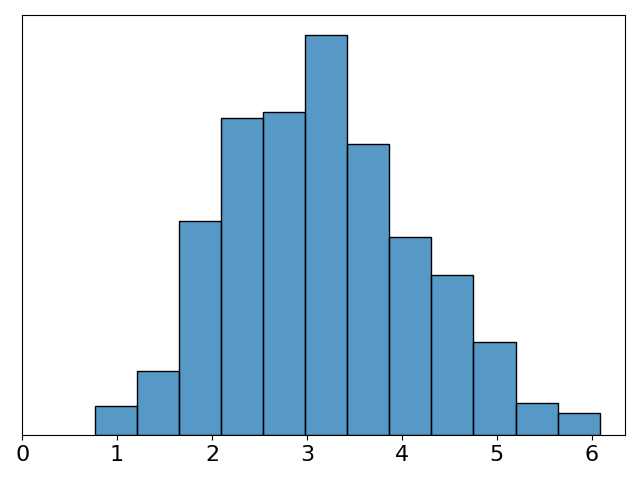}
\caption{Wall Time (Hours)}\label{thefig:elapsedhours}
\end{subfigure}
\begin{subfigure}[b]{.19\linewidth}
\includegraphics[width=\linewidth]{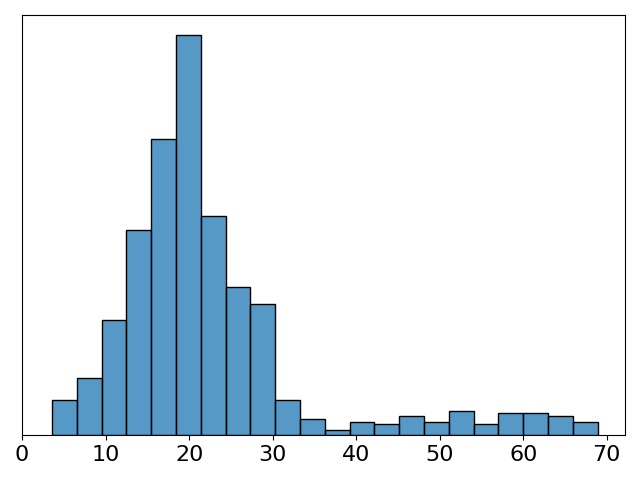}
\caption{Memory (GB)}\label{thefig:maxmemory}
\end{subfigure}
\begin{subfigure}[b]{.19\linewidth}
\includegraphics[width=\linewidth]{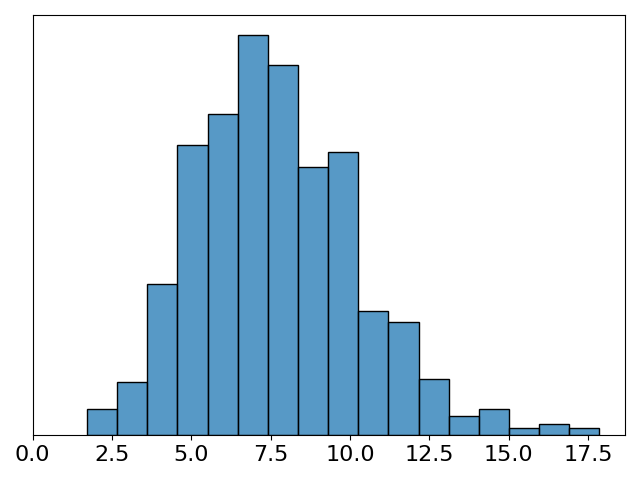}
\caption{CPU Hours}\label{thefig:cpuhours}
\end{subfigure}
\begin{subfigure}[b]{.19\linewidth}
\includegraphics[width=\linewidth]{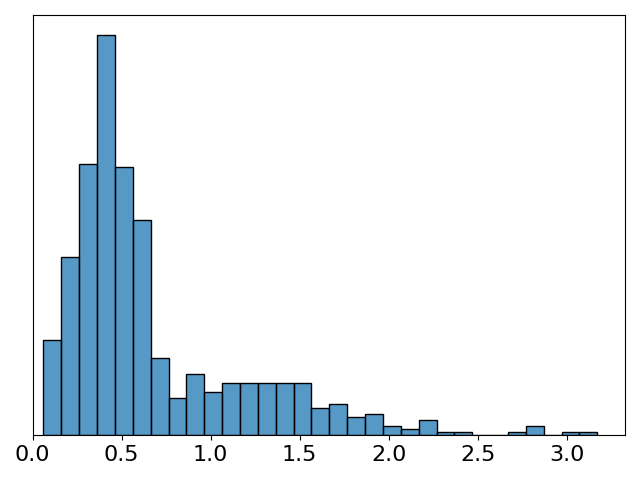}
\caption{GPU Hours}\label{thefig:gpuhours}
\end{subfigure}
\begin{subfigure}[b]{.19\linewidth}
\includegraphics[width=\linewidth]{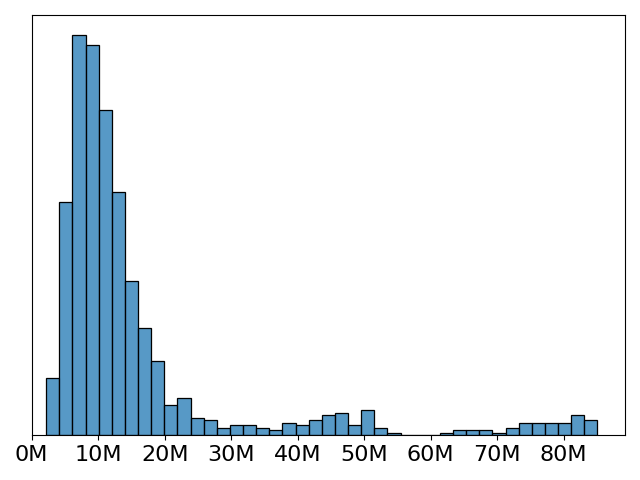}
\caption{\# Triangles per scene}\label{thefig:tricounts}
\end{subfigure}
\caption{Resource requirements for creating a pair of stereo 1080p images using \projectname{}. Our mesh resolutions scale with the output image resolution, such that individual mesh faces are barely visible. As a result, these statistics will change for different image resolutions.}
\label{thefig:resourcerequirements}
\end{figure*}

We benchmark \projectname{} on 2 \textit{Intel(R) Xeon(R) Silver 4114 @ 2.20GHz} CPUs and 1 NVidia-GPU (one of GTX-1080, RTX-[2080, 6000, a6000] or a40) across 1000 independent trials. We show the distribution in Fig.~\ref{thefig:resourcerequirements}. The average wall time to produce a pair of 1080p images is 3.5 hours. About one hour of this uses a GPU, for rendering specifically. More CPUs per-image-pair will decrease the wall-time significantly as will faster CPUs. Our system also uses about 24Gb of memory on average.

\parbf{Pre-generated \projectname{} Data} To maximize the accessibility of our system we will provide a large number of pre-generated assets, videos, and images from \projectname{} upon acceptance.

%% file: texts/12-degrees-of-freedom.tex
\section{Interpretable Degrees of Freedom}
\label{thesec:dof}
We attempted to estimate the complexity of our procedural system by counting the number of human-interpretable parameters, as shown in the per-category totals in Table 2 of the main paper. Here, we provide a more granular break down of what named parameters contributed to these results.

\paragraph{Counting Method}

We seek to provide a conservative estimate of the expressive capacity of our system. We only count distinct human-understandable parameters. We also only include parameters that are \textit{useful}, that is if it can be randomized within some neighborhood and produce noticeably different but still photorealistic assets. Each row of Tabs. \ref{thetab:dof_materials}--\ref{thetab:dof_scenecomp} gives the names of all \textit{Intepretable DOF} that are relevant to some set of \textit{Generators}.

We exclude trivial transformations such as scaling, rotating and translating an asset. We include absolute sizes such as 'Length' or 'Radius' as parameters only when their \textit{ratio} to some other part of the scene is significant, such as the leg to body ratio of a creature, or the ratio of a sand dune's height to width. 
 
Many of our material generators involve randomly generating colors using random HSV coordinates. This has three degrees of freedom, but out of caution we treat each color as one parameter. Usually, one or more HSV coordinates are restricted to a relatively narrow range, so this one parameter represents the value of the remaining axis. Equivalently, it can be imagined as a discrete parameter specifying some named color-palette to draw the color from. Some generators also contain compact functions or parametric mapping curves, each usually with 3-5 control points. We treat each curve as one parameter, as the effect of adjusting any one handle is subtle.  

\paragraph{Results}
In total, we counted 182 procedural asset generators with a sum of 1070 distinct interpretable parameters / Degrees-of-Freedom. We provide the full list of these named parameters as Tables \ref{thetab:dof_materials}--\ref{thetab:dof_scenecomp}, placed at the end of this document as they fill several pages.

%% file: texts/17-transpiler-details.tex
\section{Transpiler}
\label{thesec:transpiler}
\begin{figure}
    \centering
    \includegraphics[width=\linewidth]{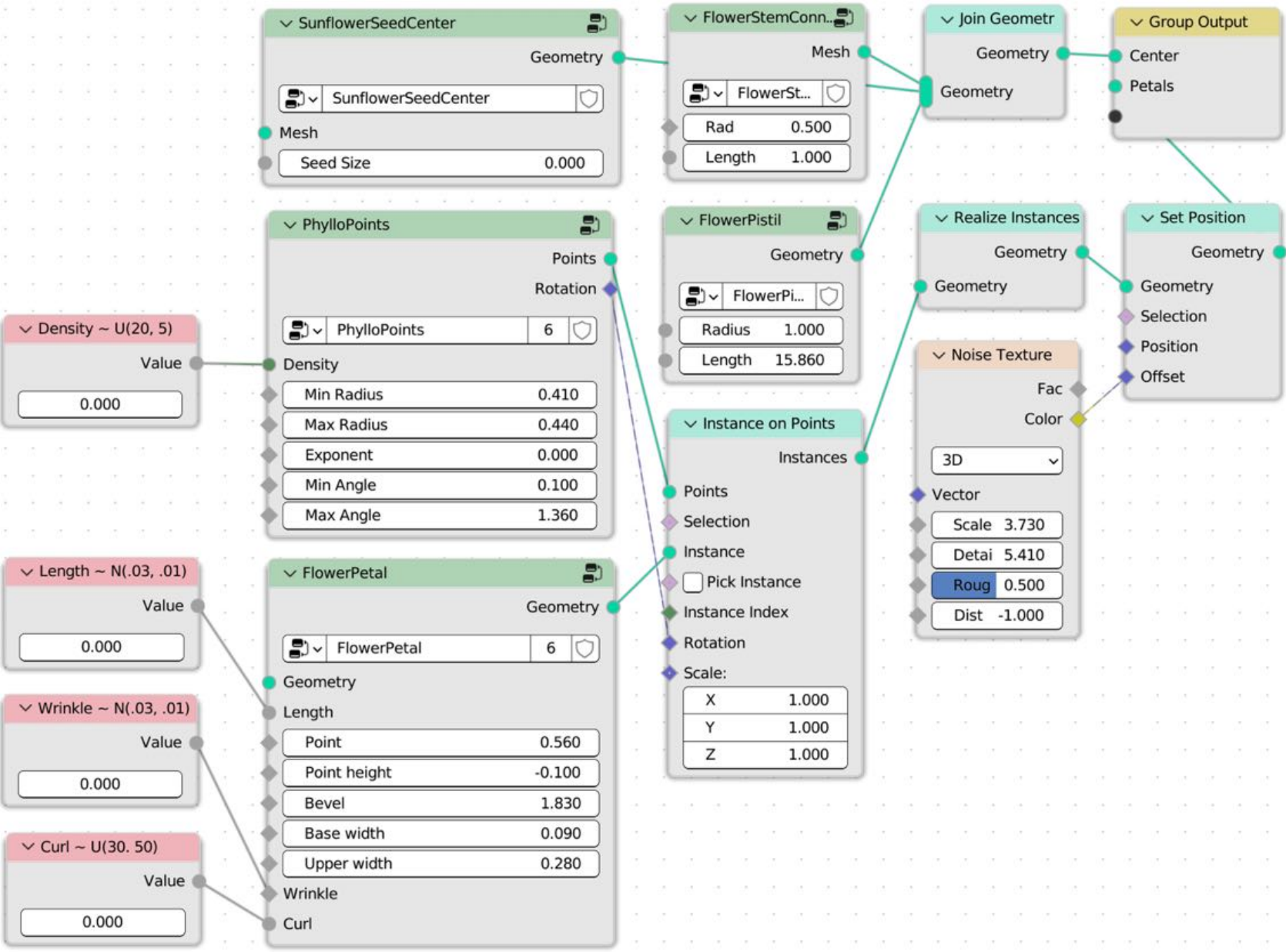}
    \caption{An example node-graph, as shown as input to the transpiler in Fig. 3 of the main paper. Dark green nodes are \textit{node groups}, containing user-defined node-graphs as their implementations. Red nodes show tuned constants, with annotations for their distribution.}
    \label{thefig:nodegroups}
\end{figure}

\paragraph{Code Generation}

In the simplest case, the transpiler is a recursive operation which performs a post-order traversal of Blender's internal representation of a node-graph, and produces a python statement defining each as a function call of it's children. We automatically handle and abstract away many edge cases present in the underlying node tree, such as enabled/disabled inputs, multi-input sockets and more. This procedure also supports all forms of blender nodes, including \textit{shader nodes, geometry nodes, light nodes} and \textit{compositor nodes}.

\paragraph{Doubly-recursive parsing}

Blender's node-graphs support many systems by which node-graphs can contain and depend on one another. Most often this is in the form of a node-group, such as the dark green boxes titled \textit{SunflowerSeedCenter} and \textit{PhylloPoints} in Fig. \ref{thefig:nodegroups}. Node-groups are user-defined nodes, containing an independent node-tree as an implementation. These are equivalent to functions in a typical programming languages, so whenever one is encountered, the transpiler will invoke itself on the node-graph implementation and package the result as a python function, before calling that function in the parent node-graph. In a similar fashion, we often use \textit{SetMaterial} nodes that reference a shader node-graph to be transpiled as a function.

\paragraph{Probability Distribution Annotation}
Node-graphs contain many internal parameters, which are artistically tuned by the user to produce a desired result. We provide a minimal interface for users to also specify the distribution of these parameters by writing small strings in the node name. See, for example, the red nodes in Fig. \ref{thefig:nodegroups}. When annotations of this format are detected, the transpiler automatically inserts calls to appropriate random number generators into the resulting code. 

%% file: texts/15-composition-in-detail.tex
\section{Scene Composition Details}

Our scenes are not individually staged images - for each image, we produce a map of an expansive and view-consistent world. One can select any camera pose or sequence of poses, which allows for video and other multi-view data generation.

To allow this, we start by sampling a full-scene ground surface. This is low-resolution, but is sufficient to approximate the surface of the terrain for the purpose of placing objects. We determine surface points using Poisson-Disc Sampling. This avoids the majority of asset-asset intersections. We modulate point density using procedural masks based on surface normals, Perlin noise and terrain attributes. Asset rotations are determined uniformly at random. Our final coarse global map is represented as a lightweight blender file with intuitive editable placeholders to represent where assets will be spawned in later steps of the pipeline. 

We provide a library of 11 optional configuration files to modify scene composition, namely \textit{Arctic, Coast, Canyon, Cave, Cliff, Desert, Forest, Mountain, Plains, River} and \textit{Underwater}. Each encodes simple natural priors such as "Cacti often grow in deserts" or "Trees are less dense on mountains", expressed as modifications to these mask and density parameters. More complex relations, like predators and prey avoiding each other, or plants not growing in shaded areas, are not currently captured. 

\subsection{Camera Selection}
We select camera viewpoints with simple heuristics designed to match the perspective of a creature or person, which are as follow:
\paragraph{Height above Ground}
In order to match the perspective of a creature or person, we sample the camera height above ground from a Gaussian distribution. (with the exception that in terrain-only scenes sometimes this height is higher to highlight some landscape features)
\paragraph{Minimum Distance}
To avoid being blocked by a close-up object and over-subdividing the geometry (which is expensive), we select camera views with a minimum distance threshold to all objects.
\paragraph{Coverage}
In order to avoid overly barren images and to highlight interesting features, we may select views such that a certain terrain component, e.g. a river, is visible. Specifically, we may require that the camera view has pixels from a specific terrain component or object type within a certain range.
\paragraph{Standard Deviation of Depth}
We compute the variance of pixel-wise depth values, and choose the viewpoint out of ten random samples with the largest variance to favor more interesting content.

\subsection{Dynamic Resolution}

In Fig.~\ref{thefig:face_area_heatmaps}, we show a visualization of triangle sizes in cm$^2$ and in pixels as viewed from the camera. Face size in meters increases proportionally to depth, whereas face size in pixels remains approximately constant. 

\subsubsection{Spherical Marching Cubes}

To generate a mesh for a specific camera view, we must extract a mesh representation of the terrain which contains dense pixel-size faces. Classical \textit{Marching Cubes} struggles to achieve this, as it evaluates the SDF at fixed intervals, which results in too-sparse geometry near the camera and too-dense geometry in the far distance. Spherical Marching Cubes is our novel adaptation of this classic algorithm to operate in spherical coordinates, which automatically creates dense geometry near the camera where it is most needed, thereby preventing waste and drastically improving performance.

Spherical Marching Cube algorithm works as follows:
\paragraph{Low Resolution Step} We divide the visible 3D space (within the frustrum camera and a certain distance range $(d_{min}, d_{max})$) into $M \times N \times R$ blocks in spherical coordinates, uniform in $\theta$ and $\phi$ and logarithmically in $r$. We convert these to cartesian coordinates and evaluate the SDF as usual. This yields a tensor of SDF values where grid cells near $d_{max}$ represent larger regions of space than those close to the camera, thereby saving space. 
\paragraph{Visible Block Search Step} We use this SDF tensor to find the closest block for each pixel with an SDF zero crossing, resulting in an approximate terrain-only depth-map. These blocks are low-resolution and may contain holes, so we check them again with dense SDF queries to determine any farther away visible blocks.
\paragraph{High Resolution Step} Finally, we evaluate dense pixel-size SDF queries for all visible blocks, and produce a finalized mesh with Marching Cubes. Theoretically the final size of this mesh can still be proportional to cube of the resolution, but in practice we find it is proportional to square. This step and the previous step can be performed iteratively to prevent all potential holes.
\paragraph{Out-of-view Part} The out-of-view part of the terrain is needed for lighting effects but done with low resolution.

\subsubsection{Parametric Surface Resolution Scaling}
NURBS and other parametric surfaces support evaluation at any mesh resolution. This is achieved by specifying some $du, dv$ as step sizes in parameter space. We provide heuristics to compute appropriate values for these step sizes such that the resulting mesh achieves a given max pixel size. 

\subsubsection{Subdivision and Remeshing}
All other assets rely on established Subdivision and Voxel-Remeshing algorithms to create dense pixel-size geometry. We provide heuristics to compute appropriate subdivision levels. Voxel Remeshing is time intensive but is especially useful for creatures and trees whose geometry can otherwise self-intersect or contain stretched faces.

%% file: texts/14-assets-in-detail.tex
\section{Asset Implementation Details}
\label{thesec:imp}
\subsection{Materials}
\label{thesec:materials}

Our materials are composed of a shader and a local geometry template. The shader procedurally generates realistic color, roughness, specularity, metallicity, subsurface scattering, and translucence parameters. The local geometry template generates corresponding geometric detail. Most often, the local geometry template simply computes a scalar field over the underlying mesh vertices, and displaces them along their normals to form a rough texture.

\subsubsection{Terrain Materials}

The majority of our terrain materials (including \textit{Mountain, Granite, Snow, Stone, Ice)} operate by combining many octaves of Perlin Noise to form geometric texture, before applying a mostly flat color. Some, such as some random variations of \textit{Mountain}, create layered color masks as a function of the world Z coordinate. Others such as \textit{Sand} follow a similar scheme, but with a procedural Wave texture instead of Perlin Noise.

Our \textit{Mud} and \textit{Sandstone} materials are particular expressive. \textit{Mud} procedurally generates puddles and slick ground by altering color, displacement and roughness jointly. Sandstone generates realistic layered sedimentary rock using noise functions and modular arithmetic based on the world Z coordinate. 

\textit{Lava} uses displacement from Perlin noise, F1-smooth Voronoi texture, and wave texture with varying scales. The shader uses Perlin noise, Voronoi texture to model hot and rocky parts, mixed with blackbody emission and a principled BSDF.

\textit{Fire} and \textit{Smoke} are comprised of multiple volumetric shaders. The first shader uses a principled volume shader with blackbody radiation whose intensity is sampled based on the amount of flame and smoke density. The smoke density is randomly sampled. The second shader imitates a high detail image captured with fast shutter speed and low exposure. The detail is brought out based on contrasting different regions of temperature and adding Perlin noise. The colors are shades of orange.

All our water materials feature glass-like surfaces with physically accurate Index-of-Refraction, combined with a volumetric scattering shader to simulate realistic underwater light bounces. The \textit{Water} shader uses these effects with geometry untouched (for use with simulators), wherase \textit{Water Surface} assumes the base geometry is a plane and adds geometric ripples.

\subsubsection{Plant Materials}
Our plant materials feature geometric and color variation using procedural Stucci, Marble and Shot Noise textures. We provide color pallettes for realistic plant and coral colors, and produce variations on these using Musgrave noise. All leafy plants feature realistic transmission and roughness properties, to simulate light filtering through translucent waxy leaves. Our \textit{Bark} and \textbf{Bark Birch} simulate bark expansion and fracturing using voronoi and perlin noise to generate displacements.


\subsubsection{Creature Material}

\paragraph{Fish}
We create two fish materials, a goldfish material and a regular fish material. Each material consists of two parts, a fish body material and a fish fin material. The fish body material creates the displacement of fish scales. To build the pattern of fish scales, we create a grid and fill every two adjacent grids in one column with a half circle. Then we move up the half circles in the even columns by one grid. The colors of regular fish are guided by one wave texture and two noise textures. The colors of goldfish are guided by two noise textures and sampled between red and yellow. The colors of fish bellies are usually white. A fish fin is created by adding periodical bumps to round planes. The weights of bump displacement are decreasing away from the fish body. Dorsal fins are sometimes serrated. The goldfish fins are translucent by mixing a principled BSDF shader, a transparent shader and a translucent shader. 

\paragraph{Bird}
Since our generated birds have particle fur, the bird material need only create a colormap. We create masks that highlight different body parts, including heads, necks, upper bodies, lower bodies and wings. Each part is assigned two similar colors guided by a noise texture. Our Bird material generator has two modes of colors, one with light colored heads and dark color bodies (emulating bald eagles and falcons), one with dark color heads and light color bodies (emulating ducks and common birds). 

\paragraph{Carnivore and Herbivore}
Carnivores and Herbivores are also equipped with particle fur and therefore color-only materials. We provide a \textit{Tiger} material, which makes short stripes by cutting a small-scale wave texture with another larger-scale wave texture. Our \textit{Giraffe} material builds spots by subtracting a F1-smooth voronoi texture from a F1 voronoi texture with the same scale. Three other spot materials build scattered and sometimes overlapped spots using noise textures and voronoi textures. Our three reptile materials build colormaps inspired by different kinds of reptiles. We pick their colors to mimic the reference reptile images, and then randomize in a small neighborhood. 

\paragraph{Beetle}
We provide a \textit{Chitin} material emulating the material of real beetles and other insects. It uses a computed "Pointiness" attribute to highlight the boundaries with sharp curvature. We color the insect head and sharp boundaries black, and other areas dark red or brown.

\paragraph{Bone, Beak \& Horns}
\textit{Bone} is most often used for animal claws and teeth. It starts wit a white to light gray color, before creating small pits in two different scales using noise and voronoi textures. Our \textit{Horn} material samples light brown to dark brown colors, with a similar mechanism for pits and scratches. Our \textit{Beak} material replicates realistic bird beaks by sampling a random yellow/orange/black color gradient along. Noise textures are used to add some small pits on it. Principled BSDF shaders are added to make the beaks more shiny.

\paragraph{Slime}
We provide a material titled \textit{Slimy}, which replicates folded shiny flesh similar to a "Blobfish". It builds thin and distorted stripe displacement using a noise texture followed by a wave texture. We create the shiny appearance by assigning high specularity and subsurface scattering parameters. 

\subsubsection{Other material}
\paragraph{Metal}
We build a silver material and an aluminium material. These use Blender's Principled BSDF shader with \textit{Metallic} set to $1$. They also have sparse sunken displacement from a noise texture. 

\subsection{Terrain}

\begin{figure}[H]
    \centering
    \includegraphics[width=\linewidth]{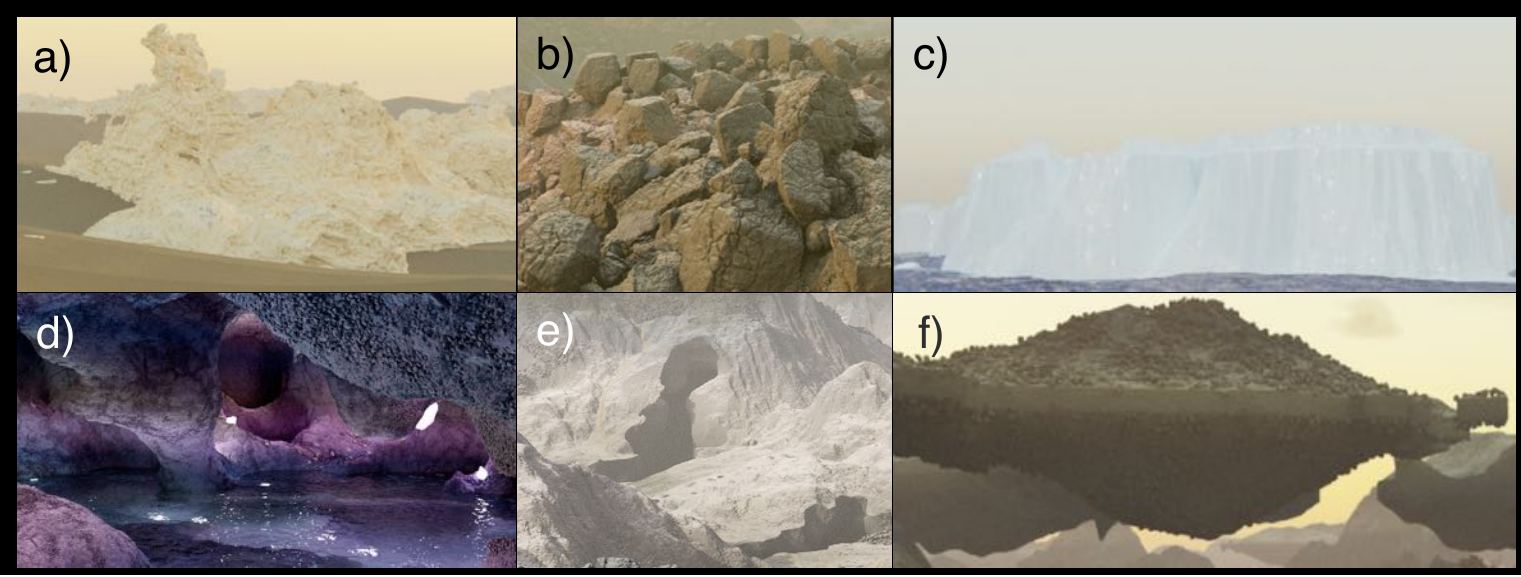}
    
  \caption{Terrain elements, including a) wind-eroded rocks, b) Voronoi rocks, c) Tiled landscapes, d/e) Caves, and f) Floating Islands.}

  \label{thefig:terrain_elements}
  
\end{figure}

\subsubsection{Terrain Elements}
The main part of terrain (\textit{Ground Surface}) is composed of a set of terrain elements represented by Signed Distance Functions (SDF).
Using SDF has the following advantages:
\begin{itemize}
    \item SDF is written in C/C++ language and can be compiled either in CPU (with OpenMP speedup) or CUDA to achieve parallelization.
    \item SDF can be evaluated at arbitrary precision and range, producing a mesh with arbitrary details and extent.
    \item SDF is flexible for composition. Boolean operations are just minimum and maximum of SDF. To put cellular rocks onto a mountain, we just query whether each corresponding Voronoi center has positive SDF of the mountain.
\end{itemize}

Terrain elements include:

    \paragraph{Wind Eroded Rocks} They are made out of Perlin noise~\cite{perlin1985image} from FastNoise Lite~\cite{FastNoiseLite} with domain warping  adapted from an article~\cite{geiss2007generating}. See Fig.~\ref{thefig:terrain_elements}(a) for an example.
    
    \paragraph{Voronoi Rocks} These are made from Cellular Noise (Voronoi Noise) from FastNoise Lite. They take another terrain element as input and generate cells that are on the surface of the given terrain element and add noisy gaps to the cells. See Fig.~\ref{thefig:terrain_elements}(b) for an example. We utilize this element to replicate fragmented rubble, small rocks and even tiny sand grains.
    \paragraph{Tiled Landscape}
    Complementary to above, we generate heterogeneous terrain elements as finite domain tiles. First, we generate a primitive tile using the A.N.T. Landscape Add-on in Blender~\cite{blender}, or a function from FastNoise Lite. This tile has a finite size, so we can simulate various natural process on it, e.g., erosion by SoilMachine~\cite{SoilMachine}, snowfall by diffusion algorithm in Landlab~\cite{Hutton_landlab_2020}~\cite{barnhart2020landlab}~\cite{hobley2017creative}. Finally, various types of tiles can be used alone or in combination to generate infinite scenes including mountains, rivers, coastal beaches, volcanos, cliffs, and icebergs. Fig.~\ref{thefig:terrain_elements}(c) shows an example of a iceberg tile. Tiles are combined by repeating them with random rotations and smoothing any boundaries. The resulting terrain element is still represented as an SDF.
    
    \paragraph{Caves}
    Our terrain includes extensive cave systems. These are cut out from other SDF elements before the mesh is created. The cave passages are generated procedurally using an Lindenmayer-System with probabilistic rules, where each rule controls the direction and movement of a virtual turtle~\cite{mark2015procedural}. These rules include turns, elevation changes, forks, and others. These passages have varying cross section shapes and are unioned with each other, leading to complicated cave systems with features from small gaps to large caverns. One can intuitively tune the cavern size, tunnel frequency and fork frequency by adjusting the likelihoods of various random rules.
    See Fig.~\ref{thefig:terrain_elements}(d) for an interior view of a cave and Fig.~\ref{thefig:terrain_elements}(e) for how it cut from mountains.
    \paragraph{Floating islands} Besides natural scenes, we also have fantastical terrain elements, e.g., floating islands (Fig.~\ref{thefig:terrain_elements}(f)) by gluing mountains and upside down mountains together.

\subsubsection{Boulders}
We start off with a mesh built from convex hull of around 32 vertices. We randomly select some faces that are large enough, so that they can be extruded and scaled. We repeat this process for two levels of extrusions: large and small. Finally we bevel the mesh, and add a displacement based on high- and low-frequency Voronoi textures. After generating the base mesh, boulders are given a rock surface and optionally a rock cover surface. See Sec.~\ref{thesec:materials} for details. Boulders are placed on the terrain mesh as placeholders.

\subsubsection{Fluid}
Most water and lava in our scenes form relatively static pools and lakes --- these are handled by generating a flat plane and applying a \textit{Surface Water} or \textit{Lava} material from Sec. \ref{thesec:materials}. 
\paragraph{Ocean} We simulate dyhnamic oceans Blender's built-in modifier to generate displacements on top of a water plane. This simulation is finite domain, so we tile it as described in \textit{Tiled Landscape} above.
\paragraph{Dynamic Fluid Simulation} We generate dynamic water and lava simulations  using Blender's built-in Fluid-Implicit-Particle (FLIP) plugin \cite{brackbill_flip_1988}. They can either simulated on a small region of the terrain or work together with Tiled Landscape, e.g., Volcanos to be reused as instances. The simulations are parameterized by sampled values of vorticity, viscosity, surface tension, flow amount, and other liquid parameters.
We simulate fire and smoke simulations using Blender's particle simulator. Our system allows for 1) Simulating fire and smoke on small random regions on the terrain or 2) Choosing arbitrary meshed assets on the scene to be set on fire or emit smoke. The simulations interact with turbulent and laminar wind flows added on the scene.
While these simulators are provided with Blender, we contribute significant engineering effort to automate their use. Typically, users manually set up individual simulations and execute them through the UI - we do so programmatically and at large scale.   

\subsubsection{Weather}
We provide procedural SDF functions for 4 realistic categories of clouds, each implemented as node-graphs. We implement rain and snow using Blender's particle system and wind simulation. We also apply atmospheric volume scattering to the entire scene to create haze, fog etc.

\subsubsection{Lighting} The majority of scenes are lit only by the sun and sky. We simulate these using the built in \textit{Nishita} \cite{nishitalighting} sky model, with randomized parameters for the Sun's position and brightness, as well as atmospheric parameters. In cave scenes, we place glowing gemstones as natural proxies for point lights. In underwater scenes, ray-traced refractive caustics are too costly at render-time, so we substitute textured spot lights. Finally, we provide an option to attach a virtual flash light or area light to the camera, to simulate a human or robot with an attached light.

\subsection{Plants}

\subsubsection{Leaves, Flowers \& Pinecones}



\paragraph{Pinecones}
Pinecones are the woody seed-bearing organs for conifers, which features scales and bracts arranged around a central axis, as shown in \fig{thefig:mushroom} b). Pinecones are made from individual buds. Each buds are sculpted from a mesh circle, with its left most point chosen as the origin. The vertices are displaced along the axis to the origin as well as along the Z axis, with scale designated by the direction from the origin to that point. We then create a mesh line along the Z-axis to form the stem of the pinecone. Pinecone buds are distributed on the axis from bottom to up with a decreasing scale and changing rotation. The rotation is composed of two parts: one along the Z axis that spread the buds around the pinecone, another along the X axis the gradually point the buds upwards. Pine shaders are made from Principled BSDF of a single color. Pinecones are scatters on the terrain mesh surfaces. 

\paragraph{Leaves}
 Our leaf generation system covers common leaf types including oval-shaped (which covers most of the broad-leaved trees), maple, ginkgo, and pine twigs. 
 
 For oval-shaped leaves, we start by subdividing a 2D plane mesh finely into grids, and evaluate each grid location with various noise functions. The leaf boundary defined by a set of control points of a Blender curve node, which specifies the width of the leaf at each location along the main stem. We then delete the unused geometry to get a rough shape of the leaf. To create veins, we use a 1D \textit{Voronoi Texture} node on a rotated coordinate system, to model the angle between the veins and the main stem. We extrude the veins and pass the height values into the Shader Node Tree to assign them different colors. We then add jigsaw-like patterns on the boundaries of the leaf, and create the cell structures using a 2D \textit{Voronoi Texture} node. We finally add wrapping effect to the leaf with another curve node.
 
 The maple leaves and ginkgo leaves are created in a very similar way as the oval-shaped leaves, except that we use polar coordinates to model rotational symmetric patterns, and the shape of the leaves is defined by a curve node in the polar coordinates. 
 
 Pine twigs are created by placing pine needles on a main stem, whose curvature and length are randomized.
 
 We use a mixture of translucent, diffuse, glossy BSDF to represent the leaf materials. The base colors are randomly sampled in the HSV space, and the distribution is tuned for each season (\textit{e.g.}, more yellow and red colors for autumn).
 
\paragraph{FlowerPlant}
We create the stem of a flower plant with a curve line together with an  cylindrical geometry. The radius of the stem gradually shrinks from the bottom to the top. Leaves are randomly attached to resampled points on the curve line with random orientations and scales within a reasonable range. Each leaf is sampled from a pool of leaf-like meshes. Additionally, we add extra branches to the main stem to mimic the forked shape of flower plants. Flowers are attached at the top of the stem and the top of the branches. Additionally, the stem is randomly rotated w.r.t the top point along all axes to generate curly looks of natural flower plants.

\subsubsection{Trees \& Bushes}

We create a tree with the following steps: 1) \textit{Skeleton Creation} 2) \textit{Skinning} 3) \textit{Leaves Placement}. 

\noindent\textbf{Skeleton Creation.} This step creates a directed tree-graph to represent the skeleton of a tree. Starting from a graph containing a single root node, we apply \textit{Recursive Paths} to grow the tree. Specifically, in each growing step, a new node is added as the child of the current leaf node. We computed the growing direction of the new node as the weighted sum of the previous direction (representing the momentum) and a random unit vector (representing the randomness). We further add an optional \textit{pulling direction} as a bias to the growing direction, to model effects such as gravity. We also specify the nodes where the tree should be branching, where we add several child nodes and apply Recursive Paths for all of them. Finally, given the skeleton created by Recursive Paths, we use the \textit{space colonization} algorithm\cite{runions2007modeling} to create dense and natural-looking branches. We scatter attraction points uniformly in a cube around the generated skeleton, and run the space colonization for a fixed amount of steps.

\noindent\textbf{Skinning.} We convert the skeleton into a Blender curve object, and put cylinders around the edges, whose radius grows exponentially from the leaf node to the root node. We then apply a procedural tree bark material to the surface of the cylinders. Instead of using a UV, we directly evaluate the values of the bark material in the 3D coordinate space to avoid seams. Since the bark patterns are usually anisotropic (\textit{e.g.}, strip-like patterns along the principal direction of the tree trunks), we use the local coordinate of the cylinders, up to some translation to avoid seams in the boundaries. 

\noindent\textbf{Leaves Placement.} We place leaves on twigs, and then twigs on trees. Twigs are created using the same skeleton creation and skinning methods, with smaller radius and more branches. Leaves are placed on the leaf nodes of the twig skeleton, with random rotation and possibly multiple instances on the same node. We use the same strategy to place the twigs on the trees, again with multiple instances to make the leaves very dense. For each tree we create 5 twig templates and reuse them all over the tree by doing \textit{instancing} in Blender, to strike a balance between diversity and memory cost.

Compared to existing tree generation systems such as the \textit{Sapling-Tree-Gen}\footnote{https://docs.blender.org/manual/en/latest/addons/add\_curve/sapling.html} addon in Blender and \textit{Speed-Tree}\footnote{https://store.speedtree.com} in UE4, our tree generation system creates leaves and barks using real geometry with much denser polygons, and thus provides high-quality ground-truth labels for segmentation and depth estimation tasks. We find this generation procedural very general and flexible, whose parameter space can cover a large number of tree species in the real world.

\begin{figure}
    \centering
    \includegraphics[width=\linewidth]{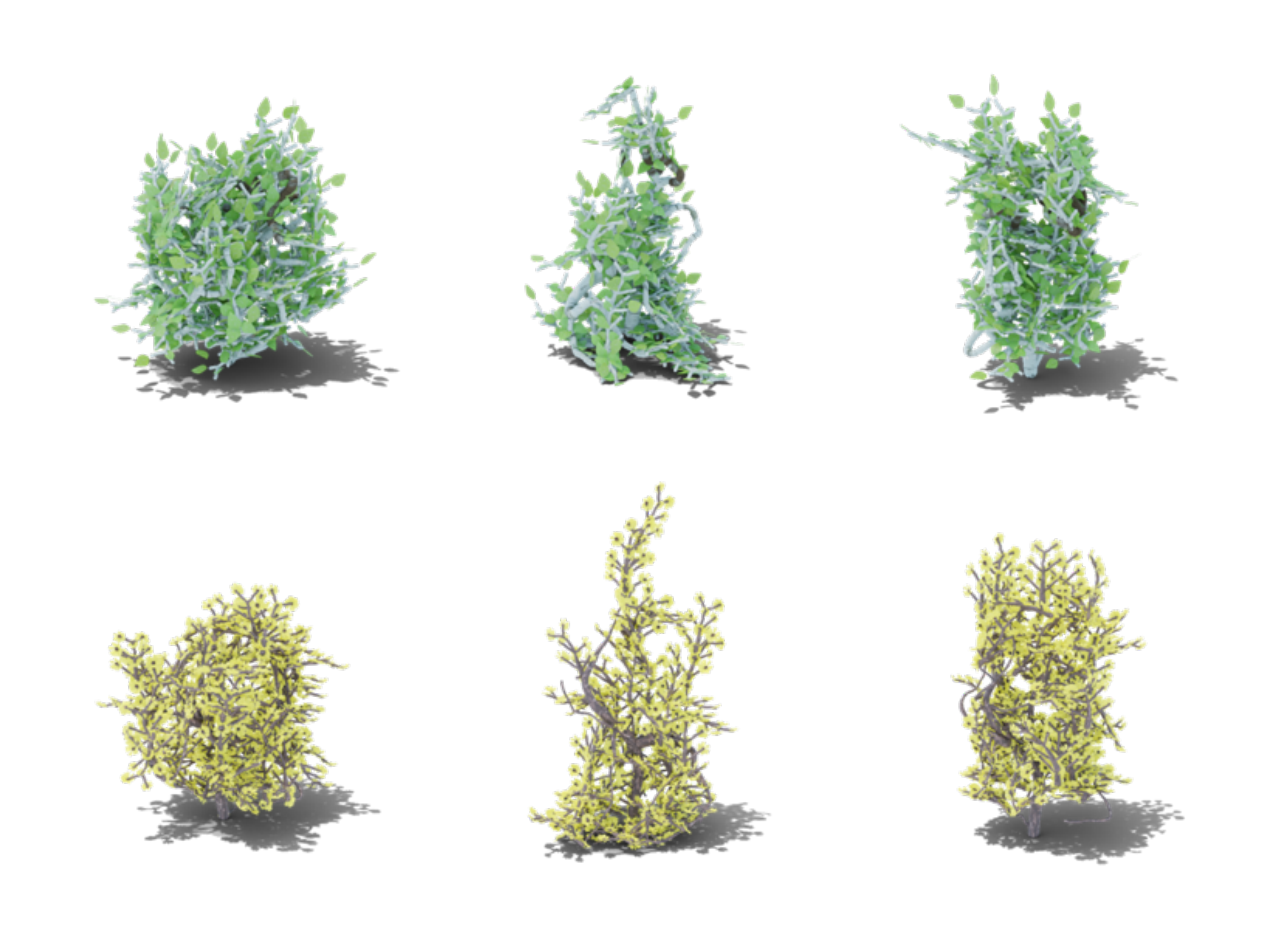}
    \caption{We control the shape of bushes by by specifying the distributions of the attraction points. Each row are the same bush species with different shapes (left to right: ball, cone, cube).}
    \label{thefig:bushes_shape}
\end{figure}

\noindent\textbf{Bushes} We also use this system to create other plants such as bushes, which have smaller heights and more branching compared to trees. Our system models the landscaping bushes that are pruned to different shapes by specifying the distributions of the attraction points in the space colonization step. Our bushes can be of either cone, cube or ball shaped, as shown in Fig.~\ref{thefig:bushes_shape}.

\begin{figure*}
    \centering
    \includegraphics[width=\linewidth]{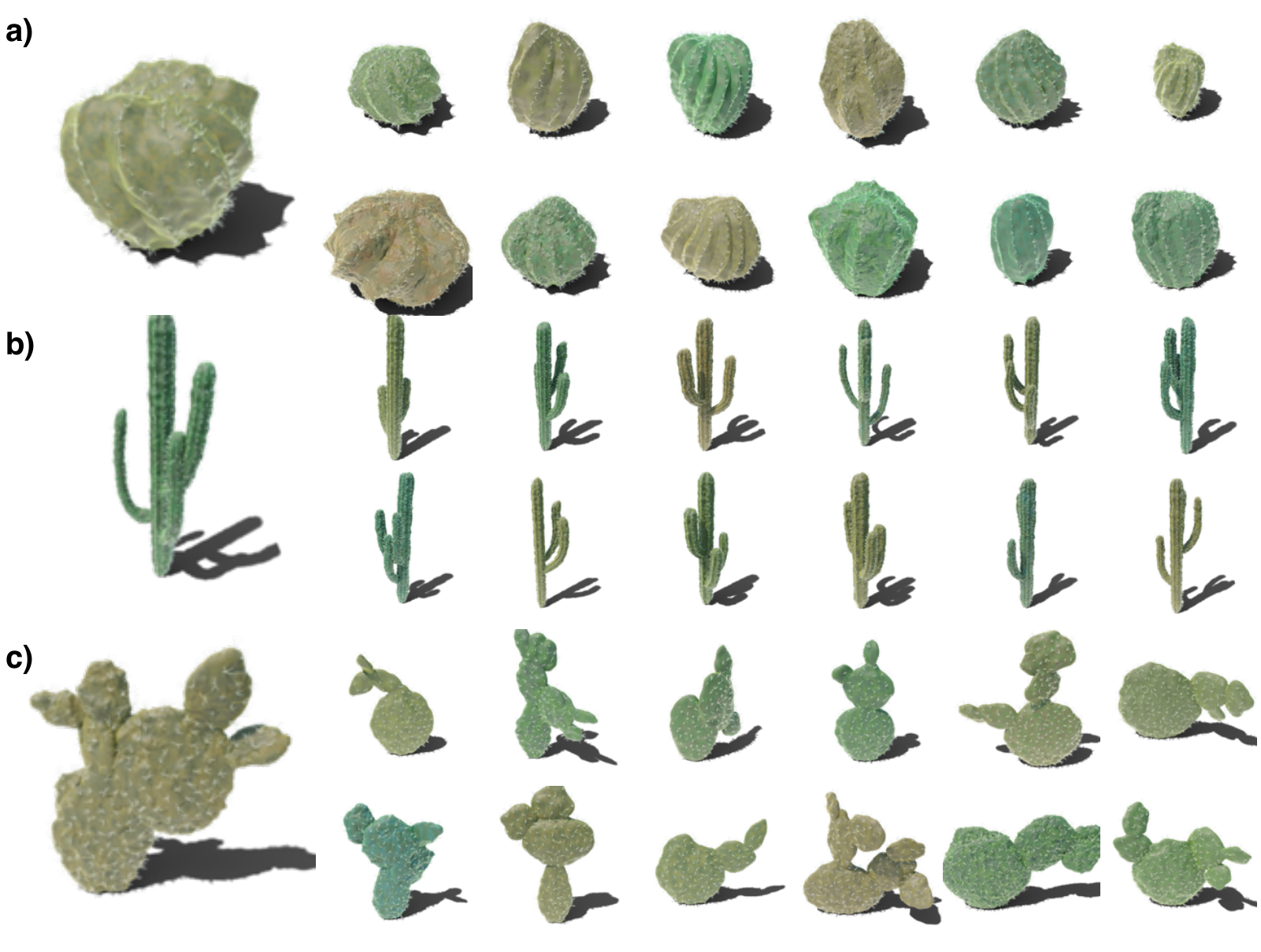}
    \caption{All classes of cacti included in \projectname{}. Each row contains one class of cactus: {\bf a)} Globular Cactus; {\bf b)} Columnar Cactus; {\bf c)} Pricky pear Cactus.}
    \label{thefig:cactus}
\end{figure*}
\subsubsection{Cactus}
\paragraph{Globular Cactus} is modeled after cactus from genus {\it Ferocactus}, as shown in \fig{thefig:cactus}a). It features a barrel-like base shape and tentacles growing from the pointy vertices of the cactus body. We implement globular cactus by first creating a star-like 2D mesh as its cross section. We then use geometry nodes to rotate, translate and scale it along the Z-axis at the same time, converting it to a 3D mesh. The rotation of the cross section mesh contributes to the desired tilt of the cactus body, and the scale determines the general shape of cactus. Finally the cactus is deformed and scale along the X and Y axis. For Globular Cactus, spikes are distributed on the the pointy vertices generated from the star and on the top most part of the cactus.

\paragraph{Columnar Cactus} is modeled after cactus from genus {\it Cereus}, as shown in \fig{thefig:cactus}b). It features an elongated body with a torch-like shape. We first generate the skeleton using our tree skeleton generation method. This time we choose a configuration with only two levels, both with a smaller momentum in path generation and a large drag towards the positive Z-axis that finally makes the cactus pointing up. From the cactus skeleton, we convert it into a 3D mesh with geometry nodes that moves a star mesh along all the splines in the tree-like skeleton, with the top end of each path having a smaller radius. For Columnar Cactus, spikes are distributed on the the pointy vertices generated from the star and on the top most part of the cactus.

\paragraph{Pricky pear Cactus} is modeled after cactus from genus {\it Opuntia}, as shown in \fig{thefig:cactus}c). It features pear-like cactus part extruded from another one. We create individual cactus parts similar to the one in Globular cactus. We it down in Y-direction and rotated it along the Z-axis so that it becomes almost planar and pear-shaped. These individual cactus parts are stacked on top of each other with an angle recursively to make the whole cactus branching like a tree. For Pricky Pear Cactus, we distribute spikes on both the front and the back faces of the cactus. 

\paragraph{Cactus spikes} After the main body of a cactus is created, we apply medium-frequency displacement on its surface and add the spike according to specifications. The low-poly spikes are made from several straight skeletons generated by the tree generation system, and are distributed on the selected areas of the cactus with some minimal distance between two instances.

\subsubsection{Fern}

\begin{figure*}
    \centering
    \includegraphics[width=\linewidth]{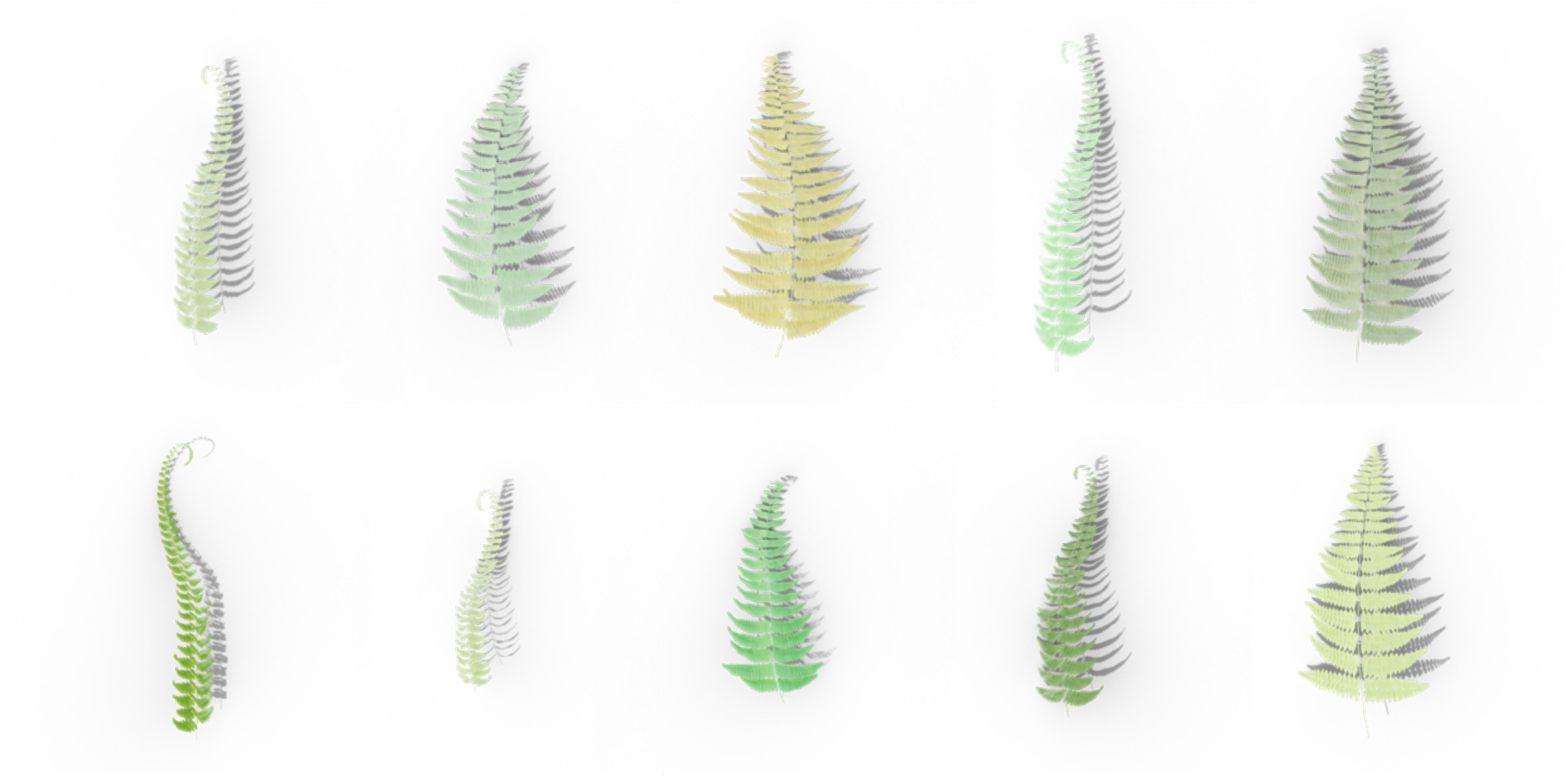}
    \caption{Assets of fern pinnae included in \projectname{}. A fern consists of a random number of pinnae in the same color with random orientations.}
    \label{thefig:pinnae}
\end{figure*}

 We create 2-pinnate fern (fern) in \projectname{}, as it is common in nature. Each fern is composed of a random number of pinnae with random orientations above the ground. Several instances of fern pinnae are illustrated in Fig.\ref{thefig:pinnae}.

\paragraph{Pinnae Composition} Each pinnae consists of a main stem and a random number of pinna attached on each side of the stem. The length of the pinnae (main stem) is controlled by a parameter \textit{age}. The main stem is first created with a mesh line with its points set along the z-axis. Then, the mesh line is randomly rotated w.r.t x, y axis around the top point to generate the curly look of a fern pinnae. The pinnae's z-axis rotation is slightly different with x, y axis rotation, as we use its curvature to represent the \textit{age} of the fern. In nature, young fern pinnae has more curly stem and grown-up fern pinnae is usually more stretched and flat. Therefore, we choose the scale of the pinnae's z-axis rotation inverse proportion to the \textit{age} of the pinnae. The external geometry of the main stem is a cylindar with its radius gradually shrinks to 0 at the top. Noticing that the bottom point is always set to world origin despite of the random rotations of the mesh line.

Before merging multiple pinnae into a fern, each pinnae is further curved along the z-axis towards the ground according to the desired orientation of the pinnae. We refer this as the gravitational rotation induced on the geomtery of pinnae. The scale of the gravitational rotation at each mesh line point is also proportion to its distance to the world origin, i.e., the longer the larger.

For 2-pinnate fern, the geometry of each pinna is similar to pinnae, i.e., a stem and leaves attached on each side. Similar to the main stem, we also curve the pinna stem randomly w.r.t x, y axis and inverse proportion to the \textit{age} w.r.t the z-axis. In our fern, the leaves are created with simple leaves-like geometry.

The whole pinnae is generated by adding leaf instances on pinna stem and then adding pinna instances on the main stem. The scale of each leaf instance is scaled to form a desired contour shape of the pinna, which is defined to grow linearly from tip to bottom with additional random noise. For pinna instances on the pinnae, we generate multiple distinct pinna versions and then randomly select one for each mesh line point. In this way, we can enough irregularity and asymmetricity on the pinnae. Furthermore, the pinna instances are also scaled according to the desired pinnae contour. In our asset, two contour modes are use. One grows linearly from top to bottom with additional random noise and the other grows linearly from top to $\frac{1}{6}$ from the bottom and then decrease linearly till the bottom of the pinnae.

Moreover, after all components are joined together, additional texture noise is added on the mesh to create more irregularities.

\paragraph{Fern Composition} Each fern is a mixture of pinnae with random orientations. In nature, these fern pinnae typically bend down towards the ground. We also add young fern pinnae standing rigidly in the center.

\subsubsection{Mushroom}
\begin{figure*}
    \centering
    \includegraphics[width=\linewidth]{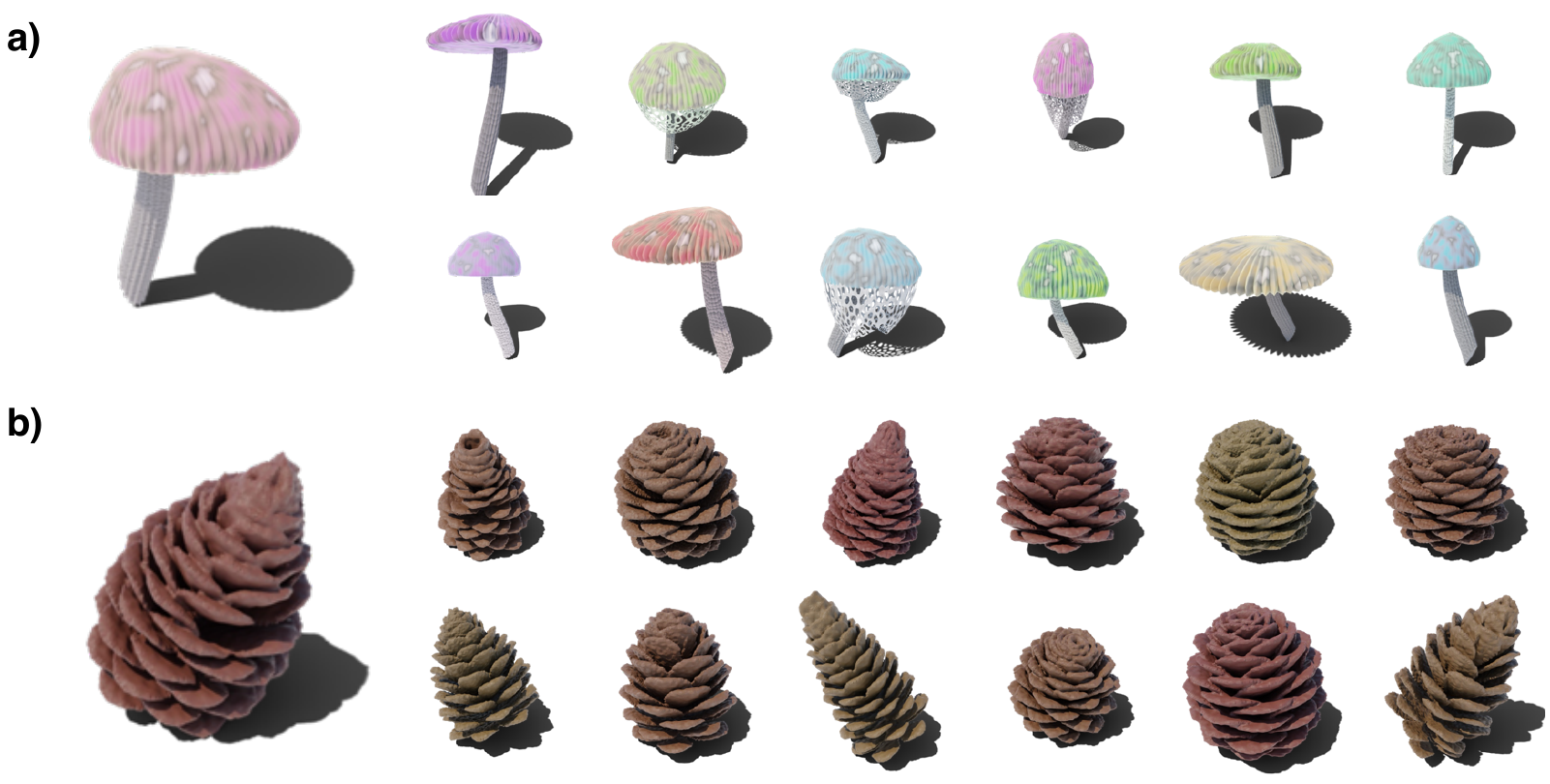}
    \caption{Mushrooms {\bf a)} and pinecones {\bf b)}.}
    \label{thefig:mushroom}
\end{figure*}

Mushrooms are modeled after real mushrooms from genus {\it Agaricus} and {\it Phallus}, as shown in \fig{thefig:mushroom} a). A mushroom is composed of its cap, its stem that supports the cap and optionally the skirt that grows from underneath the cap. The cap is made from moving a star-like mesh along the Z-axis with radius specified by multiple Bezier curves. The star-like mesh forms the pleats on the cap surface and gills underneath the caps. The stem is made from a Bezier curve skeleton and converted to a 3D mesh via geometry nodes, with its top end sticking to the cap at an angle. For the mushroom skirt, we create an invisible mesh $P$ underneath the cap that have a similar cross section as the cap. Following the same technique in the Brain Coral (See \sectapp{thetxt:reaction_diffusion}), we shrinkwrap a reaction-diffusion pattern from a icosphere $S$ onto $P$, which also mapped the $A$ field onto $P$. This time, we remove the vertices whose $A$ field is below a certain threshold from the mesh $P$, so $P$ would have honeycomb-like holes. The skirt is placed underneath the cap. Mushroom have similar material as corals, but have more white spots and lower roughness on its surface. Mushrooms are scattered on the terrain surface.

\subsubsection{Corals} Corals are marine invertebrates that live mostly on the seafloor, and are prevalent in underwater scenes. In \projectname{}, we provide a library of 8 templates for generating different classes of corals. Examples of individual coral classes are provided in \fig{thefig:coral}. We elaborate these templates for the main coral bodies as follows:

\begin{figure*}
    \centering
    \includegraphics[width=\linewidth]{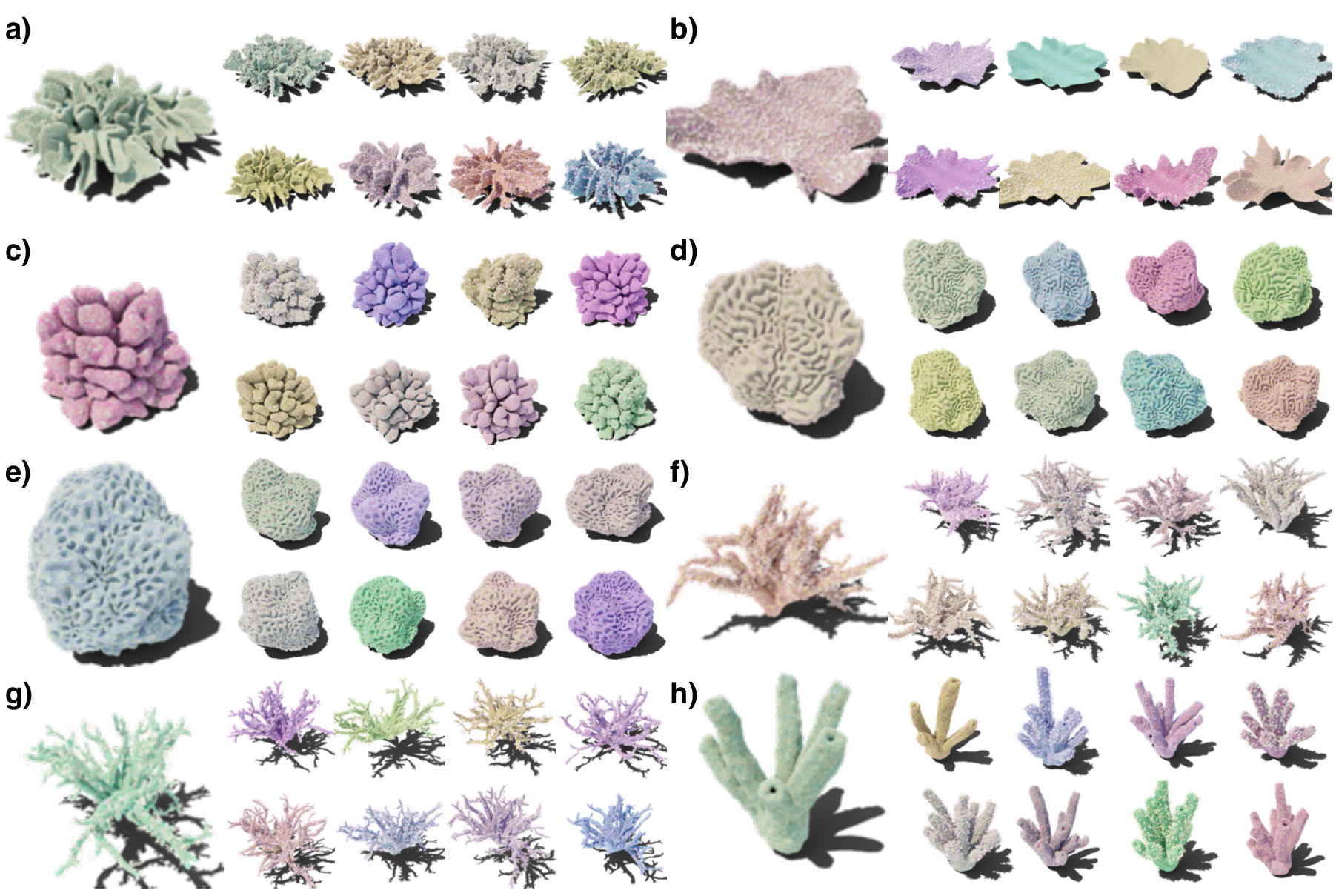}
    \caption{All classes of corals included in \projectname{}. Each block contains one class of coral: {\bf a)} Leather Coral; {\bf b)} Table Coral; {\bf c)} Cauliflower Coral; {\bf d)} Brain Coral; {\bf e)} Honeycomb Coral; {\bf f)} Bush Coral; {\bf g)} Twig Coral; {\bf h)} Tube Coral.}
    \label{thefig:coral}
\end{figure*}

\paragraph{Leather Coral}\label{thesec:diff_growth} is modeled after corals from genus {\it Sinularia}, as shown in \fig{thefig:coral}a). It features curvy surfaces that folds on it self. We implement Leather Coral using a iterative mesh generation technique named Differential Growth from \cite{diff_growth}. The mesh starts off as a mesh circle. At each iteration, a force is applied to all of its points. The force at each point is composed of an attraction force from its graph neighbors, a repulsive force from vertices that are close in 3D position, a global growth direction, and a noise vector. All of these forces can be specified by a set of parameters. Such force is applied on all vertices, and the vertices would be displaced by a distance proportional to the force. More concretely, for all vertices $v$
\begin{align*}
    \Delta {\bf x}_v = {\bf x}'_v -{\bf x}_v \propto f_{attr} +f_{rep} +f_{grow}+f_{noise}
\end{align*}

If the displacement has moved two vertices so that the edge connecting these two vertices has its length above a certain threshold, such edge would subdivided so that their lengths would fall below the threshold, creating new vertices on the edges. The aforementioned process defines a growing mesh when it is repeated for multiple iterations. In specific, for Leather Corals, we choose the parameters so that the noise force function and growth force function are large, and the growth iterations stop when there's 1k faces. To convert this mesh to the final coral mesh, we apply smooth and subsurface operation on the mesh,  followed by a solidify operation that gives the planar mesh some width, converting it into a 3D mesh. 

\paragraph{Table Coral} is modeled after corals from genus {\it Acropora}, as shown in \fig{thefig:coral}b). It features a flat table with curvy surfaces near the boundary. For Table Corals, we use the same Differential Growth method as the Leather Coral. However, we choose a different set of parameters for force application. More concretely, we use a larger repulsion force, apply less displacement on boundary vertices, and stop the process after there is 400 faces in the mesh. 

\paragraph{Cauliflower Coral} is modeled after corals from genus {\it Pocillopora}, as shown in \fig{thefig:coral}c). It features wart-like growth on its surface. We implement Cauliflower Coral after \cite{Kobayashi1993}'s simulation of dendritic crystal growth. In this growth simulation setup, we have two density fields $A$ and $B$ over 3D space. The simulation follows the following PDE:
\begin{align*}
    m&= \alpha \arctan{(\gamma(T-b)}/\pi)\\
    \frac{\partial A}{\partial t} &= \varepsilon^2\nabla A +A(1-A)(A-1/2+m)/\tau\\
    \frac{\partial B}{\partial t}&=\nabla B+ k\frac{\partial A}{\partial t} \\
\end{align*}
where $\alpha,\gamma,T,\varepsilon,\tau, k$ are pre-specified parameters. We run this PDE simulation on a 3D grid space with forward Euler method for 800 iterations. The resulting density map $A$, is used to generate the 3D mesh for the coral via the marching cube mesh conversion method.

\paragraph{Brain Coral}\label{thetxt:reaction_diffusion} is modeled after corals from genus {\it Diploria}, as shown in \fig{thefig:coral}d). It features groovy surface and intricate patterns on the surface of the coral. We first create the surface texture with reaction-diffusion system simulation. In particular, we start off from a mesh icosphere, and run Gray-Scott reaction-diffusion model on its vertex graph, where edges in the mesh are the edges of the graph. This simulation has two fields $A,B$ on individual vertices, following the equation of:

\begin{align*}
    \frac{\partial A}{\partial t}&=r_A\nabla A-A^2B+f(1-A)\\
    \frac{\partial B}{\partial t}&=r_B\nabla B-A^2B-(f+k)B\\
\end{align*}

where $r_A,r_B,f,k$ are pre-specified parameters. After 1k iterations, the field $A$ is stored to the mesh sphere $S$. Then we build another polygon mesh $P$, which follows simple deformation by geometry nodes. We apply shrinkwrap object modifier from $S$ to $P$, which also maps the field $A$ onto mesh $P$. The shrinkwrap object modifiers `wraps' the surface of $A$ onto $P$ by finding the projected points of $A$ along $A$'s normal direction. We displace $P$'s surface using the projected field $A$, which forms the grooves on the mesh surface. For Brain Corals, we choose the parameters so that $f=\sqrt{k}/2-k$ for some specific $k$, i.e. the kill rate and feed rate of system are on the saddle-node bifurcation boundary, so that the surface is groovy.

\paragraph{Honeycomb Coral} is modeled after corals from genus {\it Favia} or {\it Mussismilia}, as shown in \fig{thefig:coral}e). It features honeycomb-shaped holes on the surface of the coral. We use the same Gray-Scott reaction-diffusion model \cite{gray1994chemical} as the Brain Corals. We choose the parameters so that $f=\sqrt{k}/2-k-0.001$ for some specific $k$, i.e. the kill rate and the feed rate of the system are between the saddle-node bifurcation boundary and the Hopf bifurcation boundary, which yields the honeycomb-shaped holes.

\paragraph{Bush Coral} is modeled after corals from genus {\it Acropora}, which includes the Staghorn coral, as shown in \fig{thefig:coral}f). We implement bush coral using our tree skeleton generator as discussed. In particular, the skeleton of tree coral has three levels of configuration, with each level of configuration specifying how a branch would grow in terms of directions and length, as well as where new branches would emerge. After the tree skeleton is generated, we convert the 1D skeleton to 3D mesh by specifying radius at each vertex of the skeleton, with the vertices closer to endpoints having a smaller radius.

\paragraph{Twig Coral} is modeled after corals from genus {\it Oculina}, as shown in \fig{thefig:coral}g). We use the same tree skeleton generator as in Bush Corals. We choose a separate set of parameters so that Twig Corals are more low-lying and less directional than Bush Corals.

\paragraph{Tube Coral} is modeled after not corals, but sponges from genus {\it	Aplysina}, which none-the-less lives in the same habitat as corals, as shown in \fig{thefig:coral}h). We first generate the base mesh as the dual mesh of a deformed icosphere, whose faces have between 5-6 vertices. The optionally extrude some its upward facing faces along the direction that is approximately along the positive Z-axis. This types of extrusion happens multiple times with different multiple extrusion length. The extruded mesh will have its topmost face removed and will be applied with the solidify object modifier so that the mesh would become a hollow tube.

\paragraph{Coral tentacles} After the main body of a coral is created, we add a high frequency noise onto the coral surface. We then add tentacles to the coral mesh. Each tentacle has a low-poly mesh generated from the tree generator system that sprawls outside the coral body. Tentacles are distributed on certain parts of the coral body, like on top-facing surfaces or outermost surfaces.

\subsubsection{Other sea plants}
\begin{figure*}
    \centering
    \includegraphics[width=\linewidth]{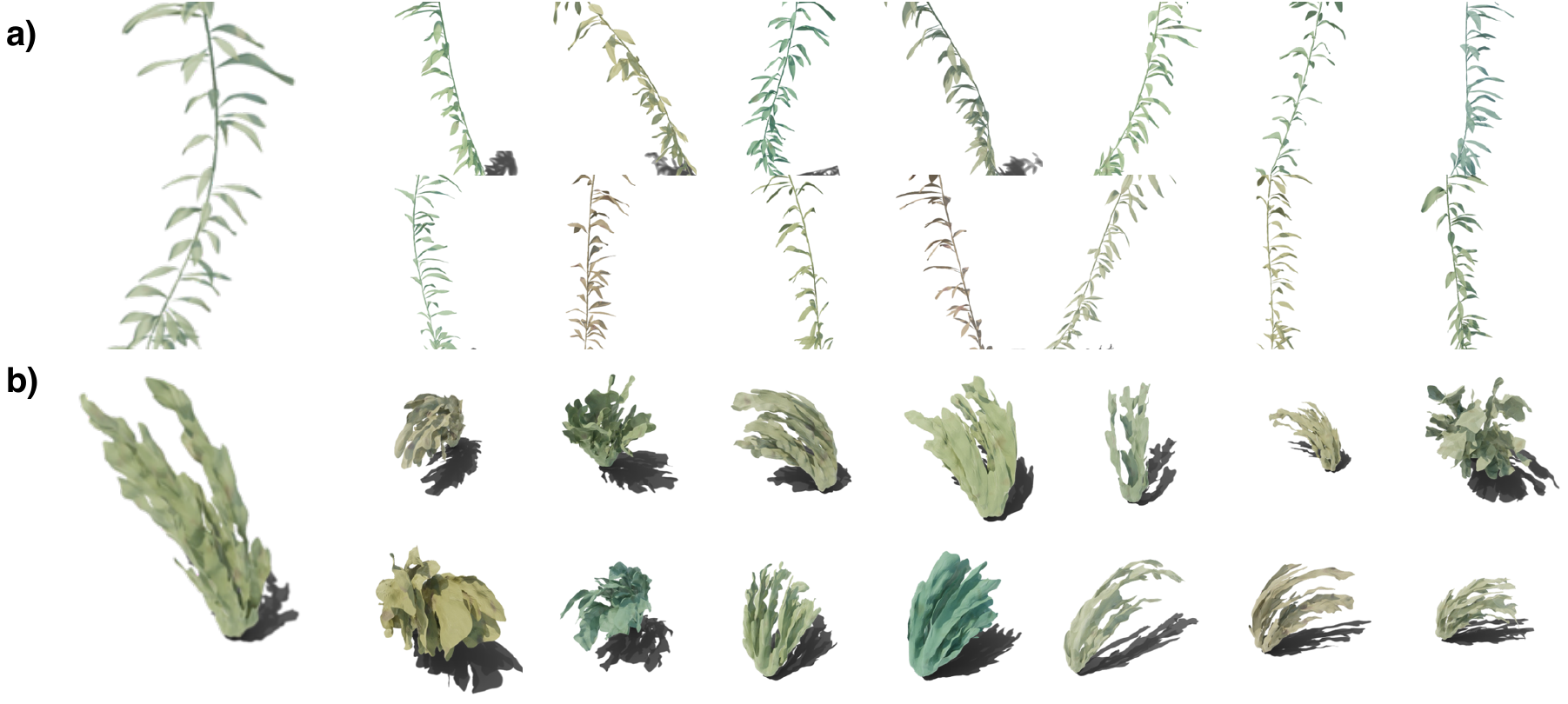}
    \caption{Kelps {\bf a)} and seaweeds {\bf b)} examples, each occupying one row.}
    \label{thefig:kelps}
\end{figure*}

Besides corals, we also provide assets of other sea plants, like kelps and seaweeds. Both kelps and seaweeds observe the global oceanic current, which is unique for the entire scene. We show examples of kelps and seaweeds in \fig{thefig:kelps}.

\paragraph{Kelps} Kelps are large brown algae that lives in shallow waters, as shown in \fig{thefig:kelps}a). To build kelps, we first build meshes for individual kelp leaves. To do so, we first create a planar mesh which is bounded by two sinusoid functions from two sides. It is then deformed along the Z axis an subsurfaced and make a wavy shape. For the kelp stem, we first plot a Bezier curve with its control points following a Brownian process with drift towards the sum of the positive Z direction and the oceanic current drift. The curve is turned into a mesh with a small radius, and this becomes the mesh of the kelp stem. Along the stem we scatter points with fixed intervals, where we place kelp leaves along its normal direction. Finally, we rotate the kelp leaves towards somewhere lower than the oceanic current vector, so that they are affected by both the oceanic current and gravity.

\paragraph{Seaweeds} Seaweeds are another class of marine algae that are shorter than kelps, as shown in \fig{thefig:kelps}b). We create seaweed assets using the Differential Growth model as in the Leather Coral assets (See Section \sectapp{thesec:diff_growth}). In particular, we choose the parameters so that it has a large growth force towards the positive Z axis. We apply smoothing and subsurfacing to the 2D mesh and solidify it into a 3D mesh. Then the seaweed is bent towards the direction of the oceanic current by a varying degree using the simple deform object modifier. Seaweeds are scattered on the surface of the terrain mesh. 

\subsection{Surface Scatters}

\begin{figure*}
    \centering
    \includegraphics[width=\linewidth]{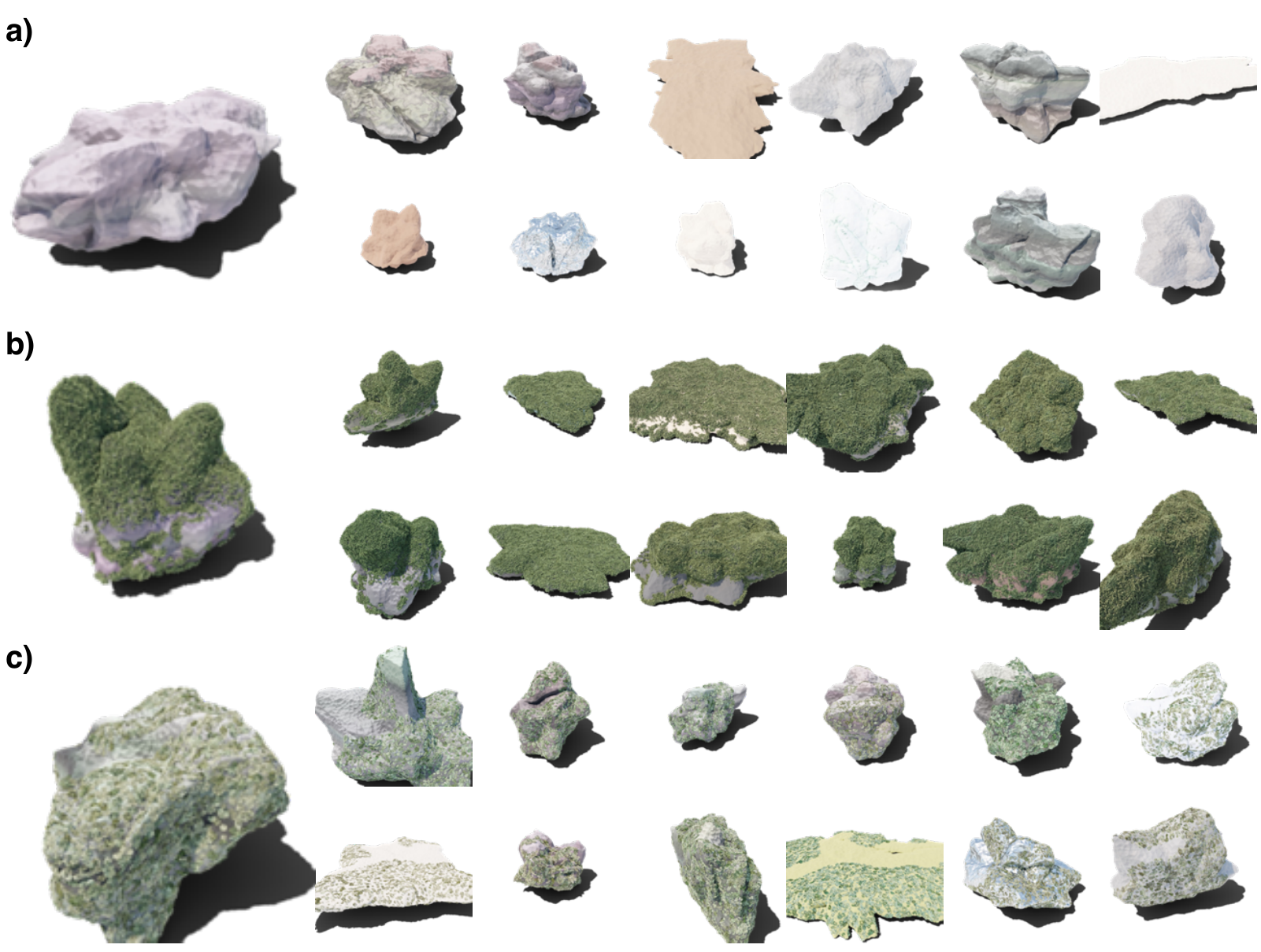}
    \caption{Boulder assets with different rock cover surfaces. In particular, each row of boulders are under {\bf a)} no surface; {\bf b)} moss surface; {\bf \c)} lichen surface.}
    \label{thefig:boulders}
\end{figure*}
\paragraph{Moss} forms dense green clumps or mats on a flat surface. We create individual moss using Bezier curves as skeletons, and then turning them into 3D mesh. We distribute individual moss instance on the boulder surfaces with the Z axis aligned at an angle with the surface normal. Such angle of rotation is guided by a Musgrave texture so that mosses in a neighbourhood have similar rotations. Moss instances have shaders that is built from a mixture of several yellowish to greenish colors, with the color variation determined by a Musgrave texture. Moss can grow on three different places of the boulder, from the faces with a higher Z coordinate than a threshold, or those whose face normal is within a threshold of the positive Z axis, or near edges where there are concave edges.

\paragraph{Lichen} is a composite organism that arises from algae that forms a mat on rock surfaces. Individual lichens are made from Differential Growth specified in \sectapp{thesec:diff_growth}. The color of lichens comes from a mixture of yellowish-greenish color and white colors, with the mixing ratio guided by a Musgrave texture. Lichen can grow on either boulders or on tree trunks. On boulders, lichens are distributed on all faces of the boulder, or the lower portions of tree trunks with a minimal distance between two instances.  

\paragraph{Slime mold} are organisms shaped like gelatinous slime that lives on decaying plant materials. We first designate around 20 initial seedling vertices where the slime mold can grow from, and assign a random weight proportional to the local convexity to all edges in the chosen area of the mesh. Then we use geometry nodes to compute the shortest path from each vertex to any of the seedling vertices, and connect these shortest path. These shortest paths from the skeleton of the slime mold. Slime mold only grows on the lower portion of tree trunks

\begin{figure*}
    \centering
    \includegraphics[width=\linewidth]{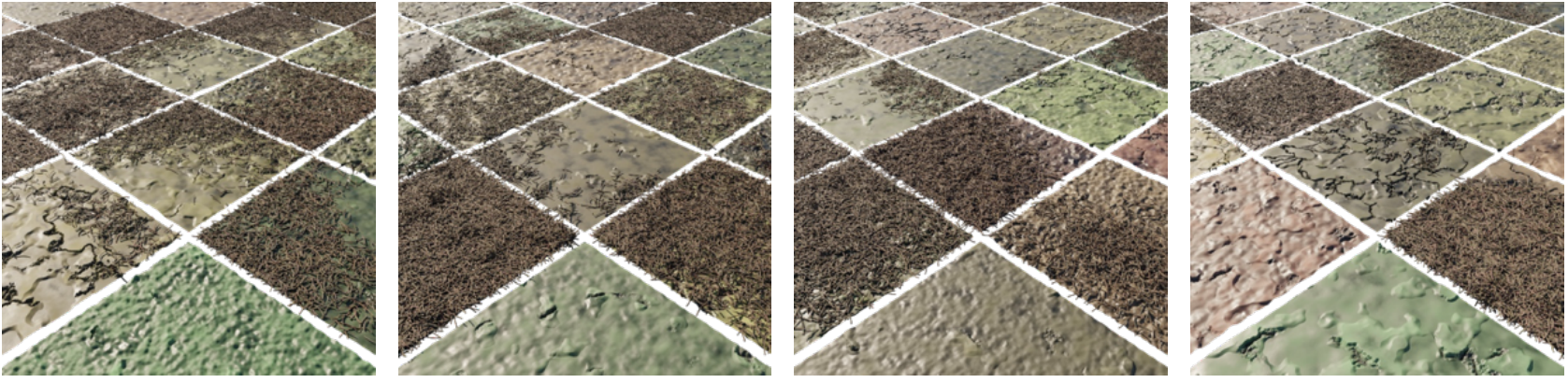}
    \caption{Pine needles scattered onto the ground with varying density.}
    \label{thefig:pine_needle}
\end{figure*}

\paragraph{Pine needles}
Pine needles are the leaves of the pine tree that have fallen onto the ground, as shown in \fig{thefig:pine_needle}. Pine needles are made from a segment of a ellipsoid and are typically of brownish and greenish colors. Pine needles are scattered on certain parts of the terrain surface based on noise texture. Pine needles are scattered on different heights so that pine needles of a certain color are above pine needles of another color.

\subsection{Marine Invertebrates}

\subsubsection{Mollusk}
\begin{figure*}
    \centering
    \includegraphics[width=\linewidth]{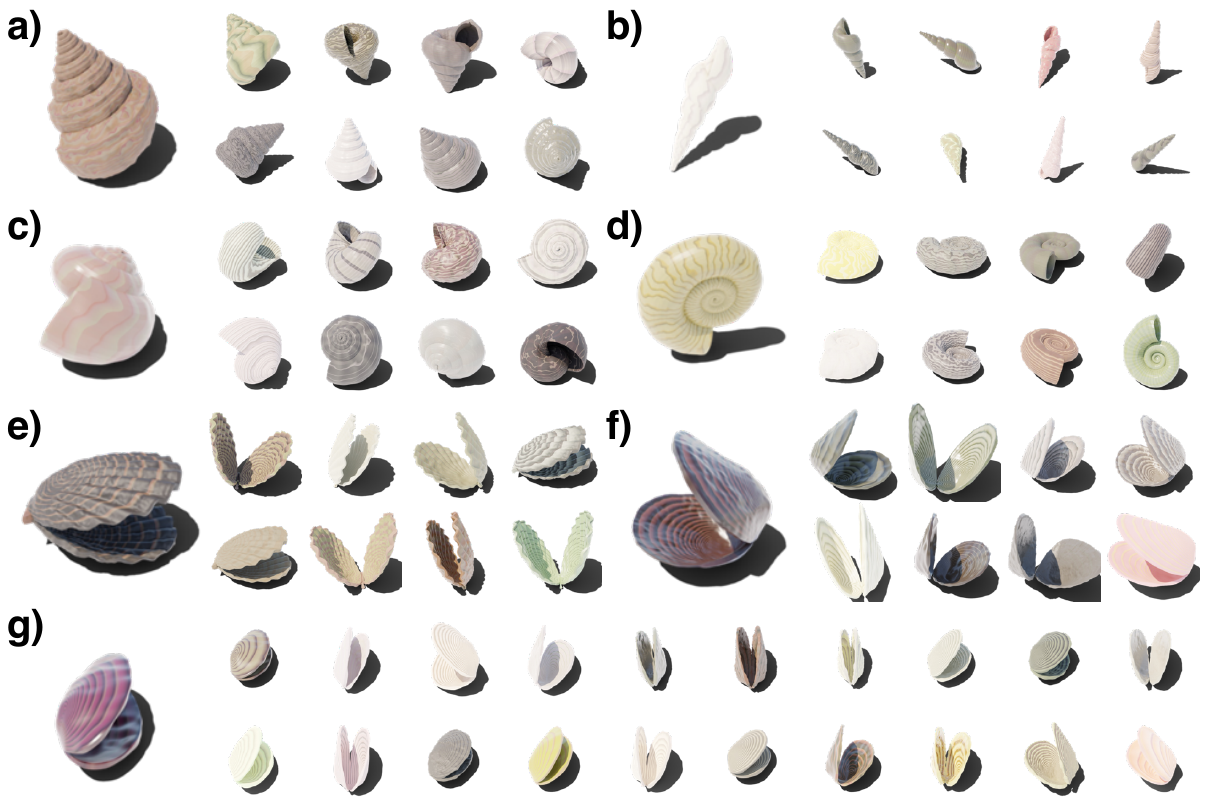}
    \caption{Different classes of mollusks, with each block representing a class of mollusk: {\bf a)} Conch, {\bf b)} Auger, {\bf c)} Volute, {\bf d)} Nautilus, {\bf e)} Scallop, {\bf f)} Clam, {\bf g)} Mussel.}
    \label{thefig:mollusk}
\end{figure*}

Mollusk is the collection of animals that includes most snails and shells, as shown in \fig{thefig:mollusk}.

\paragraph{Snails} is modeled after animals mostly from the class {\it Gastropoda}. It features snails of following shape: Conch (\fig{thefig:mollusk} a), from family {\it Strombidae}, Auger (\fig{thefig:mollusk} b), from family {\it Terebridae}), Volute (\fig{thefig:mollusk} c),from family {\it Volutidae}) and Nautilus (\fig{thefig:mollusk} d),from a different class {\it Cephalopod}). All these class of animals share the commonality that they live in a shell that grows and rotates at a constant angle as the soft body of the animal grows, which can be modeled using the array object modifier in Blender. We first build the cross section of these snails from the interpolation of a start and a ellipsoid, which gives the space for the soft body to live in, as well as the pointy spikes on some snails' surface. Then we apply the array object modifier onto the cross section so that it is rotated with a constant angle, scaled at a constant ratio, and displaced at a constant interval. These series of cross section can uniquely define the cross section at each stage of the growth, with which we can bridge the edge loops to finally form a 3D mesh for the whole snail. Parameters in this process includes the displacement of cross section along and orthogonal to the axis, the total number cycles of rotations, the ratio that the scale of cross section shrinks, and the other parameters with regard to the shape of the base cross section. The parameters are set differently for individual class of snails: Conch and Auger have large displacement along their axis while Volute and Nautilus have almost none; Conch is made from more spiky cross section mesh, and has less overlapping chambers than Auger; Nautilus has a faster-shrinking cross section, almost no displacement along the axis, and have less over lapping chambers than Volute.

\paragraph{Shells} is modeled after animals mostly from the class {\it Bivalvia}. It features shells of the following shape: Scallop (\fig{thefig:mollusk} e), from family {\it Pectinidae}), Clam (\fig{thefig:mollusk} f), from family {\it Veneridae} and Mussel (\fig{thefig:mollusk} g), from family {\it Mytilidae}). These animals share the commonality that they are covered by two symmetrical shells joining at a point that folds around the soft body of the individual, with both shells growing gradually from inside the shell. To build assets for these shells, we first generate individual shells. We first create a mesh circle and select one point on the circle as its origin. For all points on the circle, we scale its distance from the origin, with the ratio determined by the direction from the origin to the target point. Then we choose a point above the XY-plane, and interpolate between the previous mesh and the newly selected point. The interpolation ratio is determined by a point's distance to the boundary, so that the boundary points are the XY plane, which creates the convex shape of an individual shell. We mirror the shell at an angle, and now we have the 3D mesh of the shell. We have different designs for distinct class of shells: Scallops are given a wavy pattern depending on vertices' direction to the origin, and have girdles near the origin; Clams are the most basic shells with no alternations; Mussels are made from shells that are similar to ellipsoids with large eccentricities. 

\paragraph{Mollusk material} Both snails and shells grows along a certain direction and leaves a changing color pattern along its growing direction. We define a 2D coordinate $(U,V)$ the mollusks surface for mapping textures, whether $U$ is the growth direction and $V$ is orthogonal to it. For snails, $U$ is the direction of displacement for its cross section mesh, and $V$ is along the boundary of the cross section mesh. For shells, $U$ is the direction from the center of the shell to the boundary of shell, and $V$ is along the boundary of the shell. We design a saw-like wavy texture that progresses along either $U$ or $V$ directions, which creates interchanging color patterns along or orthogonal to the direction of growth. Both snails and shells are given a low-frequency surface displacement, and scatter on the terrain mesh.

\subsubsection{Other marine invertebrates}
\paragraph{Jellyfish} is composed of its cap and tentacles. For the cap, we first generate two  mesh uv spheres with one above another, then scale and deform both spheres. Then we subtract the sphere below from the the sphere above, creating the cap with two surfaces, one facing towards the positive Z axis and another facing towards the negative Z axis. Tentacles are made from ribbons along the Z axis, which are later deformed along the X axis and tapered, and finally rotated around the Z axis. Tentacles of varying sizes are placed around the lower surface of the cap. The jellyfish shader is made from a mixture of colored emission, a principled BSDF with transmission, and a colored transparent shader, whose mixing ratio is guided by Fresnel coefficients. We use a more transparent material for outer surface of the cap and shorter tentacles, which are more peripheral parts of the body, and a more opaque material for inner surface of the cap and longer tentacles, which are core organs of a jellyfish. Jellyfish are scattered with a random offset above the ground mesh.

\paragraph{Urchin} is a spiny, globular animal living on the sea floor. For modeling urchin assets, we first start with an icosphere. For each face of the icosphere, we extrude it outwards by a small distance, scale it down, and extrude it inwards to form the girdle. We then extrude the faces outwards by a varying but large distance, and scale it down to zero so that they form the spikes that ground on the urchin. The bases of an urchin are from a darker color and the spikes are from a lighter color between purple and yellow, with the girdle's color somewhere in between. 

\subsection{Creatures}
\label{thesec:creature}

\subsubsection{Creature Construction}

Each creature genome is a tree of parameters, with nodes specifying parts and edges specifying attachment.

Each node contains a dictionary of named input parameters for one of our part templates (Sec. \ref{thesec:supp_creature_parts}). We compute all parts in isolation before proceeding to attach them. This part template must produce 1) a mesh and 2) a skeleton line. The skeleton line is a 3D parametric curve specifying the center line of the part, and is used for attachment and rigging. Requiring part templates to produce a center line is not a limitation - for NURBS and most node-graph parts it is trivial to obtain. Additionally, this output can be omitted for any part not intended to have further children attached to it. Each part template may also produce additional metadata for use in the animation and material stages. 

Each edge contains a coordinate $(u, v, r)$ to determine the attachment location. $(u, v) \in [0, 1]^2$ specifies a location on the parent mesh's surface. For arbitrary meshes, this is computed by travelling $u$ percent of the way along the parent's skeleton and raycasting orthogonally to it, with angle $360^{\circ}*v$. If the parent part is a NURBS, one can instead query $(u, v)$ on it's parametric surface. Finally, we use $r$ to interpolate between the found surface point, and the corresponding skeleton point, which has the effect of controlling how close to the surface the part is mounted.

Finally, each edge specifies a relative rotation used to pose the part. Optionally, this rotation can be specified relative to the parent part's skeleton tangent, or the attachment surface normal. 

\subsubsection{Creature Animation}
As an optional extra output, each part template may specify where and how it articulates, by specifying some number of \textit{joints}. Each \textit{joint} provides specifies rotation constraints as min/max euler angles, and a parameter $t \in [0, 1]$, specifying how far along the skeleton curve it lies. If a part template specifies no joints, it's skeleton is assumed to be rigid, with only a single joint at $t=0$. We then create animation bones spanning between all the joints, and insert additional bones to span between parts.

Any joint may also be tagged as a named inverse kinematics (IK) target, which are automatically instantiated. These targets provide intuitive control of creature pose - a user or program can translate / rotate them to specify the pose of any tagged bones (typically the head, feet, shoulders, hips and tail), and inverse kinematics will solve for all remaining rotations to produce a good pose. 

We provide simple walk, run and swim animations for each creature. We procedurally generate these by constructing parametric curves for each IK target. These looped curves specify the position of each foot as a function of time. They can be elongated or scaled in height to achieve a gallop or trot, or even rotated to achieve a crab walk or walking in reverse. Once the paths are determined, we further adjust gait by choosing overall RPM, and offsets for how synchronized each foot will be in it's revolution.
 
\subsubsection{Genome Templates}

\begin{figure*}
    \centering
    \includegraphics[width=\linewidth]{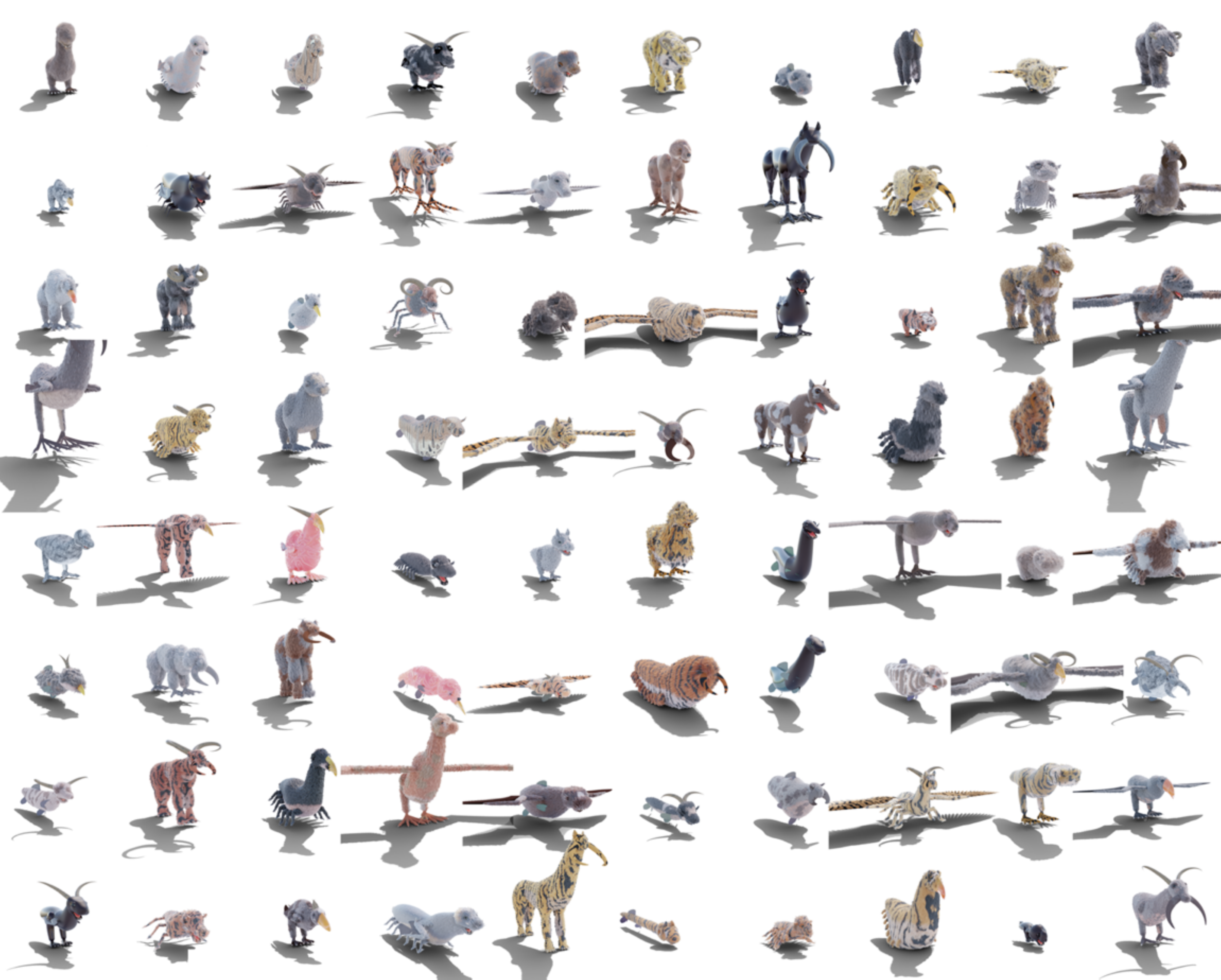}
    \caption{We provide procedural rules to combine all available creature parts, resulting in diverse fantastical combinations. Here we show a random, non-cherry-picked sample of 80 creatures. Despite diverse limb and body plans, all creatures are functionally plausible and possess realistic fur and materials.}
    \label{thefig:compositional_creatures}
\end{figure*}

In the main paper, we show the results of our realistic \textit{Carnivore, Herbivore, Bird, Beetle} and \textit{Fish} templates. These templates contain procedural rules to determine tree structure by adding legs, feet, head and appropriate details to create a realistic creature. Each template specifies distributions over all attachment parameters specified above, which provides additional diversity ontop of that of the parts themselves. Tree topology is mostly fixed for each template, although some elements like the quantity of insect legs and presence of horns or fish fins are random.

Our creature system is modular by design, and supports infinite combinations besides the realistic ones provided above. For example, we can randomly combine various creature bodies, heads and locomotion types to form a diverse array of creatures shown in Fig \ref{thefig:compositional_creatures}. As we continue to implement more independent creature parts and templates, the possible combinations of these random creature genomes will exponentially increase. 

Creature genomes also support a semi-continuous interpolation operation. Interpolation is trivial for creatures with identical tree structure and part types - one can perform part-wise linear interpolation of node and edge parameters. When tree topology or part types don't match, we recursively compute a matching for each node's children which minimizes the difference of edge attachment parameters, then perform linear interpolation on any node parameters with matching names. To interpolate between a present and missing genome node, we scale the part down from its original size to 0, which results in small vestigial arms or tails on the intermediate creatures. When part types do not align exactly, there is a discrete transition halfway through interpolation, so interpolation is not continuous in all cases.

\subsubsection{Creature Parts}
\label{thesec:supp_creature_parts}

\paragraph{NURBS Parameterization}

Many of our creature body and head templates are comprised of non-uniform rational B-splines (NURBS). NURBS are the generalized 3D analog of a B\'ezier curve. In order to form a closed shape, we pinch each NURBS surface closed at its ends, and loop it's handles in the V direction to form a closed cylinder as a starting point. We set the U and V knot-vectors to be \textit{Pinned Uniform}, and instead rely on densely-spaced or coincident handles to create sharp edges where necessary. 

By default, a NURBS cylinder is represented as an $N \times M$ array of 3D handle locations. We find the space of all NURBS handle configurations too high dimensional and unstructured to randomize directly. Adding Gaussian noise to handle locations produces lumpy, unrealistic creatures, and is unlikely to coordinate to create phenomena like bent limbs or widened midsections. Instead, we randomize under a factored representation. Specifically, we start with a $Nx3$ array of radii and relative angles, which stores a center line for the part as polar-coordinate offsets. Accumulating these produces an $Nx3$ skeleton line. We arrange N \textit{profile shapes} around this center line, each stored as M 3D points centered about the origin. This representation has just as many parameters as the original, but randomizing it produces better results. Adding noise to the polar skeleton angles and radii produces macro-scale changes in body or head shape, and multiplying the profiles by random scalars can easily change the radius or cross section, independent of where along the skeleton that profile is located. 

As a starting point for this randomization, we determined skeleton and profile values which visually match a large number of reference animal photos, including \textit{Ducks, Gulls, Robins, Cheetahs, Housecats, Tigers, Wolves, Bluefish, Crappie Fish, Eels, Pickernel Fish, Pufferfish, Spadefish, Cows, Giraffe, Llama} and \textit{Goats}. Rather than make a discrete choice of which mean values to use, we take a random convex combination of all values of a certain category, before applying the extensive randomization described above. 

\paragraph{Horns} are modeled by a transpiled Blender node graph. The 2D shapes of horns are based on spiral geometry node, supporting adjustable rotation, start radius, end radius and height. The 3D meshes then are constructed from 2D shapes by curve-to-mesh node, along with density-adjustable depth-adjustable ridges. Along with the model, we provide three parameter sets to create goats, gazelles and bulls' horn templates.

\paragraph{Hooves} are modeled by NURBS. We start with a cylinder, distributing control points on the side surface evenly. For the upper part of the cylinder, we scale it down and make it tilted to the negative X axis, which makes its shape closed to the horseshoe. The model also has an option to move some control points toward the origin, in order to create cloven hooves for goats and bison. Along with the model, we provide two parameter sets to create horses' and goats' horns templates.

\paragraph{Beaks} are modeled by NURBS. Bird beaks are composed of two jaws, generally known as the upper mandible and lower mandible. The upper part starts with a half cone. We use the exponential curve instead of the linear curve of the side surface of the cone to obtain the natural beak shape. The model also has parameters that control how much the tip of the beak hooks and how much the middle and bottom of the beak bulge, to cover different types of beaks. The lower part is modeled by the same model of the upper part with reversed Z coordinates. Along with the model, we provide four parameter sets to create eagles, herons, ducks and sparrows' beaks templates.

\paragraph{Node-Graph Creature Parts}

All part templates besides those mentioned above are implemented as node-graphs. We provide an extensive library of node-groups to ease the construction of creature parts. The majority of these involve placement and querying of parameterized tubes, which we use to build muscular legs, arms and head parts. For example, our \textit{Tiger Head} and \textit{Quadruped Leg} templates contain nodes to construct the main central form of each part, followed by placement of several tubes along their length to create muscles and detailed forms. This results in a randomizable representation of face and arm musculature, which produces the detailed carnivore heads and legs shown in the main paper. These node-graph tools can also be layered ontop of NURBS as a base representation.

%% file: texts/18-dof-tables.tex
\begin{table*}
\centering
\footnotesize
\begin{tabular}{p{0.2 \textwidth}|l|p{0.7\linewidth}} 
\toprule
\multirow{2}{*}{Material Generators} & Interpretable & \multirow{2}{*}{Named Parameters} \\
& DOF & \\
\midrule
Mountain & 2 & Noise Scale, Cracks Scale \\
Sand & 5 & Color Brightness, Wave Scale, Wave Distortion, Noise Scale, Noise Detail \\
Cobblestone & 13 & Stone Scale, Uniformity, Depth, Crack Width, Stone Colors (5), Mapping Positions (2), Roughness \\
Dirt & 9 & Low Freq. Bump Size, Low Freq Bump Height, Crack Density, Crack Scale, Crack Width, Color1, Color2, Noise Detail, Noise Dimension \\
Chunky Rock & 4 & Chunk Scale, Chunk Detail, Color1, Color2 \\
Glowing & 1 & Color1 \\
Granite & 6 & Speckle Scale, Color1, Color2, Speckle Color1, Speckle Color2, Speckle Color 3 \\
Ice & 7 & Color, Roughness, Distortion, Detail, Uneven Percent, Transmission, IOR \\
Mud & 12 & Wetness, Large Bump Scale, Small Bump Scale, Puddle Depth, Percent water, Puddle Noise Distortion, Puddle Noise Detail, Color1, Color2, Color3, WaterColor1, WaterColor2 \\
Rock & 0 &  \\
Sandstone & 18 & Ridge Polynomial (2), Ridge Density, Ridge High Freq., Ridge Noise Mag., Ridge Noise Scale, Ridge Disp:Offset Magnitude, Roughness, Crack Magnitude (2), Crack Scale, Color1, Color2, Dark Patch Percentages (3), Micro Bump Scale, Micro Bump Magnitude \\
Snow & 3 & Average Roughness, Grain Scale, Subsurface Scattering \\
Soil & 10 & Pebble Sizes (2), Pebble Noise Magnitudes, Pebble Roundness, Pebble Amounts, Voronoi Scale, Voronoi Mag., Base Colors (2), Darkness Ratio \\
Stone & 10 & Rock Scale, Rock Deepness (2), Noise Detail, Noise Roughness, Crack Scale, Crack Width, Color1, Color2, Roughness \\
Aluminium & 2 & Bump Offset, XY Ratio \\
\midrule
Fire & 2 & Blackbody Intensity, Smoke Density \\
Smoke & 2 & Color, Density \\
Ocean & 5 & Wave Scale, Choppiness, Foam, Main Color, Cloudiness \\
Lava & 10 & Color, Rock Roughness, Amount of Rock, Lava Emission, Min Lava Temp., Max Lava Temp., Voronoi Noise, Turbulence, Wave Scale, Perlin Noise \\
Surface water & 5 & Color, Scale, Detail, Lacunarity, Height \\
Water & 6 & Ripple Scale, Detail, Ripple Height, Noise Dimension, Lacunarity, Color \\
Waterfall & 3 & Color, Foam Color, Foam Density \\
\midrule
Bark & 9 & Displacement Scale, Z Noise Scale, Z Noise Amount, Z Multiplier, Primary Voronoi Scale, Primary Voronoi Randomness, Secondary Voronoi Mix Weight, Secondary Voronoi Scale, Color \\
Bark Birch & 5 & Noise Scale (2), Noise Detail (2), Displacement Scale,  \\
Greenery & 13 & Color Noise (3), Roughness Noise (3), Roughness Min/Max (2), Translucence Noise (3), Translucence Min/Max (2) \\
Wood & 3 & Scale, XY Ratio, Offset \\
Grass & 6 & Wave Scale, Wave Distortion, Musgrave Scale, Musgrave Distortion, Roughness Min/Max (2), Translucence \\
Leaf & 2 & Base Color, Vein Color \\
Flower & 5 & Diffuse Color, Translucent Color, Translucence, Center Colors (2), Center Color Coeff. \\
Coral Shader & 5 & Bright Color, Dark Color, Light Color, Fresnel Color, Musgrave Scale \\
Slime Mold & 7 & Edge Weight, Spline Parameter Cutoff, Seedlings Count, Min Distance, Bright Color, Dark Color, Musgrave Scale \\
Lichen, Moss & 6 & Bright Color, Dark Color, Musgrave Scale, Density, Min Distance, Instance Scale \\
\midrule
Bird & 7 & Bird Type, Head Ratio, Stripe Width, Stripe Noise, Neck Ratio, Color1, Color2. \\
Bone & 3 & Bump Scale, Bump Frequency, Bump Offset. \\
Chitin & 3 & Boundary Width, Noise Weight, Thorax Size. \\
Horn & 8 & Noise Scales (2), Noise Details (2), Mapping Control Points (4) \\
Reptile Brown & 3 & Circle Scale, Circle Boundary, Noise \\
Fish Body & 7 & Scale Size, Scale Noise, Fish Type, Scale Offset, Color1 Ratio, Color2 Ratio, Noise \\
Fish Fin & 7 & Offset Z, Offset Y, Shape, Bump Noise, Fin Type, Bump Weight, Transparency \\
Giraffe & 4 & Scale, Noise, Circle Width, Belly \\
Reptile & 3 & Scale, Offset, Noise \\
Reptile Gray & 2 & Noise1, Noise2 \\
Reptile 2-Color & 2 & Color1, Color2 \\
Scale & 3 & Scale Size, Scale Noise, Scale Rotation \\
Slimy & 2 & Scale, Offset \\
Spot Sparse & 3 & Spot Scale, Color1, Color2 \\
3-Color Spots & 2 & Spot1 Ratio, Spot2 Ratio \\
Tiger & 4 & Belly, Stripe Distortion, Stripe Frequency, Stripe Shape \\
2-Color Spots & 4 & Offset, Spot Scale, Ratio, Noise \\
Mollusk & 8 & UV Pattern Ratio, Scale, Distortion, Pattern Type, Hue Range, Saturation Range, Value Range, Colors Per Pattern \\
\midrule
Num. Generators:  50 & Total: 271 &  \\
\bottomrule
\end{tabular}
\caption{Our full system contains 182 procedural asset generators and 1070 interpretable DOF. Here we show parameters for just our \textit{Material Generators}.} 
\label{thetab:dof_materials}
\end{table*}

\begin{table*}
\centering
\footnotesize
\begin{tabular}{p{0.2 \textwidth}|l|p{0.7\linewidth}} 
\toprule
\multirow{2}{*}{Terrain Generators} & Interpretable & \multirow{2}{*}{Named Parameters} \\
& DOF & \\
\midrule
3D Noise, Wind-Eroded Rocks & 0 &  \\
Caves & 3 & Cavern Size, Tunnel Frequency, Fork Frequency   \\
Voronoi Rocks, Grains & 2 & Rock Frequency, Warping Frequency \\
Sand Dunes & 2 & Dune Frequency, Warping Frequency \\
Mountains, Floating Islands & 2 & Mountain Frequency, Num. Scales \\
Coast line & 2 & Coast curve frequency, Height mapping function \\
Ground Slope & 0 &  \\
Still Water, Ocean & 0 &  \\
Atmosphere & 0 &  \\
Tiled Landscape & 0 &  \\
\midrule
Scene Types (Arctic, Canyon, Cave, Cliff, Waterfall, Coast, Desert, Mountain, Plain, River, Underwater, Volcano) & 6 & Tile Types, Tile Heights, Tile Frequency, Element Probabilities, Water Level, Snow \\
\midrule
Num. Generators:  26 & DOF: 17 &  \\
\bottomrule
\end{tabular}
\caption{Our full system contains 182 procedural asset generators and 1070 interpretable DOF. Here we show parameters for just our \textit{Terrain Generators}. Terrain is heavily simulation and noise-based, so has few interpretable DOF but uncountable internal complexity.} 
\label{theref:dof_terrain}
\end{table*}

\begin{table*}
\centering
\footnotesize
\begin{tabular}{p{0.2 \textwidth}|l|p{0.7\linewidth}} 
\toprule
Lighting, Weather & Interpretable & \multirow{2}{*}{Named Parameters} \\
\& Fluid Generators & DOF & \\
\midrule
Dust, Rain, Snow, Windy Leaves & 6 & Density, Mass, Lifetime, Size, Damping, Drag, \\
Cumulus, Cumulonimbus, Stratocumulus, Altocumulus & 13 & Density, Anisotropy, Noise Scale, Noise Detail, Voronoi Scale, Mix Factor, Increased Emission, Angular Density, Mapping Curve (6) \\
Atmospheric Fog, Dust & 5 & Density Min, Density Max, Color, Noise Scale, Anisotropy \\
Lava/Water & 6 & Viscocity, Viscocity Exponent, Surface Tension, Velocity Coord, Spray Particle Scale, Flip Ratio \\
Fire/Smoke & 12 & Max Temp, Gas Heat, Bouyancy, Burn Rate, Flame Vorticity, Smoke Vorticity, Dissolve Speed, Noise Scale, Noise Strength, Surface Emission, Turbulence Scale, Turbulence Strength \\
Sky Light & 8 & Overall Intensity, Sun Size, Sun Intensity, Sun Elevation, Altitude, Air Density, Dust Density, Ozone Density \\
Caustics & 5 & Scale, Sharpness, Coordinate Warping, Power, Spotlight Blending \\
Glowing Rocks & 3 & Wattage, Colors, Shape Distortion \\
Camera Lighting (Flashlight, Area Light) & 3 & Wattage, Light Size, Blending \\
\midrule
Num. Generators:  19 & DOF: 61 &  \\
\bottomrule
\end{tabular}
\caption{Our full system contains 182 procedural asset generators and 1070 interpretable DOF. Here we show parameters for just our \textit{Lighting, Weather and Fluid Generators}}.
\end{table*}

\begin{table*}
\centering
\footnotesize
\begin{tabular}{p{0.2 \textwidth}|l|p{0.7\linewidth}} 
\toprule
\multirow{2}{*}{Rock Generators} & Interpretable & \multirow{2}{*}{Named Parameters} \\
& DOF & \\
\midrule
Rocks & 3 & Aspect Ratio, Deform, Roughness \\
Stalagmite / Stalactite & 3 & Num. Extrusions, Length, Z Offset Variance \\
Boulder & 6 & Initial Vertices Count, Is Slab, Large Extrusion Probability, Small Extrusion Probability, Large Extrusion Distance, Small Extrusion Distance \\
\midrule
Num. Generators:  4 & DOF: 12 &  \\
\bottomrule
\end{tabular}
\caption{Our full system contains 182 procedural asset generators and 1070 interpretable DOF. Here we show parameters for just our \textit{Rock Generators}.} 
\end{table*}

\begin{table*}
\centering
\footnotesize
\begin{tabular}{p{0.2 \textwidth}|l|p{0.6\linewidth}} 
\toprule
Plant \& Underwater & Interpretable & \multirow{2}{*}{Named Parameters} \\
Generators & DOF & \\
\midrule
Flower & 8 & Center Radius, Petal Dimensions (2), Seed Size, Petal Angle Range (2), Wrinkle, Curl \\
Maple & 7 & Stem Curve Control Points, Stem Rot. Angle, Polar Mult. X, X Wave Control Points, Y Wave Control Points, Warp End Rad., Warp Angle \\
Pine & 4 & Midpoint (2), Length, X Angle Mean \\
Broadleaf & 14 & Midrib Length, Midrib Width, Stem Length, Vein Asymmetry, Vein Angle, Vein Density, Subvein Scale, Jigsaw Scale, Jigsaw Depth, 
Midrib Control Points, Shape Control Points, Vein Control Points, Wave X, Wave Y \\
Ginko & 11 & Stem Control Points, Shape Curve Control Points, Vein Length, Blade Angle, Polar Multiplier, 
Vein Scale, Wave, Scale, Margin Scale, X Wave Control Points, Y Wave Control Points \\
Pinecone & 13 & Bud Float Curve Angles, Bud Float Curve Scales, Bud Float Curve Z Displacements, Bud Instance Rotation Perturbation, Bud Instance Probability, Bud Instance Count, Profile Curve Radius, Max Bud Rotation, Rotation Frequency, Stem Height, Bright Color, Dark Color , Musgrave Scale \\
Urchin & 12 & Subdivision, Z Scale, Bevel Percentage, Spike Probability, Girdle Height, Extrude Height, Spike Scale, Base Color, Girdle Color, Spike Color, Transmission, Subsurface Ratio \\
Seaweed & 7 & Ocean Current, Deform Angle, Translation Scale, Expansion Scale, Bright Color, Dark Color, Musgrave Scale \\
Jellyfish & 16 & Cap Height, Cap Scale, Cap Perturbation Scale, Long Opaque Tentacles Count, Short Transparent Tentacles Count, Arm Screw Angle, Arm Screw Offset, Arm Taper Factor, Arm Displacement Strength, Arm Min Distance, Arm Placement Angle Threshold, Bright Color, Dark Color, Transparent Color, Fresnel Color, Musgrave Scale \\
Kelp & 14 & Ocean Current, Axis Shift, Axis Length, Axis Noise Stddev, Leaf Scale, Leaf Float Curve Length, Leaf Float Curve Width, Leaf Float Curve Z Displacement, Leaf Rotation Perturb, Leaf Tilt, Leaf Instance Probability, Leaf Instance Rotation Stride, Leaf Instance Rotation Interpolation Factor, Leaf Instance Count \\
Shells (Scallop, Clam, Mussel) & 9 & Top Control Point, Shell Interpolation Ratio, Shell Float Curve Angles, Shell Float Curve Scales, Radial Groove Scale, Radial Groove Frequency, Hinge Length, Hinge Width, Angle Between Shells \\
Snail (Volute, Nautlius, Conch) & 8 & Cross Section Affine Ratio, Cross Section Spiky Perturbation, Cross Section Concavity, Lateral Movement, Longitudinal Movement, Rotation Frequency, Scaling Ratio, Loop Count \\
Reaction Diffusion Coral & 10 & Intialization Bump Count, Initialization Bump Stride, Timesteps, Step size, Diffusion Rate A, Diffusion Rate B, Feed Rate, Kill Rate, Perturbation Scale, Smooth Scale \\
Tube Coral & 6 & Face Perturbation, Short Extrude Length Range, Long Extrude Length Range, Extrusion Direction Perturbation, Drag Direction, Extrusion Probability \\
Laplacian Coral & 8 & Timesteps, Kill Rate, Step size, Tau, Eps, Alpha, Gamma, Equilibrium Temperature \\
Tree Coral & 9 & Branch Count, Secondary Branch Count, Tertiary Branch Count, Horizontal Span, Length, Secondary Length, Tertiary Length, Base Radius, Radius Decay Ratio \\
Diff. Growth Coral & 8 & Colony Count, Max Polygons, Noise Factor, Step Size, Growth Scale, Drag Vector, Replusion Radius, Inhibit Shell Factor \\
Coral Tentacles & 7 & Min Distance, Z Angle Threshold, Radius Threshold, Density, Branch Count, Branch Length, Color \\
Grass Tuft & 10 & Num. Blades, Length Std., Curl Mean, Curl Std., Curl Power, Blade Width Variance, Taper Mean, Taper Variance, Base Spread, Base Angle Variance \\
Fern & 15 & Pinna Rotation (2), Pinnae Rotation (2), Pinnae Gravity, Age, Age Variety, Num Pinna, Pinnae Contour, Num Pinnae Varieties, Num Leaves, Leaf Width Randomness, Num Pinnae, Pinnae Rotation Randomness (2),  \\
Mushroom & 17 & Cross Section Float Curve Angles, Cross Section Float Curve Scales, Cross Section Center Offset, Cross Section Z Rotation, Stem Length, Stem Radius, Cap Groove Ratio, Cap Scale Ratio, Cap Radius Float Curve Height, Cap Radius Float Curve Radius, Has Web, Umbrella Radius, Umbrella Height, Bright Color, Dark Color , Light Color, Musgrave Scale \\
Flower Stem & 15 & Branch Leaf Rotation, Branch Leaf Instance, Branch Stem Radius, Branch Rotation Coeff, Branch Leaf Density, Stem Branch Density, Stem Branch Scale, Stem Branch Range, Stem Leaf Instance, Stem Leaf Rotation, Stem Flower Instance, Stem Flower Scale, Stem Rotation Coeff. STem Radius, Num Versions, Rotation Z \\
Cactus Spikes & 6 & Branches Count, Secondary Branches Count, Min Distance, Top Cap Percentage, Density, Color \\
Globular Cactus & 5 & Groove Scale, Groove Count, Rotation Frequency, Profile Curve Height, Profile Curve Radius \\
Columnar Cactus & 9 & Radius Decay Branch, Radius Decay Root, Radius Smoothness Leaf, Branch Count, Nodes Per Branch, Nodes Per Second Level Branch, Base Radius, Perturbation Scale, Groove Scale \\
Pricky Pear Cactus & 5 & Leaf Profile Curve Width, Leaf Profile Curve Height, Leaf Instance Scale Ratio, Leaf Instance Placement Angles, Leaf Instance Count \\
\midrule
Num. Generators: 30 & DOF: 258 &  \\
\bottomrule
\end{tabular}
\caption{Our full system contains 182 procedural asset generators and 1070 interpretable DOF. Here we show parameters for just our \textit{Plant and Underwater Invertebrate Generators}.} 
\end{table*}

\begin{table*}
\centering
\footnotesize
\begin{tabular}{p{0.2 \textwidth}|l|p{0.7\linewidth}} 
\toprule
\multirow{2}{*}{Creature Generators} & Interpretable & \multirow{2}{*}{Named Parameters} \\
& DOF & \\
\midrule
Geonodes Quadruped Body & 12 & Start Rad., End Rad., Ribcage Proportions (3), Flank Proportions (3), Spine Coeffs (3), Aspect Ratio \\
Geonodes Bird Body & 3 & Start Rad., End Rad. Aspect Ratio, Fullness \\
Geonodes Carnivore Head  & 15 & Start Rad., End Rad., Snout Proportions (3), Aspect Ratio, Lip Muscle Coeff. (3) Jaw Muscle Coeff. (3), Forehead Muscle Coeff (3) \\
Geonodes Neck & 5 & Start Rad. End Rad., Neck Muscle Coeffs. (3) \\
NURBS Bodies/Heads (Carnivore, Herbivore, Fish, Beetle and Bird) & 8 & Start/End Dims (4), Proportions, Angle Offsets, Profile Offset, Bump Offset,  \\
Jaw & 9 & Rad1, Rad2, Width Shaping, Canine Size, Incisor Size, Tooth Density, Tooth Crookedness, Tongue Shaping, Tongue X Scale \\
QuadrupedBackLeg & 15 & Start Rad., End Rad., Aspect Ratio, Thigh Coeffs (6), Calf Coeffs (6), \\
QuadrupedFrontLeg & 21 & Start Rad., End Rad., Aspect Ratio, Shoulder Coeffs. (6), Forearm Coeffs (6), Elbow Coeffs (6) \\
Bird Leg & 9 & Start Rad., End Rad., Aspect Ratio, Thigh Coeffs (3), Shin Coeffs (3) \\
Insect Leg & 9 & Start Rad., End Rad., Carapace Rad. Spike Length, Spike Start Rad., Spike End Rad., Spike Range (2), Spike Density \\
Ridged Fin & 10 & Width, Roundness, Ridge Frequency, Offset Weight (2), Ridge Rot., Affine (2), Noise Ratio (2) \\
Feather Tail & 9 & Feather Dims (3), Max Rotation (3), Rotation Randomness (3) \\
Feather Wing & 7 & Start Rad., End Rad., Feather Density, Feather Form Sculpting, Wing Extendedness, Feather Rot. Randomness (2) \\
Beak & 17 & Curve Y, Curve Z, Hook Coeff. (2), Hook scale (2), Hook Pos. (2), Hook Thickness (2), Crown Scale, Crown Coeff. (2), Bump Scale, Bump L, Bump R, Sharpness \\
MammalEye & 9 & Radius, Eyelid Thickness, Eyelid Fullness, Tear Duct Placement (3), Eye Corner Placement (3) \\
Ear & 5 & Start Rad., End Rad., Depth, Thickness, Curl Angle \\
Insect Mandible & 4 & Start Rad, End Rad, Curl, Aspect Ratio \\
Nose & 3 & Radius, Nostril Size, Smoothness \\
Hoof & 8 & Claw Y Scale, Claw Z Scale, Claw Sag, Angle Length, Angle Rad. Start, Ankle Rad. End, Upper Shape, Lower Shape \\
Horn & 6 & Rad. Start, Rad. End, Ridge Thickness, Ridge Density, Ridge Depth, Height \\
Foot & 12 & Start Rad., End Rad., Toe Density, Toe Dimensions (3), Toe Splay, Footpad Radius, Claw Curl, Claw Dimensions (3) \\
Tail & 4 & Start Rad., End Rad., Curl, Aspect Ratio \\
Cotton, Skin, Rubber Simulation & 8 & Max Bending Stiffness, Max Compression Stiffness, Goal Spring Force, Pin Stiffness, Shear Stiffness (2), Tension Stiffness, Pressure \\
Running Animation & 6 & Steps Per Second, Stride Length, Gait spread, Stride Height, Upturn, Downstroke \\
Short Hair, Fluffy Hair, Feathers & 18 & Clump Num., Avoid Eyes Dist., Avg. Length, Avg. Puff, Length Noise (2), Puff Noise (2), Combing, Strand Noise (3), Tuft Spread, Tuft Clumping, Hair Radius, Intra-clump Noise, Length Falloff,  Roughness \\
Carnivore Genome & 15 & Head Ratio, Head Attachment, Jaw Ratio, Jaw Attachment, Eye Attachment (3), Nose Attachment, Ear Attachment (3), Shoulder Dist., Shoulder Splay, Leg Ratio \\
Herbivore Genome & 22 & Neck Start T., Hoof Angle, Foot Angle, Head Interp Temp., Jaw Ratio, Jaw Attachment, Eye Attachment (3), Nose Attachment, Ear Attachment (3), Shoulder Dist., Shoulder Splay, Leg Ratio, Include Nose, Include Horns, Horn Attachment (3), Body Interp Temp \\
Bird Genome & 17 & Head Ratio, Head Attachment, Tail Attachment (2), Leg Length Ratio, Foot Size Ratio, Leg Attachment (3), Wing Length Ratio, Wing Attachment (3), Head Ratio, Eye Attacment (3) \\
Insect Genome & 8 & Leg Density, Leg Splay, Leg Length Ratio, Include Mandibles, Mandible Attachment (3), Has Hair \\
Fish Genome & 11 & Dorsal Fin Ratio, Pelvic Fin Ratio, Pectoral Fin Ratio, Hind Fin Ratio, Fin Attachment (3), Eye Attachment (3) Body Interp Temp. \\
Random Genome & 10 & Has Wings, Locomotion Type, Hair Type, Interp Temperature, Head Type, Has Eyes, Nose Type, Has Jaw, Has Ears, Has Horns \\
\midrule
Num. Generators: 39 & DOF: 315 &  \\
\bottomrule
\end{tabular}
\caption{Our full system contains 182 procedural asset generators and 1070 interpretable DOF. Here we show parameters for just our \textit{Creature Generators}.} 
\end{table*}

\begin{table*}
\centering
\footnotesize
\begin{tabular}{p{0.2 \textwidth}|l|p{0.7\linewidth}} 
\toprule
\multirow{2}{*}{Tree Generators} & Interpretable & \multirow{2}{*}{Named Parameters} \\
& DOF & \\
\midrule
Random Tree, Pine Tree, Bush & 26 & Growth Height, Trunk Warp, Num. Trunks, Branching Start, Branching Angle, Branching Density, Branch Length, Branch Warp, Pull Dir. Vertical, Pull Dir. Horizontal, Outgrowth, Branch Thickness, Twig Density, Twig Scale, Twig Pts, Twig Branching Start, Twig Rot. Randomness, Twig Branching Density, Twig Init Z,Twig Z Randomness, Twig Subtwig Size, Twig Subtwig Momentum, Twig Subtwig Std., Twig Size Decay, Twig Pull Factor, Space Colonization Shape \\
\midrule
Num. Generators:  3 & DOF: 26 &  \\
\bottomrule
\end{tabular}
\caption{Our full system contains 182 procedural asset generators and 1070 interpretable DOF. Here we show parameters for just our \textit{Tree Generators}.} 
\end{table*}

\begin{table*}
\centering
\footnotesize
\begin{tabular}{p{0.2 \textwidth}|l|p{0.7\linewidth}} 
\toprule
\multirow{2}{*}{Scene Composition Generators} & Interpretable & \multirow{2}{*}{Named Parameters} \\
& DOF & \\
\midrule
Scene Generators (Arctic, Canyon, Cave, Cliff, Coast, Desert, Forest, Mountain, Plain, River, Under Water) & 110 & Asset/Scatter inclusion probabilties (39), Num. Creature/Plant Subspecies (4), Noise Mask Scales (30),  Mask Tapering Coeff. (12), Normal Mask Thresholds (14), Placement Densities (5), Placement Habitats (6) \\
\midrule
Num. Generators: 11 & DOF: 110 &  \\
\bottomrule
\end{tabular}
\caption{Our full system contains 182 procedural asset generators and 1070 interpretable DOF. Here we show parameters for just our \textit{Scene Composition Configs}. Each config references a terrain composition generator from Fig. \ref{theref:dof_terrain}, and specifies a realistic distribution of other assets to create a fully realistic natural environment.} 
\label{thetab:dof_scenecomp}
\end{table*}